\ifdefined\pdfminorversion
\fi
\documentclass{article}

\PassOptionsToPackage{comma,authoryear,compress}{natbib}

\usepackage[eandd,preprint]{neurips_2026}

\usepackage[utf8]{inputenc}
\usepackage[T1]{fontenc}
\usepackage[svgnames]{xcolor}
\usepackage{hyperref}
\usepackage{url}
\usepackage{microtype}
\usepackage{graphicx}
\usepackage{stfloats}
\usepackage{booktabs}
\usepackage{amsmath}
\usepackage{amssymb}
\usepackage{mathtools}
\usepackage{enumitem}
\usepackage{array}
\usepackage{tabularx}
\usepackage{longtable}
\usepackage{needspace}
\usepackage{placeins}
\usepackage[most]{tcolorbox}
\usepackage[font=small,labelfont=bf]{caption}
\usepackage{subcaption}
\let\standardcenter\center
\let\endstandardcenter\endcenter
\renewenvironment{center}{\setlength{\topsep}{2pt}\standardcenter}{\endstandardcenter}

\definecolor{YaleBlue}{RGB}{0,53,107}
\definecolor{promptpurple}{HTML}{6B006B}
\definecolor{promptbg}{HTML}{FFFDFB}

\newtcblisting{promptlisting}[2][]{%
    enhanced,
    colback=promptbg,
    colframe=promptpurple,
    coltitle=white,
    colbacktitle=promptpurple,
    fonttitle=\bfseries\small,
    title={#2},
    boxrule=0.9pt,
    arc=2mm,
    left=1.5mm,
    right=1.5mm,
    top=2.5mm,
    bottom=1.2mm,
    before skip=5pt,
    after skip=6pt,
    attach boxed title to top left={yshift=-2mm,xshift=3mm},
    boxed title style={
        colback=promptpurple,
        colframe=promptpurple,
        arc=1mm,
        boxrule=0pt,
        left=2mm,
        right=2mm,
        top=0.5mm,
        bottom=0.5mm
    },
    listing only,
    listing options={
        basicstyle=\ttfamily\scriptsize,
        breaklines=true,
        columns=fullflexible,
        keepspaces=true,
        showstringspaces=false
    },
    #1
}

\hypersetup{
    colorlinks=true,
    citecolor=YaleBlue,
    urlcolor=YaleBlue,
    linkcolor=black,
    pdftitle={VEHBench: A Stage-Local Diagnostic Benchmark for LLM-Assisted Vibration Energy Harvester Design},
    pdfauthor={Depeng Su, Yuyu Luo, Guobiao Hu},
    pdfsubject={A stage-local benchmark for LLM-assisted vibration energy harvester design},
    pdfkeywords={large language models, engineering design, vibration energy harvesting, benchmark, evaluation}
}

\definecolor{blanchedalmond}{rgb}{1.0, 0.92, 0.8}
\definecolor{carmine}{rgb}{0.59, 0.0, 0.09}
\definecolor{lightblue}{rgb}{0.22,0.45,0.70}%

\renewcommand{\mathbf}{\boldsymbol}

\makeatletter
\def\Ddots{\mathinner{\mkern1mu\raise\p@
\vbox{\kern7\p@\hbox{.}}\mkern2mu
\raise4\p@\hbox{.}\mkern2mu\raise7\p@\hbox{.}\mkern1mu}}
\makeatother

\definecolor{amaranth}{rgb}{0.9, 0.17, 0.31}
\definecolor{antiquebrass}{rgb}{0.8, 0.58, 0.46}
\definecolor{antiquefuchsia}{rgb}{0.57, 0.36, 0.51}
\definecolor{chromeyellow}{rgb}{0.31, 0.47, 0.26}

\newcolumntype{Y}{>{\raggedright\arraybackslash}X}

\title{VEHBench: A Stage-Local Diagnostic Benchmark for LLM-Assisted Vibration Energy Harvester Design}

\author{
  Depeng Su\textsuperscript{1} \quad
  Yuyu Luo\textsuperscript{2,*} \quad
  Guobiao Hu\textsuperscript{1,*}\\
  \normalfont\small
  \textsuperscript{1}Internet of Things Thrust, Information Hub\\
  \textsuperscript{2}Data Science and Analytics Thrust, Information Hub\\
  The Hong Kong University of Science and Technology (Guangzhou)\\
  \textsuperscript{*}Corresponding authors:
  \texttt{\{yuyuluo,guobiaohu\}@hkust-gz.edu.cn}
}

\begin{document}

\maketitle

\begin{abstract}
Battery-free Internet of Things (IoT) requires iterative design of vibration energy harvesters (VEHs) under coupled physical constraints, while LLMs are emerging as interface layers for engineering workflows. However, existing engineering benchmarks primarily assess final artifact validity, offering limited insights into how LLMs behave across different stages of coupled physical design. We introduce VEHBench, an engineering-native diagnostic benchmark for LLM-assisted VEH design, featuring 763 literature-grounded tasks scored by an analytical physical oracle. VEHBench evaluates four design roles: specification triage, verifier-guided search, corrupted-state recovery, and policy-conditioned selection. Experimental results reveal that LLM capability is strongly stage-dependent: no single model consistently dominates the entire workflow, and response-control profiles expose distinct behavioral patterns across design roles. VEHBench thus provides a stage-aware foundation for evaluating, selecting, routing, and improving verifier-grounded engineering LLMs. The benchmark artifact is available at \url{https://huggingface.co/datasets/AnonymousVehbench/vehbench}.
\end{abstract}

\section{Introduction}
\label{sec:intro}

Battery-free IoT reduces maintenance for dense, long-lived sensing, making vibration energy harvester (VEH) design a recurring need for low-power sensors~\citep{citroni2024efficient,zeadally2020design,naifar2024energy}. VEHs convert ambient vibration into power, but each deployment can impose different vibration spectra, power targets, size limits, materials, packaging conditions, and safety margins~\citep{safaei2019review,sadaf2024cantilever,sadaf2025harnessing}. LLMs are increasingly used as engineering workflow interfaces to translate requirements, review feedback, revise candidates, and compare trade-offs; this paper explores how assistance should be evaluated in physics-grounded VEH design.

\begin{figure*}[!t]
\centering
\includegraphics[width=\linewidth]{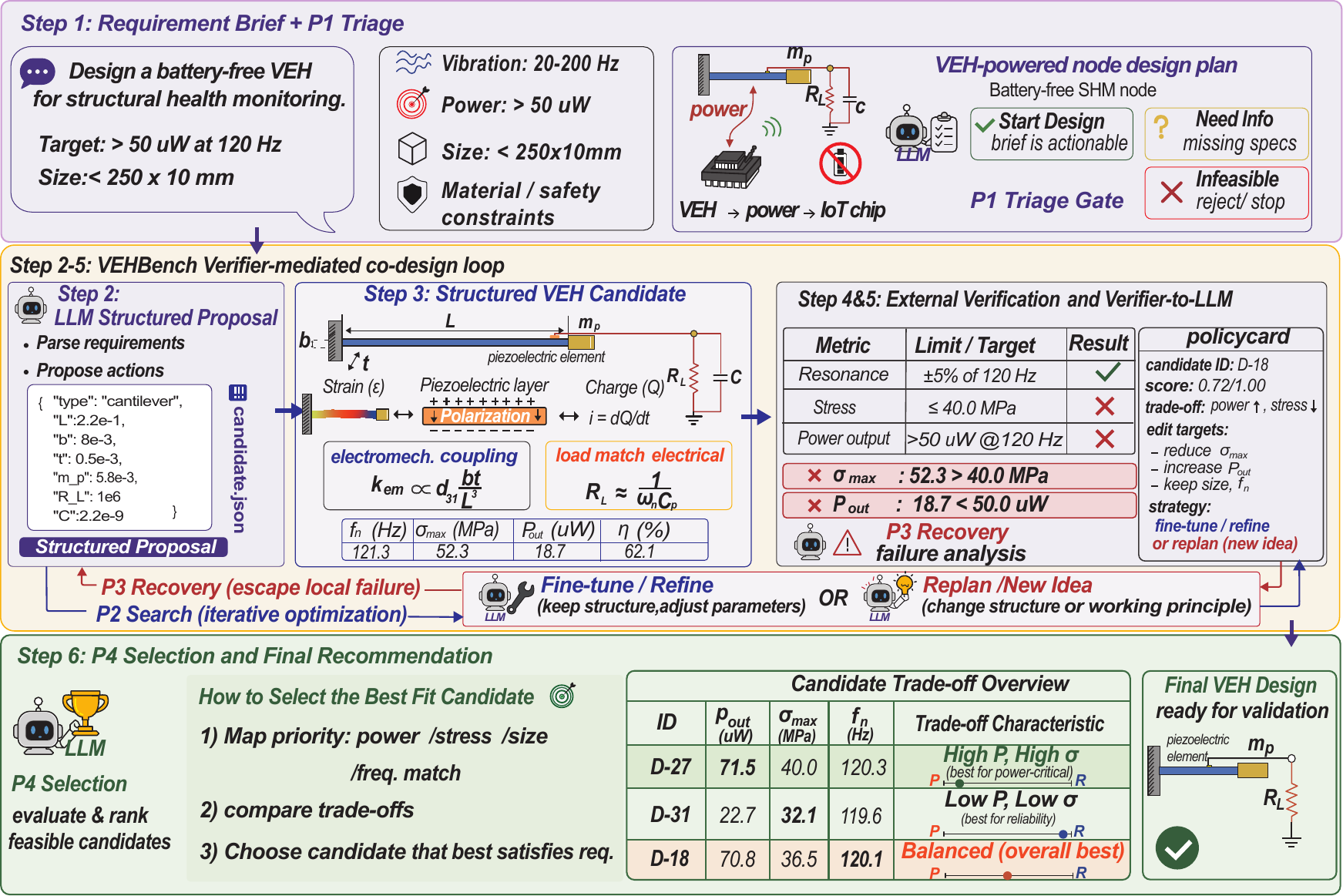}
\vspace{-2mm}
\caption{Coupled VEH co-design workflow and LLM intervention points. VEHBench evaluates how an LLM assists the verifier-grounded design loop through stage-local roles rather than treating the model as a standalone hardware designer.}
\label{fig:vehbench_workflow}
\vspace{-2mm}
\end{figure*}

VEH provides a compelling testbed for evaluating LLM-assisted engineering design since it involves tightly coupled mechanical–electrical interactions and demands expertise across structural dynamics, smart materials, circuit design, and embedded systems. For example, a cantilever-based piezoelectric energy harvester is a representative VEH due to its canonical role in the literature. Its performance is jointly governed by structural parameters (e.g., beam geometry, proof mass, and material properties) and electrical parameters (e.g., load resistance, capacitance, and electromechanical coupling), leading to strong trade-offs between objectives such as power output and bandwidth. As a result, VEH differs fundamentally from retrieval-based tasks or single-objective optimization. Figure~\ref{fig:vehbench_workflow} illustrates how an LLM can assist this coupled workflow by interpreting requirements, proposing design modifications, analyzing verifier feedback, and comparing feasible alternatives while the physical oracle remains outside the model. In this work, we propose a VEHBench to evaluate workflow assistance under physical constraints, not certified autonomous hardware design.

Current engineering evaluations have progressed by introducing external verification, including simulation-backed and physics-grounded checks~\citep{guo2025engdesign,xia2025buildarena,jadhav2024mechanical,doris2024designqa,jain2024mseval}. These benchmarks move beyond textual judgment by asking whether generated artifacts satisfy engineering constraints, but endpoint validity alone does not reveal whether a model checked brief sufficiency before search, made bounded edits after physical feedback, recovered from a misleading trajectory, or followed an explicit policy when several candidates were already feasible. These are workflow failures, not just solver failures. Recent AI evaluation has shown the value of decomposing aggregate ability into task-local roles and error types~\citep{srivastava2022bigbench,liang2023helm,kiela2021dynabench}; engineering design needs this idea in an engineering-native diagnostic layer, where stages are workflow boundaries with different trusted states, admissible actions, verifier signals, and failure consequences. For LLM-assisted engineering, a benchmark should therefore evaluate whether the model behaves appropriately at the design stage where it is used, rather than only whether the final artifact passes. We study three research questions.

\begin{itemize}[leftmargin=5mm,itemsep=2mm,topsep=1mm]
    \item \textbf{Q1: How can we evaluate LLMs in specialized engineering design tasks?} Because standardized datasets for VEH co-design remain scarce, meaningful evaluation requires the construction of an engineering-native benchmark incorporating real constraints, external verification, and stage-specific workflow assessment.
    \item \textbf{Q2: How capable are current LLMs in VEH co-design, and what insights can be gained from analyzing their failure modes?} This requires separating design responsibilities and diagnosing behavior beyond aggregate scores.
    \item \textbf{Q3: How can VEHBench support engineering applications, identify targets for improvement, and enable broader evaluation?} This requires evidence that stage-local results can inform model selection, routing, adaptation, and future verifier-grounded evaluations.
\end{itemize}

\paragraph{The first VEH-focused diagnostic benchmark.}
To answer these questions, we introduce VEHBench, an engineering-native diagnostic benchmark that evaluates LLM-assisted vibration energy harvester co-design through stage-local, verifier-grounded design tasks. Its 763 cases are not meant to compete with web-scale language benchmarks in volume; they are literature-grounded, oracle-checkable, manifest-traced engineering probes. VEHBench is built from literature-derived VEH design anchors and scored by an analytical oracle that checks physical feasibility and objective quality. Rather than treating VEH design as one endpoint task, the benchmark decomposes the workflow into four design roles: specification triage, verifier-guided search, corrupted-state recovery, and policy-conditioned selection. The detailed probe definitions are introduced in the benchmark construction section; the introduction uses them only to establish the evaluation problem. Our contributions include:

\begin{itemize}[leftmargin=5mm,itemsep=1mm,topsep=1mm]
    \item \textbf{Benchmark framework and VEHBench construction.} We propose an engineering-native diagnostic benchmark for LLM-assisted design and instantiate it as VEHBench for coupled mechanical--electrical VEH co-design. The benchmark combines literature-grounded task construction, external analytical verification, and stage-specific evaluation of design workflow behavior.

    \item \textbf{Empirical findings and interpretation framework.} We systematically evaluate current LLMs across the VEH design workflow and analyze their behavior through response-control profiles. This framework links empirical results to interpretable model behavioral characteristics, including action discipline, bounded editing, feedback conditioning, state recovery, and policy execution.

    \item \textbf{Stage-aware guidance for engineering application and model improvement targets.} We demonstrate how the benchmark can guide practical LLM-assisted design: engineers can select, prompt, route, or adapt models according to different stages of the design workflow, while AI researchers can identify critical capability gaps for engineering agents, such as specification triage, verifier-guided search, corrupted-state recovery, or policy-conditioned ranking.
\end{itemize}

\section{Related Work}
\label{sec:related_work}

\textbf{LLM-assisted engineering design.}
Recent engineering benchmarks evaluate LLMs on tasks closer to design practice. EngDesign studies whether models satisfy engineering design requirements; BuildArena emphasizes artifact construction and interaction with executable feedback; mechanical-design agent studies test iterative CAD/CAE-style reasoning; DesignQA targets engineering-document understanding; and MSEval evaluates material-selection behavior~\citep{guo2025engdesign,xia2025buildarena,jadhav2024mechanical,doris2024designqa,jain2024mseval}. Together, these works move LLM evaluation beyond linguistic plausibility toward engineering constraints, external evidence, and physically meaningful artifacts. Their main readout, however, usually remains final satisfaction, artifact validity, or task completion, making it hard to identify which design-stage behavior failed.

\textbf{Vibration energy harvesting and VEHBench positioning.}
VEH is a meaningful first substrate for LLM-assisted engineering evaluation because it is practically motivated, constraint-coupled, and analytically auditable. Modern reviews emphasize vibration and cantilever piezoelectric harvesters as compact power sources for wireless, embedded, MEMS, and IoT systems~\citep{safaei2019review,maamer2019review,sadaf2024cantilever,sadaf2025harnessing}. VEHBench follows the engineering-benchmark direction but shifts the readout from final artifact validity to stage-local design behavior. It therefore complements existing benchmarks by using an analytically checkable domain to expose where workflow behavior fails rather than treating the LLM as a single end-to-end designer; Appendix~\ref{sec:appendix_extended_related_work} expands the comparison.

\section{VEHBench: Design and Construction}
\label{sec:method}

\subsection{Design Goals, Domain Scope, and Anchors}

\paragraph{Goals.}
VEHBench evaluates LLM assistance in verifier-grounded engineering design, not autonomous hardware certification. We set four goals. \textbf{(G1) Engineering-grounded controllability:} tasks should embed realistic coupled design constraints while remaining compact enough for reproducible, deterministic evaluation. \textbf{(G2) Stage-local diagnosticity:} the benchmark should isolate failures at distinct workflow boundaries---specification, search, recovery, and selection---rather than collapse behavior into a single final-artifact score. \textbf{(G3) External and reproducible scoring:} all metrics are computed by an analytical physical oracle from structured outputs and persisted logs, without human labels or LLM-as-a-judge. \textbf{(G4) Auditable extensibility:} construction preserves source provenance, manifests, splits, oracle traces, prompts, and evaluator metadata, so the same scaffold can extend to richer simulators or other coupled domains.

\paragraph{Domain scope and anchors.}
We instantiate this framework on early-stage cantilever VEH co-design, a compact but coupled battery-free IoT setting with deployment-specific vibration spectra, power targets, size limits, materials, packaging, and safety margins~\citep{citroni2024efficient,zeadally2020design,sadaf2025harnessing}. VEHBench does not replace FEM, manufacturing review, or hardware certification; it evaluates four stage-local roles under analytical verification---specification triage, verifier-guided search, corrupted-state recovery, and policy-conditioned selection---summarized in Table~\ref{tab:inventory}. A \emph{design anchor} is a literature-derived VEH physical state that has been cleaned into normalized variables, units, bounds, assumptions, objectives, and constraints so that it can be scored by the external oracle. These anchors are the source of the stage-local probes rather than free-form prompt templates.

\subsection{Probe Design: Matching Design Stages to Measurements}

The four roles span distinct information boundaries within a design workflow: a design brief, verifier feedback during search, a corrupted trajectory, and a pool of feasible alternatives. Each boundary imposes a different trusted state, admissible action, and failure consequence. VEHBench turns these distinctions into probes by changing the \emph{trusted state}, \emph{allowed action}, and \emph{metric} while keeping the physical design substrate fixed. The probes are therefore not four difficulty levels of the same task; they are four different design roles over the same coupled physical object. This design makes the benchmark useful not only for comparing models, but also for deciding where and how an LLM should be inserted into a verifier-grounded engineering workflow.

VEHBench defines four task families, summarized in Table~\ref{tab:inventory}. \textbf{P1} occurs before search: the trusted state is only the design brief, the model must decide whether to propose, abstain, or request missing information, and the metric measures action discipline. \textbf{P2} occurs during search: the trusted state is a seed design plus oracle feedback, the model may make bounded candidate edits, and the metric measures feedback-conditioned feasible improvement. \textbf{P3} occurs after the design history has become unreliable: the model must reset, re-anchor, or stabilize rather than blindly continue, and the metric measures recovery from contaminated state. \textbf{P4} occurs after feasibility has already been established: the trusted state is an oracle-feasible candidate pool, the model ranks or selects under an explicit policy, and the metric measures policy execution. The same VEH object is shared; the model role and failure consequence change.

\begin{table*}[!t]
\caption{VEHBench probe inventory.}
\label{tab:inventory}
\centering
\scriptsize
\setlength{\tabcolsep}{3.2pt}
\renewcommand{\arraystretch}{1.08}
\begin{tabularx}{\textwidth}{l p{24mm} Y Y Y Y}
\toprule
Probe & Design stage & Input state & Model action & What it measures & Practical use \\
\midrule
P1 & Specification triage & VEH brief with complete, missing, or infeasible fields & Propose, abstain, or request information & Action discipline & Gate search before design begins \\
P2 & Search & Seed design plus oracle feedback & Make bounded edits & Feedback conditioning and feasible improvement & Improve candidate designs \\
P3 & Recovery & Corrupted or misleading design trajectory & Reset, re-anchor, or stabilize & State recovery & Prevent failed-trajectory cascade \\
P4 & Selection & Feasible candidate pool plus explicit policy & Rank or select candidates & Policy execution & Choose a final design under deployment priorities \\
\bottomrule
\end{tabularx}
\end{table*}

\subsection{Anchor-Based Construction Pipeline with Built-in Quality Control}

\begin{figure*}[!t]
\centering
\includegraphics[width=1\linewidth]{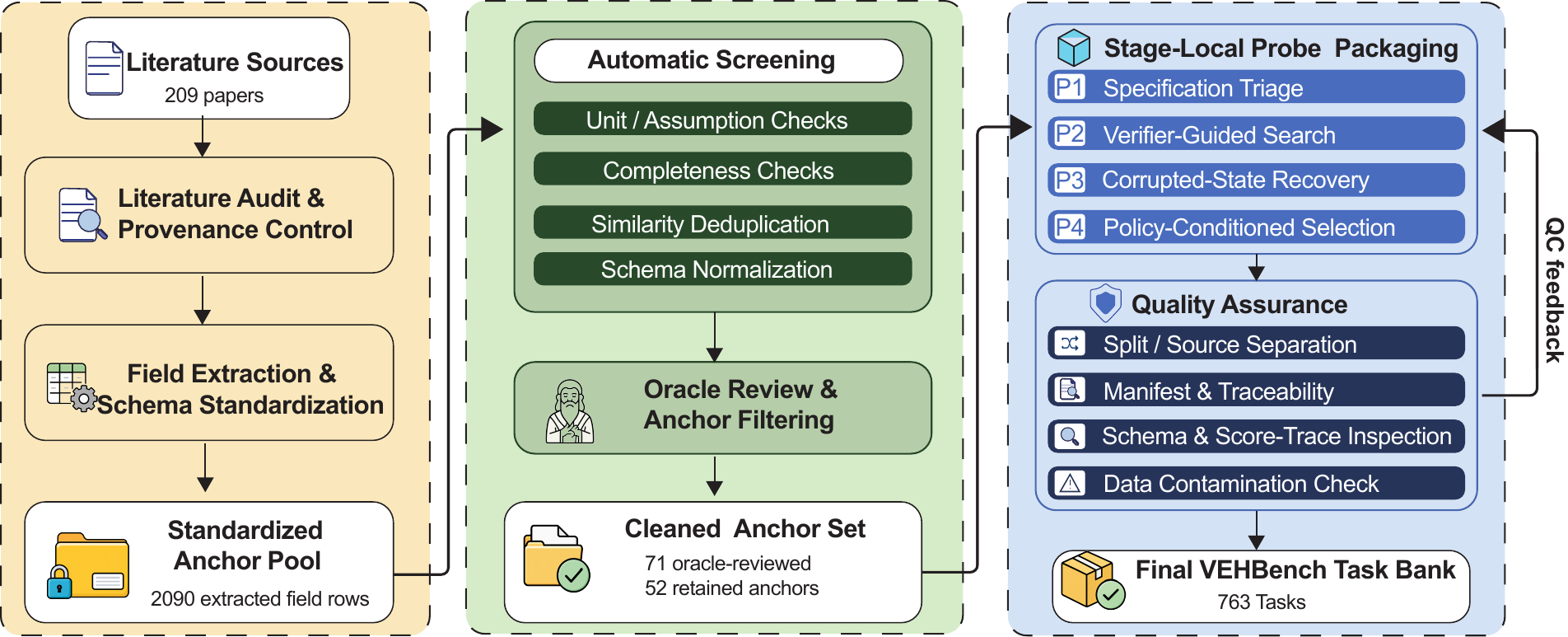}
\vspace{-3mm}
\caption{Benchmark construction pipeline from literature audit to manifest-backed release, with quality-control gates throughout.}
\label{fig:construction_pipeline}
\vspace{-3mm}
\end{figure*}

VEHBench uses \textbf{anchor-based probe generation}. Instead of freely writing engineering prompts, we start from literature-derived VEH design anchors, clean them into oracle-checkable physical states, and then generate stage-specific probes around those states. This keeps the benchmark grounded in engineering literature while allowing controlled variation over workflow stage. Figure~\ref{fig:construction_pipeline} summarizes the full construction path from literature audit to manifest-backed release.

The construction pipeline has six quality-controlled steps.

\textbf{Step 1: Literature audit and provenance control.}
We audit VEH papers and extract fields relevant to cantilever energy harvesting, including geometry, material, layer structure, excitation, target frequency, load resistance, power, stress, displacement, and feasibility constraints. Each extracted record is tracked by source metadata so that later tasks can be traced back to their construction path.

\textbf{Step 2: Issue screening and unit control.}
Extracted records are screened for missing units, inconsistent assumptions, incomplete variables, ambiguous operating conditions, and unsupported formulations. Records that cannot be made physically checkable are excluded before task generation.

\textbf{Step 3: Oracle review and physics control.}
Candidate anchors are reviewed against the analytical VEH oracle. Only anchors whose assumptions and variables can support feasibility and objective scoring are retained. This prevents task labels depending on unverifiable or internally inconsistent physics.

\textbf{Step 4: Anchor cleaning and schema control.}
Accepted anchors are converted into canonical design contexts with normalized variables, bounds, assumptions, objectives, and constraint definitions. This step ensures that tasks differ because of design role, not accidental formatting or schema artifacts.

\textbf{Step 5: Probe packaging and split control.}
Cleaned anchors are packaged into P1--P4 task families with stage-specific prompts, admissible actions, metrics, and oracle-facing metadata. P2--P4 use source-anchor separation for optimizer-facing splits, while P1 is constructed as a matched certification-style triage stage.

\textbf{Step 6: Release packaging and reproducibility control.}
The final tasks are released with manifests, evaluator-facing metadata, oracle traces, prompt templates, generation scripts, split reports, and model-log formats. This makes each task auditable at the source, schema, split, and evaluation levels.

\vspace{-5mm}
\subsection{Oracle Scoring, Data Quality, and Release}

VEHBench keeps physical evaluation outside the language model. Each task is generated from a cleaned literature anchor and specifies design variables, bounds, objectives, budgets, and any exposed design history. Model outputs are parsed into structured actions and passed to the analytical oracle, which returns feasibility, active violations, objective values, and task-specific feedback. The 763 released tasks are derived from a 209-paper audit and 52 cleaned anchors, screened for unit and assumption consistency and packaged into stage-local probes. The resulting benchmark is therefore high-density and high-quality: every task traces to a real VEH parameter set, and every score is computed by a deterministic physical verifier rather than by human judgment. This pipeline protects two forms of validity: physical validity, via deterministic oracle scoring; and diagnostic validity, via stage-specific probe design. The oracle supports P1 triage, P2 feasible search and objective quality, P3 corrupted-state recovery, and P4 policy-conditioned ranking over an oracle-feasible pool. The release mirrors this quality chain with task banks, manifests, evaluator code, prompts, splits, and per-model logs, making every reported score traceable to a source anchor and deterministic physical verifier.

\section{Empirical Findings: Stage Boundaries Create Distinct Failure Regimes}
\label{sec:experiments}

\subsection{Experimental Setup}

We evaluate 12 complete model runs on the 763-task VEHBench benchmark. Each run covers P1--P4. Full prompts, splits, run modes, and metric definitions are reported in the appendix.

\paragraph{Scoring and profiles.}
All scores are deterministic functions of persisted JSONL logs; no score uses human annotation or LLM-as-a-judge. Headline metrics are P1-Composite, P2 final feasible power ratio, P3-Success, and P4 Kendall \(\tau_b\). P2 assigns zero to infeasible finals; P3 counts final feasible recovery after the full budget; P4 ranks a fixed feasible pool against oracle policy order. Higher is better for headline scores; Figure~\ref{fig:stage_error_landscape} reports lower-is-better, non-exclusive error rates. Statistical intervals use nonparametric bootstrap, Wilson intervals, and paired bootstrap where appropriate. To diagnose stage-specific behavior and failure modes, we extract response-control profiles from the same logs: action discipline, edit style, feedback conditioning, state-reset effort, and policy execution. Profiles are used only for diagnosis, not ranking; Appendix~\ref{sec:master_metric_index} indexes all metrics and Appendix~\ref{sec:metric_formulas} gives full definitions.

\subsection{Finding 1: Stage Boundaries Change Rankings and Failure Modes}

\begin{table*}[t]
\caption{Headline VEHBench results over 12 complete model runs. Columns report P1-Composite, P2 final feasible power ratio, P3-Success, and P4 Kendall \(\tau_b\); higher is better.}
\label{tab:main_results}
\centering
\footnotesize
\setlength{\tabcolsep}{5pt}
\renewcommand{\arraystretch}{1.08}
\begin{tabular}{lrrrr}
\toprule
Model & P1 Comp. \(\uparrow\) & P2 ratio \(\uparrow\) & P3 Succ. \(\uparrow\) & P4 Tau \(\uparrow\) \\
\midrule
qwen3-max & \textbf{0.574} & 0.1955 & 30.1\% & 0.835 \\
gemini-3.1-pro-preview & 0.549 & \textbf{0.3904} & 37.2\% & 0.824 \\
\texttt{o4-mini} & 0.518 & 0.1551 & 26.3\% & 0.780 \\
deepseek-r1 & 0.504 & 0.2413 & 42.9\% & 0.833 \\
gpt-5.4 & 0.428 & 0.1329 & 42.3\% & \textbf{0.887} \\
hunyuan-hy3-preview & 0.425 & 0.1609 & \textbf{47.4\%} & 0.839 \\
deepseek-v3 & 0.369 & 0.1565 & 35.3\% & 0.860 \\
\texttt{llama-3.3-70b} & 0.361 & 0.1197 & 3.2\% & 0.714 \\
mimo-v2.5-pro & 0.317 & 0.1073 & 45.5\% & 0.843 \\
\texttt{qwen3.6-plus} & 0.306 & 0.2063 & 27.6\% & 0.877 \\
deepseek-v4-pro & 0.306 & 0.1294 & 28.8\% & 0.794 \\
claude-4.6-sonnet & 0.207 & 0.2394 & 16.0\% & 0.840 \\
\bottomrule
\end{tabular}
\end{table*}

\begin{figure*}[!t]
\centering
\includegraphics[width=\linewidth]{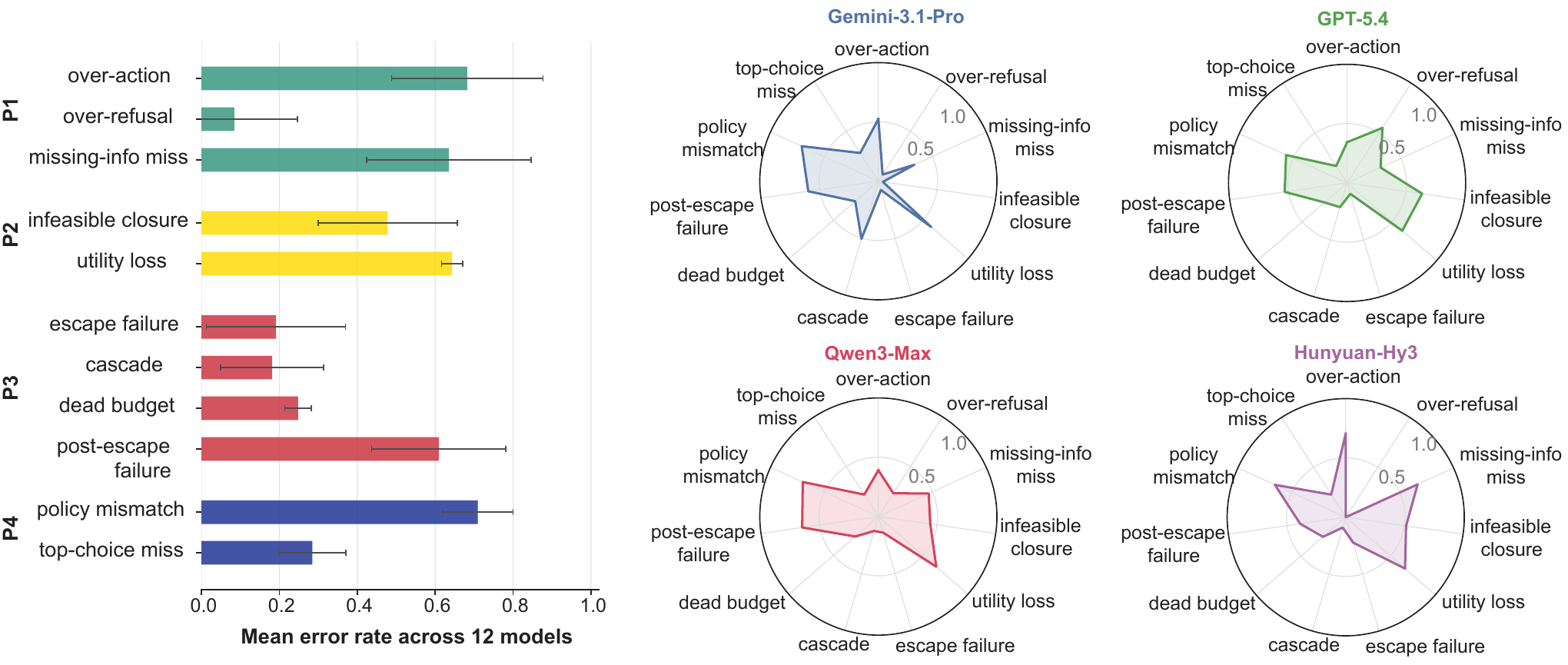}
\vspace{-4mm}
\caption{Stage-level error families and representative profiles. Bars show 12-model mean error rates; radar panels illustrate model-specific signatures on the same axes.}
\label{fig:stage_error_landscape}
\vspace{-3mm}
\end{figure*}

Table~\ref{tab:main_results} shows that no model leads the full VEHBench workflow. The four stages have four different winners: qwen3-max for P1 triage, gemini-3.1-pro-preview for P2 search, hunyuan-hy3-preview for P3 recovery, and gpt-5.4 for P4 selection. Stage leaderboards are weak proxies for one another: pairwise Spearman correlations range from \(-0.26\) to \(0.38\) (Appendix Table~\ref{tab:appendix_stage_rank_corr}). The immediate question is therefore not only which model wins, but how each stage fails. Figure~\ref{fig:stage_error_landscape} answers this by grouping stage-specific, non-exclusive diagnostic error rates across the 12-model roster. The radar panels are illustrative rather than additional headline metrics: they show that the same stage-level error axes produce different model-specific failure signatures.

P1 is an entry-control problem. We diagnose triage failures with over-action, over-refusal, and missing-information miss, rather than raw accuracy, because proposing can look superficially acceptable while failing safety-critical action cases (Appendix~\ref{sec:appendix_fulltable_metric_dictionary}; Table~\ref{tab:appendix_calibration}). Across the 12-model roster, mean over-action is 0.683, mean missing-information miss is 0.635, and mean over-refusal is only 0.085 (Appendix Table~\ref{tab:appendix_stage_error_decomp}). The dominant P1 error is therefore not excessive caution; it is entering design when the brief is infeasible or underspecified. qwen3-max has the best P1 composite because it suppresses unsafe entry and missing-information misses relative to most models (Appendix Table~\ref{tab:appendix_p1_full}).

P2 is a physical search-closure problem. We separate infeasible closure, destructive or invalid updates, and utility loss so that well-formed edits alone are not rewarded. Mean infeasible closure is 0.478 and mean utility loss is 0.644, while destructive and protocol-invalid proxies are near zero (Appendix Table~\ref{tab:appendix_stage_error_decomp}). Most models can follow the output protocol but still fail to close the physical search loop; gemini-3.1-pro-preview is distinctive because its closure failure is only 0.038, even though useful-power loss remains substantial (Appendix Table~\ref{tab:appendix_p2_full}).

P3 deserves a separate readout because its failures occur after exposure to a corrupted trajectory. Its diagnostics separate escape, cascade, dead budget, and post-escape recovery. Mean escape failure is only 0.191, but mean post-escape failure is 0.609 (Appendix Table~\ref{tab:appendix_stage_error_decomp}). The bottleneck is therefore not mainly noticing the trap; it is stabilizing after escape. hunyuan-hy3-preview wins P3 because its post-escape failure is much lower than the roster average (Appendix Table~\ref{tab:appendix_p3_full}). This post-escape gap bridges to the profile and state-interface analyses below.

P4 has a different surface again. Mean policy mismatch is 0.709, while dominance error is 0.008 and parse failure is 0.001 (Appendix Table~\ref{tab:appendix_stage_error_decomp}). P4 failures are therefore not mainly formatting, infeasible-candidate, or Pareto-dominance errors; models usually parse and compare feasible candidates, but fail to execute the stated engineering policy (Appendix Table~\ref{tab:appendix_p4_full}). These four surface failures motivate the response-control diagnosis in Finding 2.

\subsection{Finding 2: Profiles Diagnose Stage-Specific Failure Modes}

Response-control profiles add a diagnostic layer to the surface errors in Finding 1: they map failures to behavioral signals in the logs. They are not independent causal variables. Several profile indicators are drawn from the same metric families as the stages they diagnose, so strong intended diagonal correlations should be read as instrumentation checks rather than causal discovery. The useful signal is the full sign pattern, the off-diagonal transfer or conflict, and cases where profile effort decouples from task outcome. The complete scaling rule, indicator sets, and 12-model profile table are in Appendix~\ref{sec:metric_formulas}, Appendix Table~\ref{tab:appendix_profile_indicator_sets}, and Appendix Table~\ref{tab:profile_quantification}.

\begin{table*}[t]
\centering
\scriptsize
\setlength{\tabcolsep}{3pt}
\renewcommand{\arraystretch}{1.08}
\caption{Profile--stage Spearman correlations, computed from Appendix Tables~\ref{tab:profile_quantification} and~\ref{tab:appendix_p1_full}--\ref{tab:appendix_p4_full}.}
\label{tab:profile_stage_corr}
\begin{tabularx}{\textwidth}{p{27mm}rrrrY}
\toprule
Profile axis & P1 Comp. & P2 ratio & P3 Succ. & P4 Tau & Diagnostic tension \\
\midrule
Action discipline & 0.972 & 0.273 & 0.396 & -0.119 & Safe gating can suppress exploratory design updates \\
Edit style & 0.158 & 0.839 & -0.189 & -0.077 & Local edits can follow corrupted trajectory history \\
Feedback conditioning & 0.543 & 0.741 & 0.587 & 0.217 & Feedback guides search; stale context can mislead \\
State-reset effort & -0.074 & -0.392 & 0.217 & 0.119 & Reset helps recovery but disrupts search continuity \\
Policy execution & -0.133 & 0.552 & 0.217 & 0.762 & Policy discipline can constrain design exploration \\
\bottomrule
\end{tabularx}
\end{table*}

Table~\ref{tab:profile_stage_corr} separates three patterns. First, the intended diagonals check that the profiles summarize the stages they diagnose: action discipline tracks P1, edit style tracks P2, and policy execution tracks P4. Second, some priors transfer across boundaries. Feedback conditioning is positive for P1, P2, and P3, making it the closest profile dimension to a general engineering-assistance signal; policy execution also transfers to P2, suggesting that fixed-candidate evaluation and verifier-guided search both reward disciplined use of external constraints. Third, some priors conflict across boundaries. State-reset effort is negatively associated with P2, and edit style is weakly negative with P3: behaviors that help clean search are not automatically useful when the trajectory state is corrupted.

The P1 profile explains why accuracy is insufficient for entry control. qwen3-max has the best P1 composite (0.574) despite lower raw accuracy than \texttt{o4-mini} (65.4\% vs. 70.0\%) because its action discipline is more balanced across propose, missing-information, and infeasibility cases (Appendix Tables~\ref{tab:appendix_p1_full} and~\ref{tab:profile_quantification}). claude-4.6-sonnet has 57.5\% raw accuracy but a much weaker P1 composite (0.207), with ACS 0.057 and MDS 0.000 (Appendix Table~\ref{tab:appendix_p1_full}). Action discipline diagnoses the difference between being often correct and being safe at the gate.

The P2 profile diagnoses closure as directed search, not formatting. gemini-3.1-pro-preview combines top edit style (0.753) with the highest feedback conditioning (0.508), and its closure failure is only 0.038 (Appendix Tables~\ref{tab:profile_quantification} and~\ref{tab:appendix_stage_error_decomp}). \texttt{llama-3.3-70b} has moderate edit style (0.586) but the lowest feedback conditioning (0.326), and its closure failure rises to 0.630 (Appendix Tables~\ref{tab:profile_quantification} and~\ref{tab:appendix_stage_error_decomp}). Both models edit; the difference is whether edits follow verifier feedback.

The P4 profile diagnoses policy mismatch rather than parser weakness. gpt-5.4 has the best P4 Tau (0.887) and high policy execution (0.743), while \texttt{qwen3.6-plus} has the strongest policy execution profile score (0.761; Appendix Table~\ref{tab:profile_quantification}) and best exact-match rate (Appendix Table~\ref{tab:appendix_p4_full}). In contrast, \texttt{llama-3.3-70b} has the lowest policy execution score (0.622) and weakest P4 Tau (0.714) (Appendix Tables~\ref{tab:profile_quantification} and~\ref{tab:appendix_p4_full}). P4 measures evaluator-role discipline after search has been removed.

The clearest decoupling case is P3. Table~\ref{tab:p3_state_decoupling} shows why the state-reset profile should be read as recovery effort, not recovery success; its source columns are expanded in Appendix Tables~\ref{tab:profile_quantification} and~\ref{tab:appendix_p3_full}. deepseek-v4-pro has the highest state-reset score and replans in 99.4\% of tasks, but its P3 success is only 28.8\% and its recovery quality is 0.199. In contrast, hunyuan-hy3-preview and mimo-v2.5-pro replan much less aggressively but reach feasible states and preserve objective quality more often. P3 therefore requires both recognizing contaminated state and stabilizing after reset.

\begin{table*}[t]
\centering
\scriptsize
\setlength{\tabcolsep}{3.5pt}
\renewcommand{\arraystretch}{1.08}
\caption{P3 state-reset effort versus recovery outcome.}
\label{tab:p3_state_decoupling}
\begin{tabular}{lrrrrrr}
\toprule
Model & State reset & P3 Succ. & Escape & Replan & First feas. & RecQ \\
\midrule
deepseek-v4-pro & 0.737 & 28.8\% & 90.4\% & 99.4\% & 27.6\% & 0.199 \\
qwen3-max & 0.646 & 30.1\% & 85.9\% & 49.4\% & 22.4\% & 0.261 \\
deepseek-v3 & 0.629 & 35.3\% & 82.7\% & 57.7\% & 23.1\% & 0.287 \\
hunyuan-hy3-preview & 0.599 & 47.4\% & 77.6\% & 7.7\% & 40.4\% & 0.488 \\
mimo-v2.5-pro & 0.627 & 45.5\% & 63.5\% & 17.9\% & 41.0\% & 0.626 \\
\midrule
Model mean & 0.540 & 31.9\% & 80.9\% & 23.7\% & 21.5\% & 0.275 \\
\bottomrule
\end{tabular}
\end{table*}

 Model-level profiles then identify usable failure modes rather than a global best model. gemini-3.1-pro-preview is interaction-oriented, matching its P2 lead. gpt-5.4 is evaluator-oriented, with the highest policy execution (0.743) and P4 Tau (0.887), but weaker edit style (0.482) and P2 ratio (0.133) (Appendix Tables~\ref{tab:profile_quantification}, \ref{tab:appendix_p2_full}, and~\ref{tab:appendix_p4_full}). These patterns expose two actionable failure classes: representation/interface failures, where recovery effort does not stabilize state, and boundary-fit failures, where a useful prior at one stage does not transfer to another.

\subsection{Finding 3: Diagnostics Translate into Engineering Controls}

The two failure classes above point to two controls. For representation failure, we change the P3 state interface. For boundary-fit tension, we analyze stage-aware model selection.

\begin{figure*}[!t]
\centering
\includegraphics[width=\linewidth,trim=0 0 0 0,clip]{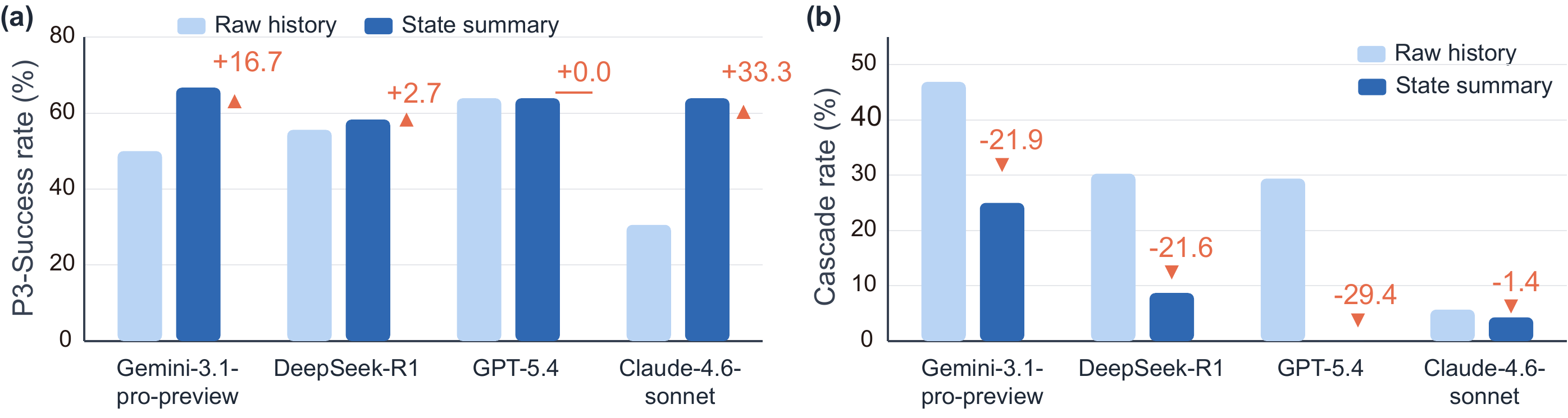}
\vspace{-5mm}
\caption{P3 state-summary intervention. It improves recovery for most models and reduces cascade at the corrupted-state boundary; full numeric results are in Appendix Table~\ref{tab:appendix_p3_intervention_delta_ci}.}
\label{fig:mechanism_evidence}
\vspace{1.5mm}
\end{figure*}

\paragraph{Control 1: State interface design.}
The P3 post-escape gap suggests a representation failure: the model may try to recover, but raw corrupted history remains in context and prevents stabilization. We therefore replace raw trajectory history with a verifier-authored state summary that preserves the current trusted proposal, best feasible candidate, and active violations while removing the contaminated conversational trace. On a stratified 36-task subset, this state-interface change improves model-mean recovery from 50.0\% to 63.2\% and lowers cascade from 28.1\% to 9.5\% (Figure~\ref{fig:mechanism_evidence}). The response is profile-dependent: claude-4.6-sonnet gains 33.3 pts in success, suggesting that raw-history format was suppressing usable design behavior; gemini-3.1-pro-preview gains 16.7 pts, consistent with its strong edit-style profile benefiting from a cleaner state; gpt-5.4 holds success fixed but reduces cascade to 0.0\%; deepseek-r1 gains only 2.8 pts. The mechanism is therefore not ``longer prompting.'' It is stage-local state control.

\paragraph{Control 2: Stage-aware selection simulation.}
Finding 2's profile-stage matrix shows why routing is needed: state-reset effort is negatively associated with P2, and edit style is weakly negative with P3. Table~\ref{tab:stage_aware_router} turns this into a deployment view using full 12-model held-out splits. Stage specialists win locally but pay a cross-stage tax: qwen3-max, hunyuan-hy3-preview, and gpt-5.4 drop to mean normalized scores of 0.738, 0.755, and 0.745 when forced across the workflow. gemini-3.1-pro-preview is the best single-model fallback not because it dominates, but because its weakness is less severe, giving a mean of 0.892. The insight is that aggregate deployment quality depends less on peak strength than on avoiding boundary-specific collapse.

\begin{table*}[t]
\caption{Stage-aware selection on full 12-model held-out splits. Stage cells report raw held-out scores; Mean norm. averages scores after dividing each stage by its held-out stage-best score.}
\label{tab:stage_aware_router}
\centering
\scriptsize
\setlength{\tabcolsep}{2.7pt}
\renewcommand{\arraystretch}{1.04}
\begin{tabularx}{\textwidth}{p{44mm}YYYYr}
\toprule
Strategy & P1 & P2 & P3 & P4 & Mean norm. \\
\midrule
P1 specialist: qwen3-max & 0.594 & 0.185 & 23.9\% & 0.839 & 0.738 \\
P2 specialist: gemini-3.1-pro-preview & 0.552 & 0.385 & 32.6\% & 0.821 & 0.892 \\
P3 specialist: hunyuan-hy3-preview & 0.413 & 0.157 & 45.7\% & 0.815 & 0.755 \\
P4 specialist: gpt-5.4 & 0.454 & 0.148 & 38.0\% & 0.888 & 0.745 \\
\textbf{Validation-selected router} & 0.552 (gemini-3.1-pro-preview) & 0.385 (gemini-3.1-pro-preview) & 42.4\% (mimo-v2.5-pro) & 0.821 (deepseek-v3) & 0.945 \\
\midrule
\textbf{Ex-post stage-best envelope} & 0.594 (qwen3-max) & 0.385 (gemini-3.1-pro-preview) & 45.7\% (hunyuan-hy3-preview) & 0.888 (gpt-5.4) & 1.000 \\
\bottomrule
\end{tabularx}
\end{table*}

The split-based routing simulation in Appendix~\ref{sec:appendix_stage_router} separates realized routing gain from remaining headroom. The validation-selected router improves mean normalized held-out score from 0.892 to 0.945, almost entirely by switching P3 from gemini-3.1-pro-preview to mimo-v2.5-pro and raising normalized recovery from 0.714 to 0.929. The ex-post envelope remains 1.000, leaving 0.055 headroom from validation misses of held-out stage leaders. Thus VEHBench supports two concrete controls: state-interface design for representation failures and stage-aware routing for profile-boundary mismatch.

\subsection{Robustness and Scope Checks}

We run five controls. Three address internal threats: a selection--generation audit confirms ranking is not reducible to generation (26.1--60.9 pt A--B gaps); a query-matched CMA-ES baseline rules out P2 as pure numerical search (ratio 0.099 vs. the strongest LLM); and a prompt-control audit finds no consistent correction of known deficits. Two support construct validity: thinking-mode coverage helps recovery but does not predict the four stage leaders; the circuit pilot reproduces stage spreads under a second engineering domain. Full results are reported in Appendix~\ref{sec:appendix_isomorphic}--\ref{sec:appendix_circuit_pilot}.

\section{Discussion}

The empirical findings suggest that engineering LLMs should be evaluated and deployed through stage-specific use of external verification, not as monolithic designers. VEHBench is complementary to simulation-backed evaluation: verifiers determine physical validity, while diagnostic probes show whether the model used that validity signal appropriately at each design stage.

History is a control variable, not just context. In clean search, raw trajectory history can carry useful constraint information; in corrupted recovery, the same history can become contamination. The state-summary intervention supports stage-specific state representation: preserve trajectory during clean search, sanitize state during recovery, and expose feasible candidates plus policy during selection.

The broader contribution is the diagnostic chain, not any single model ranking. Surface error decomposition identifies what fails, response-control profiles diagnose the behavior behind failure, and stage-local controls test whether interface or routing changes can improve outcomes. These profiles are operational diagnostics, not causal claims about cognition. VEHBench remains scoped to analytically verifiable cantilever VEH co-design and does not replace FEM or hardware certification; the next step is to instantiate the same scaffold with richer simulators and other coupled domains.

\section{Conclusion}
\label{sec:conclusion}

We introduced VEHBench, an engineering-native diagnostic benchmark for LLM-assisted VEH co-design. It combines literature-grounded task construction, analytical verification, and stage-local evaluation of design behavior. The central conclusion is that LLM capability in engineering design is role-dependent: aggregate rank is less informative than stage compatibility. VEHBench provides both an evaluation tool and a design guide for verifier-grounded engineering agents: select, route, adapt, and identify improvement targets according to the stage where they are used.

\clearpage
\bibliographystyle{plainnat}
\nocite{akhtar2024croissant,alibaba2026qwenDeepThinking,citroni2024efficient,deepseek2025r1,deepseek2026v4preview,doris2024designqa,dorner2025limits,efron1993introduction,erturk2008bimorph,erturk2009experimental,gebru2021datasheets,google2026geminithinking,guo2025engdesign,hansen2021coco,hansen2023cma,jadhav2024mechanical,jain2024mseval,kiela2021dynabench,liang2023helm,maamer2019review,mitchell2019modelcards,naifar2024energy,openai2026o4mini,pu2025overbench,qwen2026qwen36plus,raji2021aiaccountability,sadaf2024cantilever,sadaf2025harnessing,safaei2019review,siliconflow2026hy3preview,srivastava2022bigbench,wang2006nanogenerators,williams1996microelectric,xia2025buildarena,xiong2025stealtheval,xu2010nanowire,zeadally2020design}
\bibliography{main}

\clearpage
\appendix
\setcounter{equation}{0}
\renewcommand{\theequation}{A-\arabic{equation}}
\newpage

\section*{\hspace{-4mm} \centering Appendix}
\vspace{3mm}

\section{Supplementary Roadmap}

The appendix supports the benchmark claims in the main paper rather than serving as a storage room for extra leaderboards. It is organized into four support blocks:
\begin{itemize}[leftmargin=5mm,itemsep=1mm,topsep=1mm]
    \item \textbf{Dataset and artifact audit:} construction provenance, artifact boundary, contamination scope, P3 intervention sampling, and release contents.
    \item \textbf{Physical oracle, metric, and statistical definitions:} closed-form VEH equations, uncertainty intervals, calibration references, and formal definitions for every headline and diagnostic metric.
    \item \textbf{Full results and split checks:} complete P1--P4 full-bank tables, split-resolved tables, reasoning/thinking coverage, response-control profiles, and stage-aware selection/error decomposition.
    \item \textbf{Mechanism and cross-domain audits:} extended related work, prompts, failure cases, selection--generation isomorphism, Tier~3 controlled-prompt results, CMA-ES, and the circuit audit.
\end{itemize}

\subsection{Master Metric Index}
\label{sec:master_metric_index}

Tables~\ref{tab:master_metric_index_p1p4}--\ref{tab:master_metric_index_controls} provide a compact index of the metrics used in Section~\ref{sec:experiments} and the main appendix result tables. The index is a lookup layer: the Definition column points to the appendix section or table that gives the formula, scaling rule, or source calculation.

{\scriptsize
\setlength{\tabcolsep}{1.5pt}
\renewcommand{\arraystretch}{1.08}
\begin{longtable}{@{}p{20mm}p{9mm}p{13mm}p{21mm}p{49mm}p{20mm}@{}}
\caption{P1--P4 metrics used in main results.}
\label{tab:master_metric_index_p1p4}\\
\toprule
Metric & Probe & Type & Used in & Formula sketch / meaning & Definition \\
\midrule
\endfirsthead
\toprule
Metric & Probe & Type & Used in & Formula sketch / meaning & Definition \\
\midrule
\endhead
\midrule
\multicolumn{6}{r}{Continued on next page} \\
\endfoot
\bottomrule
\endlastfoot
\multicolumn{6}{@{}l}{\textbf{P1: specification triage}} \\
P1-Composite & P1 & Headline & Table~\ref{tab:main_results}; Finding 1 & Weighted score combining Macro-F1, ACS, MDS, IDS, and subtype F1; higher means safer entry control. & App.~\ref{sec:metric_p1} \\
Accuracy & P1 & Diagnostic & Finding 2; App. tables & Fraction of P1 tasks whose predicted action equals the gold triage action. & App.~\ref{sec:metric_p1} \\
Macro-F1 & P1 & Diagnostic & App. tables & Unweighted mean of F1 over propose, infeasible, and request actions. & App.~\ref{sec:metric_p1} \\
ACS & P1 & Diagnostic & App. tables & Propose recall multiplied by one minus spurious-propose rate. & App.~\ref{sec:metric_p1} \\
MDS & P1 & Diagnostic & Finding 2; App. tables & Missing-info request recall multiplied by one minus spurious-request rate. & App.~\ref{sec:metric_p1} \\
IDS & P1 & Diagnostic & App. tables & Infeasible recall multiplied by one minus spurious-infeasible rate. & App.~\ref{sec:metric_p1} \\
Subtype F1 & P1 & Diagnostic & App. tables & Macro-F1 over six subtype families, used as a small component of P1-Composite. & App.~\ref{sec:metric_p1} \\
Over-action & P1 & Error family & Fig.~\ref{fig:stage_error_landscape}; Finding 1 & Rate of proposing design when the gold action is not propose. & App.~\ref{sec:metric_p1}; App.~\ref{sec:metric_diagnostic_dictionary} \\
Over-refusal & P1 & Error family & Fig.~\ref{fig:stage_error_landscape}; Finding 1 & Spurious request plus spurious infeasible rates on tasks whose gold action is different. & App.~\ref{sec:metric_diagnostic_dictionary} \\
Missing-info miss & P1 & Error family & Fig.~\ref{fig:stage_error_landscape}; Finding 1 & \(1-\) missing-info recall; denominator is gold missing-info cases. & App.~\ref{sec:metric_diagnostic_dictionary} \\
\midrule
\multicolumn{6}{@{}l}{\textbf{P2: verifier-guided design search}} \\
P2 ratio / P2b & P2 & Headline & Table~\ref{tab:main_results}; Finding 1 & Mean \(F_iP_i/P_i^{\mathrm{BKF}}\); infeasible final designs contribute zero. & App.~\ref{sec:metric_p2}; App.~\ref{sec:metric_diagnostic_dictionary} \\
Final feasible rate & P2 & Diagnostic & Finding 2; App. tables & Fraction of P2 tasks whose final candidate satisfies all oracle constraints. & App.~\ref{sec:metric_p2} \\
CondRatio & P2 & Diagnostic & App. tables & Mean objective ratio only among tasks that close feasibly. & App.~\ref{sec:metric_diagnostic_dictionary} \\
Utility loss & P2 & Error family & Fig.~\ref{fig:stage_error_landscape}; Finding 1 & \(1-\) CondRatio, clipped to \([0,1]\); objective shortfall after feasible closure. & App.~\ref{sec:metric_diagnostic_dictionary} \\
Infeasible closure & P2 & Error family & Fig.~\ref{fig:stage_error_landscape}; Finding 1 & \(1-\) final feasible rate; failure to close the physical search loop. & App.~\ref{sec:metric_diagnostic_dictionary} \\
Destructive edit & P2 & Error family & Fig.~\ref{fig:stage_error_landscape} & Feasible-to-infeasible transition rate after an edit. & App.~\ref{sec:metric_diagnostic_dictionary} \\
Invalid/no-op proxy & P2 & Error family & Fig.~\ref{fig:stage_error_landscape} & Protocol-invalid rate used as the stable released proxy for invalid or no-op behavior. & App.~\ref{sec:appendix_stage_error_decomposition} \\
P2a first feasible & P2 & Diagnostic & App. tables & Fraction feasible after the first update step. & App.~\ref{sec:metric_p2} \\
Improvement rate / P2c & P2 & Diagnostic & App. tables & Fraction of consecutive proposals with higher normalized utility. & App.~\ref{sec:metric_diagnostic_dictionary} \\
AUC & P2 & Diagnostic & App. tables & Mean best-so-far normalized objective over the fixed query budget. & App.~\ref{sec:metric_diagnostic_dictionary} \\
Queries & P2 & Diagnostic & App. tables & Mean number of oracle calls used by the model trajectory. & App.~\ref{sec:metric_diagnostic_dictionary} \\
\midrule
\multicolumn{6}{@{}l}{\textbf{P3: corrupted-state recovery}} \\
P3-Success & P3 & Headline & Table~\ref{tab:main_results}; Finding 1 & Final feasibility after the corrupted-state recovery budget. & App.~\ref{sec:metric_p3} \\
Recovery quality / RecQ & P3 & Diagnostic & Table~\ref{tab:p3_state_decoupling}; App. tables & Objective ratio among successful final recoveries. & App.~\ref{sec:metric_p3}; App.~\ref{sec:metric_diagnostic_dictionary} \\
First feasible & P3 & Diagnostic & App. tables & Whether the trajectory reaches feasibility before the final step. & App.~\ref{sec:metric_diagnostic_dictionary} \\
Escape rate & P3 & Diagnostic & Finding 1; App. tables & Fraction of tasks where the model moves away from the known corrupted trap. & App.~\ref{sec:metric_p3}; App.~\ref{sec:metric_diagnostic_dictionary} \\
Cascade rate & P3 & Diagnostic & Fig.~\ref{fig:mechanism_evidence}; App. tables & Fraction of escaped tasks that introduce a new coupled violation. & App.~\ref{sec:metric_p3}; App.~\ref{sec:metric_diagnostic_dictionary} \\
Dead-budget rate & P3 & Diagnostic & Fig.~\ref{fig:stage_error_landscape}; App. tables & Fraction of tasks with no meaningful recovery action. & App.~\ref{sec:metric_p3}; App.~\ref{sec:metric_diagnostic_dictionary} \\
Replan rate & P3 & Diagnostic & Table~\ref{tab:p3_state_decoupling}; App. tables & Fraction of tasks with explicit reset or replanning behavior. & App.~\ref{sec:metric_diagnostic_dictionary} \\
Escape failure & P3 & Error family & Fig.~\ref{fig:stage_error_landscape}; Finding 1 & \(1-\) escape rate. & App.~\ref{sec:metric_diagnostic_dictionary} \\
Post-escape failure & P3 & Error family & Fig.~\ref{fig:stage_error_landscape}; Findings 1--3 & \(1-\mathrm{Success}/\max(\mathrm{Escape},\epsilon)\), clipped to \([0,1]\). & App.~\ref{sec:metric_diagnostic_dictionary} \\
\midrule
\multicolumn{6}{@{}l}{\textbf{P4: policy-conditioned selection}} \\
Full Tau & P4 & Headline & Table~\ref{tab:main_results}; Finding 1 & Kendall \(\tau_b\) between model ranking and oracle policy ranking over the full pool. & App.~\ref{sec:metric_p4}; App.~\ref{sec:metric_diagnostic_dictionary} \\
Exact match & P4 & Diagnostic & App. tables & Full-order exact match to the oracle policy order. & App.~\ref{sec:metric_p4}; App.~\ref{sec:metric_diagnostic_dictionary} \\
Top-1 accuracy & P4 & Diagnostic & App. tables & Whether the oracle-best candidate is ranked first. & App.~\ref{sec:metric_p4}; App.~\ref{sec:metric_diagnostic_dictionary} \\
Top-2 set accuracy & P4 & Diagnostic & App. tables & Whether the unordered top-two candidate set is preserved. & App.~\ref{sec:metric_p4}; App.~\ref{sec:metric_diagnostic_dictionary} \\
Pareto Tau & P4 & Diagnostic & App. tables & Ranking consistency on Pareto-relevant comparisons. & App.~\ref{sec:metric_diagnostic_dictionary} \\
BARS & P4 & Diagnostic & App. tables & Balanced-active ranking score combining tau, policy-sensitive pairs, and exact match. & App.~\ref{sec:metric_p4}; App.~\ref{sec:metric_diagnostic_dictionary} \\
Policy mismatch & P4 & Error family & Fig.~\ref{fig:stage_error_landscape}; Finding 1 & \(1-\) all-policy-sensitive pair accuracy. & App.~\ref{sec:metric_diagnostic_dictionary} \\
Dominance error & P4 & Error family & Fig.~\ref{fig:stage_error_landscape} & Pareto-violation rate. & App.~\ref{sec:metric_diagnostic_dictionary} \\
Parse failure & P4 & Error family & Fig.~\ref{fig:stage_error_landscape} & Parse-error rate in the ranking output. & App.~\ref{sec:metric_diagnostic_dictionary} \\
Top-choice miss & P4 & Error family & App. error tables & \(1-\) top-1 accuracy. & App.~\ref{sec:metric_diagnostic_dictionary} \\
\end{longtable}
}

{\scriptsize
\setlength{\tabcolsep}{1.5pt}
\renewcommand{\arraystretch}{1.08}
\begin{longtable}{@{}p{24mm}p{14mm}p{20mm}p{52mm}p{24mm}@{}}
\caption{Profile, intervention, routing, and robustness metrics.}
\label{tab:master_metric_index_controls}\\
\toprule
Metric & Analysis & Used in & Formula sketch / meaning & Definition \\
\midrule
\endfirsthead
\toprule
Metric & Analysis & Used in & Formula sketch / meaning & Definition \\
\midrule
\endhead
\bottomrule
\endlastfoot
Action discipline & Profile & Table~\ref{tab:profile_stage_corr}; Finding 2 & Direction-normalized average of P1 action alignment, macro-F1, recalls, propose precision, and non-invalid rate. & App.~\ref{sec:metric_profiles}; Table~\ref{tab:appendix_profile_indicator_sets} \\
Edit style & Profile & Table~\ref{tab:profile_stage_corr}; Finding 2 & Direction-normalized average of bounded local edits, feasibility preservation, directed updates, final feasibility, non-destructive edits, and protocol validity. & App.~\ref{sec:metric_profiles}; Table~\ref{tab:appendix_profile_indicator_sets} \\
Feedback conditioning & Profile & Table~\ref{tab:profile_stage_corr}; Finding 2 & Direction-normalized average of violation reduction, utility improvement, best-so-far AUC, and post-feedback feasibility signals. & App.~\ref{sec:metric_profiles}; Table~\ref{tab:appendix_profile_indicator_sets} \\
State-reset effort & Profile & Tables~\ref{tab:profile_stage_corr}, \ref{tab:p3_state_decoupling} & Direction-normalized average of escape, replan, escape quality, non-cascade, non-dead-budget, and state-summary gains. & App.~\ref{sec:metric_profiles}; Table~\ref{tab:appendix_profile_indicator_sets} \\
Policy execution & Profile & Table~\ref{tab:profile_stage_corr}; Finding 2 & Direction-normalized average of P4 tau/BARS, policy-sensitive pair accuracy, exact/top-1 accuracy, and non-error terms. & App.~\ref{sec:metric_profiles}; Table~\ref{tab:appendix_profile_indicator_sets} \\
\midrule
State-summary \(\Delta\) success & Intervention & Fig.~\ref{fig:mechanism_evidence}; Finding 3 & Paired task-level P3-Success difference between state-summary and raw-history prompts. & Table~\ref{tab:appendix_p3_intervention_delta_ci} \\
State-summary \(\Delta\) cascade & Intervention & Fig.~\ref{fig:mechanism_evidence}; Finding 3 & Paired task-level cascade-rate difference between state-summary and raw-history prompts. & Table~\ref{tab:appendix_p3_intervention_delta_ci} \\
Intervention subset & Intervention & Finding 3; App. sampling & Stratified 36-task P3 subset balanced across splits and subtype families. & Table~\ref{tab:appendix_p3_subset_provenance} \\
\midrule
Mean normalized score & Routing & Table~\ref{tab:stage_aware_router}; Finding 3 & Average over stages after dividing each held-out stage score by its held-out stage-best score. & App.~\ref{sec:appendix_stage_router} \\
Validation-selected router & Routing & Table~\ref{tab:stage_aware_router}; Finding 3 & Per-stage model chosen on validation split and evaluated only on held-out splits. & App.~\ref{sec:appendix_stage_router} \\
Ex-post stage-best envelope & Routing & Table~\ref{tab:stage_aware_router}; Finding 3 & Ex-post held-out best model per stage; upper bound, not deployable selection. & App.~\ref{sec:appendix_stage_router} \\
Router gain & Routing & Finding 3 & Mean normalized score of validation router minus best single-model fallback. & App.~\ref{sec:appendix_stage_router} \\
Remaining headroom & Routing & Finding 3 & \(1-\) validation-router mean normalized score. & App.~\ref{sec:appendix_stage_router} \\
\midrule
Selection-generation gap & Robustness & Sec.~4.5 & A-selection success minus B-generation success on isomorphic probe groups. & App.~\ref{sec:appendix_isomorphic} \\
CMA-ES P2 ratio & Robustness & Sec.~4.5 & Query-matched optimizer final feasible objective ratio under the same oracle. & App.~\ref{sec:appendix_cmaes} \\
Controlled-prompt delta & Robustness & Sec.~4.5 & Targeted prompt condition minus neutral/default condition on pre-registered weak cells. & App.~\ref{sec:tier3_full} \\
Thinking-mode coverage & Robustness & Sec.~4.5; App. tables & Grouped stage means by whether thinking mode was used in the reported run. & Tables~\ref{tab:appendix_reasoning_grouping}--\ref{tab:appendix_reasoning_groupmeans} \\
Circuit audit spread & Robustness & Sec.~4.5 & Stage-leader spread reproduced under a second closed-form engineering domain. & App.~\ref{sec:appendix_circuit_pilot} \\
\end{longtable}
}

\paragraph{Reporting conventions.}
All headline metrics are higher-is-better. Error-family rates are lower-is-better and non-exclusive: they can use different denominators and are not intended to sum to one. Unless stated otherwise, roster means average per-model rates over the 12 complete P1--P4 runs. Held-out routing combines \texttt{test\_id} and \texttt{test\_ood} by task-count weights; validation splits are used only for model selection in the router simulation.

\section{VEH Analytical Oracle Formulas}
\label{sec:appendix_veh_oracle_formulas}

This section gives the closed-form equations used by the analytical VEH oracle. The oracle is a fast single-mode Euler--Bernoulli model for a unimorph piezoelectric cantilever with one substrate layer, one piezoelectric layer, a possible tip mass, harmonic base excitation, and a resistive electrical load. It is the external verifier used to compute physical feasibility and objective values; it is not a FEM model or a hardware-certification model. The electromechanical structure follows cantilevered piezoelectric energy-harvester models~\citep{erturk2008bimorph,erturk2009experimental}; broader self-powered nanogenerator context is cited separately in the appendix discussion~\citep{wang2006nanogenerators,xu2010nanowire}. The implementation follows the same order as \texttt{diagbench.physics.oracle.PiezoelectricOracle}.

\begin{center}
\captionof{table}{Symbols used by the VEH analytical oracle. All internal computations use SI units; task files expose lengths in mm, thicknesses in \(\mu\)m, tip mass in g, excitation in Hz and g, and output power in \(\mu\)W.}
\label{tab:appendix_veh_symbols}
\centering
\scriptsize
\setlength{\tabcolsep}{3pt}
\renewcommand{\arraystretch}{1.08}
\begin{tabularx}{\textwidth}{p{17mm}p{25mm}p{18mm}Y}
\toprule
Symbol & Meaning & Unit & Notes \\
\midrule
\(L,b\) & beam length and width & m & from \texttt{beam\_length\_mm}, \texttt{beam\_width\_mm} \\
\(h_s,h_p\) & substrate and piezo layer thickness & m & from \texttt{substrate\_thickness\_um}, \texttt{piezo\_thickness\_um} \\
\(m_t,R_L\) & tip mass and load resistance & kg, \(\Omega\) & from \texttt{tip\_mass\_g}, \texttt{load\_resistance\_ohm} \\
\(f_e,\omega\) & excitation frequency and angular frequency & Hz, rad/s & \(\omega=2\pi f_e\) \\
\(a_g,a\) & base acceleration amplitude & g, m/s\(^2\) & \(a=9.80665\,a_g\) \\
\(E_s,E_p\) & substrate and piezo Young's moduli & Pa & material-table values \\
\(\rho_s,\rho_p\) & substrate and piezo densities & kg/m\(^3\) & material-table values \\
\(d_{31},e_{31}\) & piezo strain and stress constants & m/V, C/m\(^2\) & \(e_{31}=d_{31}E_p\) \\
\(\epsilon_{33}^{T}\) & piezo permittivity at constant stress & F/m & \(\epsilon_{33}^{T}=\epsilon_0\epsilon_{33,r}\), \(\epsilon_0=8.854187817{\times}10^{-12}\) F/m \\
\(\zeta\) & mechanical damping ratio & -- & default \(0.01\) unless specified in the task environment \\
\(\bar y\) & transformed neutral-axis location from substrate bottom & m & modulus-weighted centroid \\
\(EI\) & composite bending stiffness & N m\(^2\) & parallel-axis theorem over substrate and piezo layers \\
\(m_{\mathrm{eff}}\) & first-mode effective mass & kg & \(m_{\mathrm{eff}}=0.2357m_b+m_t\) \\
\(\omega_r,f_r\) & resonant angular frequency and frequency & rad/s, Hz & first-mode clamped-free approximation \\
\(\theta\) & electromechanical coupling coefficient & N/V = C/m & single-mode unimorph coupling \\
\(C_p,Z_e\) & internal capacitance and electrical load impedance & F, \(\Omega\) & \(Z_e=R_L/(1+j\omega R_LC_p)\) \\
\(W,V\) & tip displacement and load voltage amplitudes & m, V & harmonic steady-state amplitudes \\
\(P,\sigma,\eta_f\) & average load power, root stress, frequency error & W, Pa, \% & task outputs are \(\mu\)W, MPa, and \% \\
\bottomrule
\end{tabularx}
\end{center}

\paragraph{Input conversion and material parameters.}
The task design vector is
\begin{equation}
\mathbf{x}=(L_{\mathrm{mm}},\,b_{\mathrm{mm}},\,h_{s,\mu\mathrm{m}},\,h_{p,\mu\mathrm{m}},\,m_{t,\mathrm{g}},\,R_L),
\end{equation}
and the excitation context is \((f_e,a_g)\). The oracle converts to SI units by
\begin{equation}
\begin{aligned}
L&=10^{-3}L_{\mathrm{mm}}, &
 b&=10^{-3}b_{\mathrm{mm}}, &
 h_s&=10^{-6}h_{s,\mu\mathrm{m}}, \\
 h_p&=10^{-6}h_{p,\mu\mathrm{m}}, &
 m_t&=10^{-3}m_{t,\mathrm{g}}, &
 a&=9.80665\,a_g .
\end{aligned}
\end{equation}
The piezo material supplies \((d_{31},\epsilon_{33,r},E_p,\rho_p)\) and the substrate material supplies \((E_s,\rho_s)\). Unless specified, the default pair is PZT-5A on stainless steel. The oracle supports PZT-5A, PZT-5H, MFC-M8528, and PVDF for the piezo layer, and stainless steel, aluminum, brass, and titanium for the substrate.

\paragraph{Composite section and bending stiffness.}
Let \(A_s=bh_s\) and \(A_p=bh_p\) be the layer areas. The layer centroids measured upward from the substrate bottom are
\begin{equation}
y_s=\frac{h_s}{2},\qquad y_p=h_s+\frac{h_p}{2}.
\end{equation}
The transformed neutral axis is the modulus-weighted centroid
\begin{equation}
\bar y=\frac{E_sA_sy_s+E_pA_py_p}{E_sA_s+E_pA_p}.
\end{equation}
The second moments of area about the neutral axis are
\begin{equation}
I_s=\frac{bh_s^3}{12}+A_s(y_s-\bar y)^2,\qquad
I_p=\frac{bh_p^3}{12}+A_p(y_p-\bar y)^2,
\end{equation}
and the composite bending stiffness is
\begin{equation}
EI=E_sI_s+E_pI_p .
\end{equation}

\paragraph{Equivalent mass and resonance.}
The distributed beam mass is
\begin{equation}
m_b=(\rho_sh_s+\rho_ph_p)bL.
\end{equation}
The first-mode effective mass uses the standard clamped-free cantilever factor
\begin{equation}
m_{\mathrm{eff}}=\alpha_\phi m_b+m_t,\qquad \alpha_\phi=0.2357 .
\end{equation}
The analytical first-mode resonance is approximated by
\begin{equation}
\omega_r=\sqrt{\frac{3EI}{m_{\mathrm{eff}}L^3}},\qquad
f_r=\frac{\omega_r}{2\pi}.
\end{equation}
The frequency-matching diagnostic used by the constraints is
\begin{equation}
\eta_f=100\frac{|f_r-f_e|}{f_e}.
\end{equation}

\paragraph{Electromechanical coupling and capacitance.}
The piezoelectric stress constant is
\begin{equation}
e_{31}=d_{31}E_p .
\end{equation}
For full piezo coverage and a tip-normalized first bending mode, the root-slope factor is \(\gamma_1=1.3765\). The single-mode electromechanical coupling coefficient is
\begin{equation}
\theta=-e_{31}\,b\,(y_p-\bar y)\frac{\gamma_1}{L}.
\end{equation}
The internal capacitance of the piezo layer is
\begin{equation}
C_p=\epsilon_{33}^{T}\frac{bL}{h_p},\qquad
\epsilon_{33}^{T}=\epsilon_0\epsilon_{33,r}.
\end{equation}

\paragraph{Electrical load and harmonic response.}
For excitation angular frequency \(\omega=2\pi f_e\), the resistive load and piezo capacitance form the complex electrical impedance
\begin{equation}
Z_e(\omega)=\frac{R_L}{1+j\omega R_LC_p}.
\end{equation}
The coupled single-mode denominator is
\begin{equation}
\begin{aligned}
D(\omega)
&=m_{\mathrm{eff}}(\omega_r^2-\omega^2)
+j\,2\zeta m_{\mathrm{eff}}\omega_r\omega
+\theta^2j\omega Z_e(\omega).
\end{aligned}
\end{equation}
The steady-state tip displacement amplitude under base acceleration is
\begin{equation}
W=\frac{m_{\mathrm{eff}}a}{|D(\omega)|}.
\end{equation}
This is the dynamic frequency-response displacement used by the implementation, not the static cantilever deflection \(FL^3/(3EI)\). The voltage amplitude across the load is
\begin{equation}
V=|\theta|\,\omega\,|Z_e(\omega)|\,W,
\end{equation}
and the average load power is
\begin{equation}
P=\frac{V^2}{2R_L},\qquad P_{\mu\mathrm{W}}=10^6P.
\end{equation}
Here \(V\) is a peak harmonic amplitude. Equivalently, \(P=V_{\mathrm{rms}}^2/R_L\) with \(V_{\mathrm{rms}}=V/\sqrt{2}\).
The reported tip displacement is
\begin{equation}
W_{\mathrm{mm}}=10^3W.
\end{equation}

\paragraph{Root stress and feasibility constraints.}
The oracle uses a root-bending approximation for substrate tensile-face stress. With \(c_s=\bar y\) denoting the distance from the neutral axis to the bottom substrate face,
\begin{equation}
\sigma=E_s c_s\frac{3W}{L^2},\qquad
\sigma_{\mathrm{MPa}}=10^{-6}\sigma .
\end{equation}
Given task limits \(\sigma_{\max}\), \(W_{\max}\), \(\eta_{\max}\), and \(P_{\min}\), the oracle reports signed slacks
\begin{equation}
\begin{aligned}
s_{\sigma} &= \sigma_{\max}-\sigma_{\mathrm{MPa}},\\
s_W &= W_{\max}-W_{\mathrm{mm}},\\
s_f &= \eta_{\max}-\eta_f,\\
s_P &= P_{\mu\mathrm{W}}-P_{\min}.
\end{aligned}
\end{equation}
Positive slack means the corresponding constraint is satisfied. A candidate is feasible iff
\begin{equation}
s_{\sigma}\ge 0,\qquad s_W\ge 0,\qquad s_f\ge 0,\qquad s_P\ge 0.
\end{equation}
If a task omits a limit, the implementation uses the default values \(\sigma_{\max}=50\) MPa, \(W_{\max}=5\) mm, \(\eta_{\max}=5\%\), and \(P_{\min}=1~\mu\mathrm{W}\). The oracle returns \((f_r,P_{\mu\mathrm{W}},\sigma_{\mathrm{MPa}},W_{\mathrm{mm}},\eta_f)\), the four signed slacks above, and the intermediate values \(m_{\mathrm{eff}}\), \(EI\), \(\theta\), and \(C_p\) for audit.

\section{Construction Audit and Artifact Availability}

The current VEH suite is not a loose aggregation of prompts. Its optimizer-facing stages are traceable to a concrete literature-extraction and admission pipeline. The upstream extraction audit covers 209 papers and 2090 field rows, with 151 papers carrying at least one issue and 77 carrying a high-risk issue. From this audit, 71 cantilever candidates are forwarded to oracle review, and the current P2 backbone admits 52 cleaned structure-review anchors. The current benchmark then packages these anchors into split-specific task banks whose manifests store source tables, input manifest hashes, seeds, and artifact SHA256 values.

\Needspace{16\baselineskip}
\begin{center}
\captionof{table}{Construction audit for the current benchmark release. The table separates source scale, split policy, and release evidence so that the benchmark can be audited as a dataset rather than only as a set of scores.}
\label{tab:appendix_construction_audit}
\centering
\scriptsize
\setlength{\tabcolsep}{3pt}
\renewcommand{\arraystretch}{1.08}
\begin{tabularx}{\textwidth}{p{17mm}p{24mm}p{27mm}p{18mm}Y}
\toprule
Stage & Upstream source & Split policy & Final inventory & Audit detail \\
\midrule
P1 & vetted VEH anchor contexts rendered through eight certification templates & three matched partitions under three seeds; subtype-balanced and not DOI-disjoint & 240 = 80/80/80 & legacy \texttt{dev}/\texttt{test\_id}/\texttt{test\_ood} labels are bookkeeping only; partitions use 20/19/19 anchor contexts with pairwise overlap 19/19/18 \\
P2 & 52 cleaned structure-review anchors from the extraction audit & optimizer-facing splits are DOI-disjoint: 26 oracle-domain anchors feed \texttt{dev}/\texttt{test\_id}, 26 projected out-of-domain-kept anchors feed \texttt{test\_ood} & 208 = 64/40/104 & 8 unit-suspect papers retained with explicit flags; admission report records source-anchor counts and no dropped task ids \\
P3 & dual-constraint trap projection from P2 & inherits P2 optimizer-facing split boundary & 156 = 64/40/52 & manifests record source P2 task bank, BKF table, split manifest, and trap policy \\
P4 & policy-conditioned ranking pools derived from P2 BKF records & inherits P2 optimizer-facing split boundary & 159 = 93/36/30 & 53 realized base pools across the release; manifests record profile counts, top-1 distinctness, and source-task lineage for each pool \\
\bottomrule
\end{tabularx}
\end{center}

\begin{center}
\captionof{table}{Task-statistics overview for the released VEH benchmark snapshot. Counts are computed from the released JSONL task banks and their split manifests.}
\label{tab:appendix_task_statistics}
\centering
\scriptsize
\setlength{\tabcolsep}{3.5pt}
\renewcommand{\arraystretch}{1.08}
\begin{tabularx}{\textwidth}{p{26mm}X}
\toprule
Dimension & Released snapshot \\
\midrule
Probe inventory & P1/P2/P3/P4 contain 240/208/156/159 tasks. \\
Split sizes & P1: 80/80/80; P2: 64/40/104; P3: 64/40/52; P4: 93/36/30 over \texttt{dev}/\texttt{test\_id}/\texttt{test\_ood}. \\
Subtype coverage & P1 has 8 triage subtypes; P2 has 4 search subtypes; P3 uses the dual-constraint trap family; P4 uses policy-conditioned full-ranking pools. \\
Difficulty coverage & P1: 60 base, 30 moderate, 150 hard; P2: 52 easy, 52 medium, 104 hard; P3: 156 hard; P4: 12 easy, 51 medium, 96 hard. \\
Source anchors & P1 uses 20 matched anchor contexts; P2 uses 52 cleaned anchors; P3 traces to 50 source anchors; P4 realizes 29 source-anchor ranking pools. \\
Source-group coverage & P2 has 104 oracle-domain and 104 out-of-domain-kept tasks; P3 has 104/52; P4 has 129/30; P1 is a matched triage stress stage rather than a DOI-disjoint source split. \\
\bottomrule
\end{tabularx}
\end{center}
\FloatBarrier

The release package is designed around these manifests and is available through the artifact page at \href{https://huggingface.co/datasets/AnonymousVehbench/vehbench}{huggingface.co/datasets/AnonymousVehbench/vehbench}. VEHBench is the benchmark name used in the paper; \texttt{diagbench} is the implementation package and code namespace for the same released artifact, so mixed names in code paths do not indicate a separate benchmark or incompatible evaluator. \texttt{diagbench-734D} is the historical repository slug used during peer review, not the benchmark title. The artifact contains the exact task JSONL files and manifests for P1--P4, the oracle and evaluator code, the task-generation scripts, the admission and split reports, and the per-model JSONL run logs used for the main and appendix tables. The intended artifact boundary is therefore the same as the paper boundary: each reported benchmark table should be traceable to a specific manifest-backed snapshot rather than to an undocumented local run.

The release preserves one auxiliary model-output directory under anonymized naming for audit completeness, so repository directory counts may exceed the 12-model roster used in the paper tables. All reported tables use the frozen 12 complete P1--P4 model runs; incomplete, deprecated, or auxiliary snapshots are retained only as release evidence and are not imputed into scores.

Contamination resistance in the current release is deliberately scoped. It means DOI/title-hash deduplication in the upstream paper registry, DOI-disjoint source-anchor separation for P2--P4, manifest-level lineage tracking, and evaluation on post-fix noleak P1 runs only. It does \emph{not} mean that we claim web-scale near-duplicate search over model pretraining corpora or model-specific pretraining decontamination. P1 is intentionally outside the source-disjoint claim and should be read as a matched robustness stress stage rather than as a held-out generalization benchmark.

\section{P3 Intervention Provenance}

The P3 intervention in the main paper is intentionally not a rerun over the entire 156-task bank. It is a frozen protocol audit on a 36-task subset selected by stratified sampling over \texttt{split} $\times$ \texttt{source\_p2\_subtype}. The selection rule is: draw three tasks per observed stratum, then top up to 12 tasks per split from the remaining pool while preserving split balance. This keeps every observed family visible while preventing the intervention table from collapsing to a single split.

\begin{center}
\captionof{table}{Full-bank versus intervention-subset composition for P3. The 36-task intervention subset is stratified rather than ad hoc: every observed split/subtype family is retained, and the final subset is exactly balanced across splits.}
\label{tab:appendix_p3_subset_provenance}
\centering
\small
\begin{tabular}{lp{50mm}p{50mm}}
\toprule
Split & Full 156-task bank & 36-task intervention subset \\
\midrule
\texttt{dev} & 64 tasks: boundary\_binding 20, paper\_like 17, power\_tight 12, resonance\_tuned 15 & 12 tasks: 3 / 3 / 3 / 3 across the same four strata \\
\texttt{test\_id} & 40 tasks: boundary\_binding 10, paper\_like 4, power\_tight 15, resonance\_tuned 11 & 12 tasks: 3 / 3 / 3 / 3 across the same four strata \\
\texttt{test\_ood} & 52 tasks: projected\_boundary 18, projected\_resonance 34 & 12 tasks: projected\_boundary 3, projected\_resonance 9 after split-balanced top-up \\
\bottomrule
\end{tabular}
\end{center}
\FloatBarrier

\section{Statistical Validity}

Headline gaps should be paired with uncertainty rather than interpreted as isolated point estimates. For continuous aggregates such as P1-Composite, P2 final feasible power ratio, and P4 Kendall-style scores, we use nonparametric bootstrap intervals over task rows~\citep{efron1993introduction}. For binary proportions such as P3-Success, we report Wilson intervals. Paired bootstrap deltas are used for the main near ties and for the P3 intervention audit so that uncertainty reflects within-task covariance rather than independent resampling.

\begin{center}
\captionof{table}{Headline uncertainty table for the four main probes.}
\label{tab:appendix_ci_table}
\centering
\scriptsize
\setlength{\tabcolsep}{3.5pt}
\renewcommand{\arraystretch}{1.08}
\begin{tabular}{lp{23mm}p{18mm}p{23mm}p{25mm}p{36mm}}
\toprule
Probe & Metric & Best & 95\% CI & Key delta & Note \\
\midrule
P1 & P1-Composite & 0.574 & [0.500, 0.641] & qwen3-max $-$ gemini-3.1-pro-preview: +0.025 [-0.078, 0.121] & not a trivial propose-bias win \\
P2 & feasible power ratio & 0.3904 & [0.3598, 0.4235] & gemini-3.1-pro-preview $-$ claude-4.6-sonnet: +0.1510 [0.1117, 0.1891] & separates closure from anchoring \\
P3 & P3-Success & 47.4\% & [39.8, 55.2]\% & hunyuan-hy3-preview $-$ mimo-v2.5-pro: +1.9 pts [-3.2, 7.1] & near tie at the recovery frontier \\
P4 & Kendall Tau & 0.887 & [0.862, 0.911] & gpt-5.4 $-$ qwen3.6-plus: +0.013 [-0.024, 0.050] & calibrates ranking claims \\
\bottomrule
\end{tabular}
\end{center}
\FloatBarrier

\begin{center}
\captionof{table}{Near-tie paired tests that matter most to the main paper.}
\label{tab:appendix_near_ties}
\centering
\scriptsize
\setlength{\tabcolsep}{3pt}
\renewcommand{\arraystretch}{1.08}
\begin{tabularx}{\textwidth}{p{32mm}p{38mm}Y}
\toprule
Pair & Test object & Why this pair matters \\
\midrule
hunyuan-hy3-preview vs mimo-v2.5-pro & P3-Success / RecoveryQuality & current recovery frontier; similar success, different recovery quality \\
gpt-5.4 vs qwen3.6-plus & P4 Full Tau / Exact & structural ranking versus endpoint preference matching \\
qwen3-max vs gemini-3.1-pro-preview & P1-Composite / P2 headline & certification-versus-closure split \\
\bottomrule
\end{tabularx}
\end{center}
\FloatBarrier

\begin{center}
\captionof{table}{P3 intervention numeric results. Delta columns report paired task-bootstrap 95\% intervals for \texttt{state\_summary} versus raw history.}
\label{tab:appendix_p3_intervention_delta_ci}
\centering
\scriptsize
\setlength{\tabcolsep}{4pt}
\renewcommand{\arraystretch}{1.08}
\begin{tabular}{lcccc}
\toprule
Model & Succ. raw/sum. & $\Delta$ success & Casc. raw/sum. & $\Delta$ cascade \\
\midrule
deepseek-r1 & 55.6 / 58.3 & +2.8 pts [-11.1, +16.7] & 30.3 / 8.7 & -21.6 pts [-37.5, -7.2] \\
gpt-5.4 & 63.9 / 63.9 & +0.0 pts [-13.9, +13.9] & 29.4 / 0.0 & -29.4 pts [-45.5, -14.7] \\
gemini-3.1-pro-preview & 50.0 / 66.7 & +16.7 pts [-2.8, +36.1] & 46.9 / 25.0 & -21.9 pts [-43.6, +0.5] \\
claude-4.6-sonnet & 30.6 / 63.9 & +33.3 pts [+16.7, +50.0] & 5.7 / 4.3 & -1.4 pts [-8.6, +4.4] \\
\midrule
Model mean & 50.0 / 63.2 & +13.2 pts [+4.9, +21.5] & 28.1 / 9.5 & -18.6 pts [-26.6, -10.6] \\
\bottomrule
\end{tabular}
\end{center}

\clearpage
\section{Metric Definitions}
\label{sec:metric_formulas}

This section defines the metrics used in the main paper and appendix. All scores are computed from persisted JSONL logs by deterministic evaluator code; no metric uses human annotation or LLM-as-a-judge.

\subsection{P1: Credible Triage}
\label{sec:metric_p1}

P1 is a one-shot three-action decision problem with gold labels in
\(\{\texttt{propose\_design}, \texttt{declare\_infeasible}, \texttt{request\_missing\_info}\}\).

\textbf{Accuracy and Macro-F1.} Accuracy is micro-averaged over all P1 tasks. Macro-F1 is the unweighted mean of the three action-level F1 scores:
\begin{equation}
\text{Macro-F1}
= \frac{1}{3}\sum_{c\in\mathcal{C}} F_{1,c},
\qquad
F_{1,c}=
\frac{2\,\mathrm{Prec}_c\,\mathrm{Rec}_c}
{\mathrm{Prec}_c+\mathrm{Rec}_c}.
\end{equation}
If \(\mathrm{Prec}_c+\mathrm{Rec}_c=0\), the corresponding \(F_{1,c}\) is defined as zero.

\textbf{Discipline scores.} Let \(A_r\), \(M_r\), and \(I_r\) denote recall for \texttt{propose\_design}, \texttt{request\_missing\_info}, and \texttt{declare\_infeasible}. Let \(A_s\), \(M_s\), and \(I_s\) denote the corresponding spurious-action rates on tasks whose gold label is not that action. Then:
\begin{align}
\mathrm{ACS} &= A_r(1-A_s),\\
\mathrm{MDS} &= M_r(1-M_s),\\
\mathrm{IDS} &= I_r(1-I_s).
\end{align}
ACS penalizes habitual proposing, MDS penalizes over-requesting information, and IDS penalizes over-refusal.

\textbf{P1-Composite.} The P1 headline metric is the weighted certification score used by the evaluator:
\begin{equation}
\begin{aligned}
\mathrm{P1Comp}={}&
0.40\,F1_{\mathrm{3class}}
+0.20\,\mathrm{ACS}
+0.15\,\mathrm{MDS} \\
&+0.15\,\mathrm{IDS}
+0.10\,F1_{\mathrm{subtype}} .
\end{aligned}
\end{equation}

\subsection{P2: Verifier-Guided Design Search}
\label{sec:metric_p2}

For a P2 trajectory, let \(\mathbf{x}_t\) be the design at step \(t\), \(V(\mathbf{x}_t)\ge0\) the total normalized oracle violation, and \(T\) the final evaluated step.

\textbf{First and final feasibility.} P2a is the fraction of tasks where \(V(\mathbf{x}_1)=0\). Final feasible rate is the fraction where \(V(\mathbf{x}_T)=0\).

\textbf{P2b final feasible power ratio.} For VEH tasks, the headline metric is:
\begin{equation}
\begin{aligned}
\mathrm{P2b}
= \frac{1}{N_{\mathrm{P2}}}\sum_{i=1}^{N_{\mathrm{P2}}}
&\mathbf{1}\!\left[V(\mathbf{x}^{(i)}_T)=0\right] \\
&\cdot
\frac{P_{\mathrm{out}}(\mathbf{x}^{(i)}_T)}
{P_{\mathrm{out}}(\mathbf{x}^{(i)}_{\mathrm{BKF}})} .
\end{aligned}
\end{equation}
The circuit audit uses the analogous final feasible objective score returned by the circuit oracle.

\textbf{Conditional objective ratio.}
\begin{equation}
\mathrm{CondObj} =
\frac{
\sum_{i:\,V(\mathbf{x}^{(i)}_T)=0}
\mathrm{obj}(\mathbf{x}^{(i)}_T)}
{|\{i: V(\mathbf{x}^{(i)}_T)=0\}|}.
\end{equation}

\textbf{Trajectory diagnostics.} Violation reduction consistency is the fraction of transitions with \(V(\mathbf{x}_{t+1})\le V(\mathbf{x}_t)-10^{-6}\). Directed update rate is the fraction of violated-constraint updates that move the relevant oracle metric in the feedback-implied direction. Feasibility preservation is the fraction of feasible-to-feasible transitions. Mean log edit delta is:
\begin{equation}
\delta_{\mathrm{edit}}
=
\frac{1}{|\mathcal{D}|}\sum_{d\in\mathcal{D}}
\frac{\left|\log(x_{t+1,d}/x_{t,d})\right|}
{\log(x_d^{\max}/x_d^{\min})}.
\end{equation}
Over-edit and no-op rates threshold this quantity in the corresponding evaluator.

\subsection{P3: Post-Trap Recovery}
\label{sec:metric_p3}

P3 starts from a corrupted trajectory state. Escape rate is the fraction of tasks where at least one step reduces violation relative to the trapped state. Cascade rate is the fraction of escaped tasks where a new structurally coupled violation appears after the escape move. Dead-budget rate is the fraction of tasks with no meaningful post-trap design change.

\textbf{P3-Success and recovery quality.}
\begin{equation}
\mathrm{P3Success}
= \frac{1}{N_{\mathrm{P3}}}
\sum_i \mathbf{1}\!\left[V(\mathbf{x}^{(i)}_T)=0\right].
\end{equation}
\begin{equation}
\mathrm{RecoveryQuality}
=
\frac{
\sum_{i:\,V(\mathbf{x}^{(i)}_T)=0}
\mathrm{obj}(\mathbf{x}^{(i)}_T)}
{|\{i: V(\mathbf{x}^{(i)}_T)=0\}|}.
\end{equation}
The raw-history versus state-summary delta is the paired task-level difference in P3-Success between the two prompt representations, reported with paired bootstrap intervals.

\subsection{P4: Policy-Conditioned Ranking}
\label{sec:metric_p4}

Each P4 task contains five oracle-feasible candidates. Full Kendall \(\tau\) is Kendall \(\tau_b\) between the model ranking and oracle ranking. Exact match, top-1 accuracy, top-2 set accuracy, pairwise accuracy, and policy-flip accuracy are computed directly from the predicted order.

\textbf{BARS.} For the VEH P4-full bank, the headline BARS score is computed only on balanced-active rows:
\begin{equation}
\begin{aligned}
\mathrm{BARS}
= {}&0.55\,\tau_{\mathrm{BA}}
+0.25\,\mathrm{PairAcc}_{\mathrm{BA}} \\
&+0.20\,\mathrm{Exact}_{\mathrm{BA}},
\end{aligned}
\end{equation}
where BA denotes the balanced-active subset. For the circuit audit, the analogous ranking score uses scaled Kendall \((\tau+1)/2\), policy-flip accuracy, and exact match with the same weights.

\subsection{Response-Control Profile Scores}
\label{sec:metric_profiles}

Profile scores are log-derived diagnostics, not extra model prompts and not human annotations. They summarize how a model behaves in the released logs: whether it acts too readily or too cautiously before search, whether edits are local or destructive, whether verifier feedback changes the next action, whether corrupted history triggers reset-like behavior, and whether fixed candidates are ranked according to the stated policy.

For each dimension \(D\), the evaluator collects a fixed indicator set \(\mathcal{M}_D\), orients every indicator so higher is better, min-max scales it across the compared models, and averages:
\begin{equation}
\mathrm{Profile}_D(m)
=
\frac{1}{|\mathcal{M}_D|}
\sum_{q\in\mathcal{M}_D} z_q(m).
\end{equation}
For an indicator \(q\), the normalized value is:
\begin{equation}
z_q(m)=
\frac{s_q(m)-\min_{m'}s_q(m')}
{\max_{m'}s_q(m')-\min_{m'}s_q(m')+\epsilon},
\end{equation}
after applying an orientation transform when lower raw values are better (for example parse-error, cascade, over-edit, or spurious-action rates). We use \(\epsilon=10^{-12}\) only to avoid division by zero; constant indicators therefore contribute no cross-model separation. The compared set for the main paper is the 12 complete P1--P4 model runs. The profiles are used only for diagnosis and correlation analysis; they do not change the task-level scores in Tables~\ref{tab:main_results} or~\ref{tab:appendix_p1_full}--\ref{tab:appendix_p4_full}.

\Needspace{15\baselineskip}
\begin{center}
\begin{minipage}{\textwidth}
\centering
\captionof{table}{Indicators used for response-control profile extraction. All indicators are direction-normalized before averaging.}
\label{tab:appendix_profile_indicator_sets}
\scriptsize
\setlength{\tabcolsep}{4pt}
\renewcommand{\arraystretch}{1.08}
\begin{tabularx}{\textwidth}{lY}
\toprule
Dimension & Indicators \\
\midrule
Action discipline & P1 action-distribution alignment, macro-F1, missing recall, infeasible recall, propose precision, non-invalid rate \\
Edit style & P2 bounded-local-edit rate, feasibility preservation, directed update rate, final feasible rate, non-destructive edit rate, protocol-valid rate \\
Feedback conditioning & P2 violation reduction, P2 utility improvement, P2 best-so-far AUC, P3 violation reduction, P3 post-feedback feasible rate \\
State-reset effort & P3 trap escape, explicit replan, escape quality, non-cascade rate, non-dead-budget rate, state-summary success gain, state-summary cascade reduction \\
Policy execution & P4 scaled full Tau, balanced-active BARS, balanced-active policy-sensitive pair accuracy, exact match, top-1 accuracy, non-Pareto-violation rate, non-parse-error rate \\
\bottomrule
\end{tabularx}
\end{minipage}
\end{center}

\paragraph{Interpretation.}
The profile dimensions are intentionally behavioral and stage-local. Action discipline is expected to align with P1 because P1 is an entry-control problem. Edit style and feedback conditioning are expected to align with P2 because P2 rewards bounded verifier-guided design search. State-reset effort is expected to matter most in P3 because P3 contaminates the history itself, but it measures recovery-oriented effort rather than final recovery success. Policy execution is expected to align with P4 because search has been removed and only policy-conditioned ordering remains. In the main results, the strongest positive alignments follow the intended diagnostic pattern: action discipline tracks P1-Composite, edit style tracks P2 final feasible power ratio, and policy execution tracks P4 Tau/BARS. Weak or negative off-diagonal relationships are part of the finding: a response-control prior that is helpful at one trusted-state boundary can be neutral or harmful at another.

\section{Calibration References}

Weak baselines are useful here because the benchmark is deliberately structured. For P1, the main concern is degenerate label-prior matching. For P2, the main concern is whether headline gains reflect true improvement beyond the literature-derived seed or only feasibility preservation. For P3, the relevant weak behavior is continuity with the corrupted state rather than explicit reset. For P4, the relevant floor is chance agreement with a five-item ordering.

\begin{center}
\captionof{table}{Calibration references corresponding to the main-text weak baselines. These are not competitors to the frontier roster; they anchor how much of each stage can be explained by trivial behavior.}
\label{tab:appendix_calibration}
\centering
\small
\begin{tabular}{lp{30mm}p{23mm}p{60mm}}
\toprule
Probe & Baseline reference & Score & Interpretation \\
\midrule
P1 & always-\texttt{propose\_design} on the dev split & 56.3\% accuracy, 0.240 macro-F1, 0.0 worst-action recall & a degenerate action prior can look superficially acceptable on raw accuracy, which is why P1-Composite and subtype-balanced reporting are necessary \\
P2 & anchor-fixed heuristic on the 208-task main bank & 100\% feasible coverage, 0.3065 mean power ratio & many frontier models fail to preserve even the trivial feasible anchor, while gemini-3.1-pro-preview is the only one to beat this heuristic on the headline utility ratio \\
P3 & last-state continuation / no-reset heuristic & 0\% explicit reset by construction; weak reference for escape, cascade, and recovery columns & the useful floor is not a single scalar success rate but the failure mode produced by trusting corrupted state; P3 therefore reports escape, cascade, dead-budget, and recovery jointly \\
P4 & uniform random permutation over five feasible candidates & $\mathbb{E}[\tau]=0$, exact 0.83\%, top-1 20\% & observed ranking scores are far above chance, so P4 is not a random-agreement artifact \\
\bottomrule
\end{tabular}
\end{center}

\clearpage
\section{Full Leaderboard Tables}

The appendix carries the full all-model tables for each main probe so that the main-paper mechanism claims remain auditable. All four tables use the frozen 12-model complete roster; runs without complete P1--P4 coverage are excluded rather than imputed.

This block should be read as a diagnostic audit rather than four extra leaderboards. Read each table column-wise: the headline metric gives the rank, while the diagnostic columns identify the failure mode behind that rank. The full evaluator formulas are collected in Appendix~\ref{sec:metric_formulas}; the dictionary below states how to compute the columns used in Tables~\ref{tab:appendix_p1_full}--\ref{tab:appendix_p4_full} and the diagnostic error rates used in Figure~\ref{fig:stage_error_landscape}.

\subsection{Metric and Diagnostic-Rate Dictionary}
\label{sec:appendix_fulltable_metric_dictionary}
\label{sec:metric_diagnostic_dictionary}

All quantities are computed from released JSONL task logs and oracle traces. Unless otherwise stated, rates are first computed per model and then averaged across the 12 complete models when the main text reports a roster mean. Diagnostic error families are \emph{non-exclusive}: they use different denominators and are not intended to sum to one. For example, if a missing-information P1 task is incorrectly answered with \texttt{propose\_design}, it is both a missing-information miss and an over-action event.

\paragraph{P1 column definitions.}
Let \(g_i\) be the gold P1 action and \(\hat a_i\) the model action for task \(i\), with actions in \(\{\texttt{propose\_design}, \texttt{request\_missing\_info}, \texttt{declare\_infeasible}\}\). Accuracy is \(N^{-1}\sum_i \mathbf{1}[\hat a_i=g_i]\). Macro-F1 is the unweighted mean of the three action-level F1 scores. For action \(c\), recall and spurious-action rate are:
\begin{align}
\mathrm{Rec}_c
&=
\frac{\left|\{i: g_i=c \ \mathrm{and}\ \hat a_i=c\}\right|}
{\left|\{i: g_i=c\}\right|},\\
\mathrm{Spur}_c
&=
\frac{\left|\{i: g_i\ne c \ \mathrm{and}\ \hat a_i=c\}\right|}
{\left|\{i: g_i\ne c\}\right|}.
\end{align}
The table columns ACS, MDS, and IDS are:
\begin{align}
\mathrm{ACS}&=\mathrm{Rec}_{\texttt{propose}}(1-\mathrm{Spur}_{\texttt{propose}}),\\
\mathrm{MDS}&=\mathrm{Rec}_{\texttt{request}}(1-\mathrm{Spur}_{\texttt{request}}),\\
\mathrm{IDS}&=\mathrm{Rec}_{\texttt{infeasible}}(1-\mathrm{Spur}_{\texttt{infeasible}}).
\end{align}
Subtype F1 averages F1 over the released P1 subtypes. P1-Composite is \(0.40\) macro-F1 \(+0.20\) ACS \(+0.15\) MDS \(+0.15\) IDS \(+0.10\) subtype F1. The P1 error-family rates in Figure~\ref{fig:stage_error_landscape} are: over-action \(=\mathrm{Spur}_{\texttt{propose}}\), over-refusal \(=\mathrm{clip}_{[0,1]}(\mathrm{Spur}_{\texttt{request}}+\mathrm{Spur}_{\texttt{infeasible}})\), and missing-information miss \(=1-\mathrm{Rec}_{\texttt{request}}\).

\paragraph{P2 column definitions.}
For P2 task \(i\), let \(F_i\in\{0,1\}\) indicate final oracle feasibility, \(P_i\) be the final output power, and \(P_i^{\mathrm{BKF}}>0\) be the benchmark-known feasible reference power. P2a is first-step feasibility. Final feasible rate is \(N^{-1}\sum_i F_i\). The headline P2b final feasible power ratio is:
\begin{equation}
\mathrm{P2b}=\frac{1}{N}\sum_i F_i\,\frac{P_i}{P_i^{\mathrm{BKF}}},
\end{equation}
so infeasible final designs contribute zero. Conditional ratio is:
\begin{equation}
\mathrm{CondRatio}=\frac{1}{\sum_i F_i}\sum_{i:F_i=1}\frac{P_i}{P_i^{\mathrm{BKF}}},
\end{equation}
which measures objective quality only among closed feasible searches. Impr. is the trajectory improvement rate: the fraction of consecutive proposal steps whose utility increases, where utility is normalized power minus penalties for active frequency, stress, and displacement violations. AUC is the mean best-so-far normalized objective over the fixed query budget; Queries is the mean number of oracle calls used. The P2 error-family rates are infeasible closure \(=1-\) final feasible rate, destructive edit \(=\) feasible-to-infeasible transition rate, invalid/no-op proxy \(=\) protocol-invalid rate, and utility loss \(=1-\mathrm{CondRatio}\). Utility loss therefore asks how much objective quality remains missing \emph{after} successful feasible closure; it is not mutually exclusive with closure failure.

\paragraph{P3 column definitions.}
For P3, Success is final feasibility after the corrupted-state recovery budget. RecQ is the objective ratio among successful recoveries. First feas. is whether the trajectory reaches feasibility before the final step. Escape is the fraction of tasks where the model moves away from the known corrupted trap. Cascade is the fraction of escaped tasks that introduce a new coupled violation. Dead marks trajectories with no meaningful recovery action, and Replan marks explicit reset or replanning behavior. The P3 error-family rates are escape failure \(=1-\) escape rate, cascade \(=\) cascade rate, dead budget \(=\) dead-budget rate, and post-escape failure \(=1-\mathrm{Success}/\max(\mathrm{Escape},\epsilon)\), clipped to \([0,1]\). Post-escape failure isolates models that notice the trap but still fail to stabilize a feasible design.

\paragraph{P4 column definitions.}
For P4, all candidates are oracle-feasible before ranking. Full Tau is Kendall \(\tau_b\) between the model order and the oracle policy order over the full five-item pool. Exact is full-order exact match. Top1 and Top2 measure whether the best candidate and the top-two set are preserved. Pareto Tau measures ordering consistency on Pareto-relevant comparisons. BARS is the balanced-active ranking score defined in Appendix~\ref{sec:metric_formulas}. Parse / viol. reports parse failure and attempted candidate-modification or dominance-violation behavior. The P4 error-family rates are policy mismatch \(=1-\) all-policy-sensitive pair accuracy, dominance error \(=\) Pareto-violation rate, parse failure \(=\) parse-error rate, and top-choice miss \(=1-\) top-1 accuracy.

\begin{center}
\captionof{table}{Interpretive map for the appendix full-table block. Each table is tied to a benchmark claim rather than presented as metric inventory alone.}
\label{tab:appendix_fulltable_map}
\centering
\small
\begin{tabular}{lp{21mm}p{16mm}p{16mm}p{16mm}p{16mm}p{18mm}}
\toprule
Probe & Core columns & Diagnostic 1 & Diagnostic 2 & Diagnostic 3 & Diagnostic 4 & Primary mechanism claim served \\
\midrule
P1 & Acc / Macro-F1 / ACS / MDS / IDS / subtype F1 / composite & over-act prior & hard infeasibility misses & missing-info discipline & balanced selectivity & raw accuracy does not equal credible triage \\
P2 & P2a / P2b / final feasible rate / conditional ratio / Impr. (P2c) / queries & anchoring & closure & best-so-far improvement & query efficiency & start is not finish \\
P3 & success / recovery quality / trap escape / cascade / dead budget / explicit replan & no-escape failure & escaped-but-unrecovered & false recovery & post-escape quality & escape is not recovery \\
P4 & Full Tau / Exact / Top-1 / Top-2 / Pareto Tau / BARS & structural order & strict endpoint ranking & policy-active ranking & parse stability & structural ranking is not policy sensitivity \\
\bottomrule
\end{tabular}
\end{center}
\FloatBarrier

\clearpage
\Needspace{28\baselineskip}
\paragraph{P1 metrics and readout.}
P1 is a one-shot triage task. Accuracy counts exact agreement with the oracle label; macro-F1 averages the three action labels so that a model cannot score well by favoring the majority action. ACS, MDS, and IDS are discipline scores for \texttt{propose\_design}, \texttt{request\_missing\_info}, and \texttt{declare\_infeasible}; each multiplies recall by one minus the corresponding spurious-action rate. Subtype F1 checks whether performance holds across near-feasible, missing-info, and infeasible subtypes. The headline P1-Composite combines these terms, so it rewards balanced entry discipline rather than raw willingness to produce a design. The table shows why this is necessary: several models have similar raw accuracy, but qwen3-max and gemini-3.1-pro-preview separate because their discipline scores are more balanced, while claude-4.6-sonnet exposes over-refusal or action-mismatch behavior that accuracy alone would under-explain.

\Needspace{22\baselineskip}
\noindent\begin{minipage}{\textwidth}
\centering
\captionof{table}{P1 full table. Compact headers keep certification metrics readable.}
\label{tab:appendix_p1_full}
\scriptsize
\setlength{\tabcolsep}{3.5pt}
\renewcommand{\arraystretch}{1.08}
\begin{tabular}{llllllll}
\toprule
Model & Acc & F1 & ACS & MDS & IDS & Subtype F1 & Comp. \\
\midrule
qwen3-max & 65.4\% & 0.647 & 0.429 & 0.533 & 0.486 & 0.768 & 0.574 \\
gemini-3.1-pro-preview & 69.2\% & 0.636 & 0.430 & 0.667 & 0.218 & 0.762 & 0.549 \\
\texttt{o4-mini} & 70.0\% & 0.623 & 0.388 & 0.396 & 0.361 & 0.776 & 0.518 \\
deepseek-r1 & 66.2\% & 0.594 & 0.360 & 0.510 & 0.298 & 0.730 & 0.504 \\
gpt-5.4 & 43.3\% & 0.486 & 0.224 & 0.657 & 0.223 & 0.569 & 0.428 \\
hunyuan-hy3-preview & 66.2\% & 0.539 & 0.295 & 0.333 & 0.200 & 0.700 & 0.425 \\
deepseek-v3 & 64.6\% & 0.483 & 0.206 & 0.352 & 0.100 & 0.668 & 0.369 \\
\texttt{llama-3.3-70b} & 62.5\% & 0.467 & 0.242 & 0.200 & 0.192 & 0.669 & 0.361 \\
mimo-v2.5-pro & 61.7\% & 0.433 & 0.152 & 0.244 & 0.083 & 0.644 & 0.317 \\
\texttt{qwen3.6-plus} & 61.3\% & 0.422 & 0.143 & 0.200 & 0.100 & 0.641 & 0.306 \\
deepseek-v4-pro & 61.3\% & 0.422 & 0.143 & 0.200 & 0.100 & 0.641 & 0.306 \\
claude-4.6-sonnet & 57.5\% & 0.305 & 0.057 & 0.000 & 0.100 & 0.583 & 0.207 \\
\bottomrule
\end{tabular}

\noindent\textbf{Audit note.} The exact tie between \texttt{qwen3.6-plus} and deepseek-v4-pro is preserved from the frozen scoring snapshot rather than broken by post-hoc rounding or reranking. Exact vector ties are therefore treated as tied diagnostic rows, not as evidence for an ordering between the two models.

\noindent\textbf{Takeaway.} P1 separates credible entry control from raw action frequency: the strongest rows combine reasonable proposing with missing-information and infeasibility discipline, while weak rows expose over-refusal or over-action even when accuracy is not catastrophic.
\end{minipage}

\Needspace{14\baselineskip}
\paragraph{P2 metrics and readout.}
P2 is a verifier-guided design-search task under oracle feedback. P2a is first-step feasibility, which measures whether the first edit lands directly in a feasible region. P2b is the headline final feasible power ratio: the final design contributes only if it is feasible, and its output power is normalized by the benchmark-known feasible reference. Final feasibility separates closure from objective quality, conditional ratio reports quality only among successful searches, improvement measures best-so-far utility gain, AUC summarizes trajectory quality across the budget, and queries records how many oracle calls were needed. The main conclusion is that search closure and utility are not the same: gemini-3.1-pro-preview dominates P2b because it combines high final feasibility with high feasible utility, while models with respectable first-step feasibility can still finish with low final ratio if their subsequent edits overfit one constraint or lose power.

\begin{center}
\captionof{table}{P2 full table. P2a is first-step feasibility; P2b is the headline final feasible power ratio.}
\label{tab:appendix_p2_full}
\scriptsize
\setlength{\tabcolsep}{3.5pt}
\renewcommand{\arraystretch}{1.08}
\begin{tabular}{llllllll}
\toprule
Model & P2a & P2b & Final feas. & Cond. ratio & Impr. & AUC & Queries \\
\midrule
gemini-3.1-pro-preview & 26.0\% & 0.3904 & 96.2\% & 0.4101 & 0.7902 & 0.3388 & 2.63 \\
deepseek-r1 & 29.8\% & 0.2413 & 67.8\% & 0.3585 & 0.7901 & 0.2999 & 3.47 \\
claude-4.6-sonnet & 13.0\% & 0.2394 & 62.5\% & 0.3891 & 0.7037 & 0.2797 & 4.41 \\
\texttt{qwen3.6-plus} & 30.8\% & 0.2063 & 61.1\% & 0.3432 & 0.6466 & 0.2880 & 3.87 \\
qwen3-max & 23.6\% & 0.1955 & 55.8\% & 0.3537 & 0.6649 & 0.2894 & 4.21 \\
hunyuan-hy3-preview & 2.4\% & 0.1609 & 48.6\% & 0.3380 & 0.8033 & 0.2480 & 4.79 \\
deepseek-v3 & 7.2\% & 0.1565 & 44.2\% & 0.3576 & 0.5907 & 0.2766 & 4.68 \\
\texttt{o4-mini} & 22.6\% & 0.1551 & 42.8\% & 0.3665 & 0.7571 & 0.3119 & 3.59 \\
gpt-5.4 & 7.2\% & 0.1329 & 36.1\% & 0.3841 & 0.6828 & 0.2708 & 4.75 \\
deepseek-v4-pro & 30.3\% & 0.1294 & 41.3\% & 0.3205 & 0.6167 & 0.2935 & 4.26 \\
\texttt{llama-3.3-70b} & 23.6\% & 0.1197 & 37.0\% & 0.3277 & 0.6673 & 0.2895 & 4.38 \\
mimo-v2.5-pro & 26.4\% & 0.1073 & 33.2\% & 0.3282 & 0.6714 & 0.3075 & 4.50 \\
\bottomrule
\end{tabular}
\end{center}

\noindent\textbf{Takeaway.} P2 is won by feasible closure plus useful final power, not by first-step feasibility alone; gemini-3.1-pro-preview's lead is therefore a design-search trajectory result rather than an anchoring artifact.

\Needspace{14\baselineskip}
\paragraph{P3 metrics and readout.}
P3 starts from a corrupted trajectory. Success is final feasibility at the end of the recovery budget; recovery quality reports the objective quality among successful recoveries; first feasible identifies whether a model ever reaches a feasible state early in the recovery trace. Escape measures whether the model moves away from the known trap, cascade records whether the escape introduces a new coupled violation, dead-budget marks trajectories with no meaningful recovery action, and replan counts explicit reset or replanning behavior. These columns show that P3 is not ordinary clean design search: claude-4.6-sonnet is the clearest diagnostic row, with 96.8\% escape but only 0.6\% first-feasible and 16.0\% final success. It often notices that the raw history is wrong, but it rarely stabilizes back into a feasible design. deepseek-r1 and gpt-5.4 combine high escape with better final recovery; hunyuan-hy3-preview and mimo-v2.5-pro lead success through different trade-offs in recovery quality and cascade control.

\noindent\begin{minipage}{\textwidth}
\centering
\captionof{table}{P3 full table. Columns separate recovery success, escape, cascade, dead-budget, and replanning behavior.}
\label{tab:appendix_p3_full}
\scriptsize
\setlength{\tabcolsep}{3.5pt}
\renewcommand{\arraystretch}{1.08}
\begin{tabular}{llllllll}
\toprule
Model & Succ. & RecQ & First feas. & Escape & Cascade & Dead & Replan \\
\midrule
hunyuan-hy3-preview & 47.4\% & 0.4876 & 40.4\% & 77.6\% & 9.1\% & 25.3\% & 7.7\% \\
mimo-v2.5-pro & 45.5\% & 0.6263 & 41.0\% & 63.5\% & 4.0\% & 26.7\% & 17.9\% \\
deepseek-r1 & 42.9\% & 0.2708 & 22.4\% & 92.3\% & 26.4\% & 25.1\% & 26.3\% \\
gpt-5.4 & 42.3\% & 0.3723 & 26.3\% & 90.4\% & 21.3\% & 25.2\% & 6.4\% \\
gemini-3.1-pro-preview & 37.2\% & 0.3194 & 26.9\% & 92.3\% & 50.7\% & 25.8\% & 10.9\% \\
deepseek-v3 & 35.3\% & 0.2868 & 23.1\% & 82.7\% & 27.1\% & 27.6\% & 57.7\% \\
qwen3-max & 30.1\% & 0.2612 & 22.4\% & 85.9\% & 12.7\% & 25.7\% & 49.4\% \\
deepseek-v4-pro & 28.8\% & 0.1986 & 27.6\% & 90.4\% & 17.0\% & 24.3\% & 99.4\% \\
\texttt{qwen3.6-plus} & 27.6\% & 0.1453 & 12.8\% & 94.9\% & 23.0\% & 25.6\% & 0.6\% \\
\texttt{o4-mini} & 26.3\% & 0.1389 & 12.2\% & 69.2\% & 17.6\% & 26.6\% & 8.3\% \\
claude-4.6-sonnet & 16.0\% & 0.0960 & 0.6\% & 96.8\% & 5.3\% & 25.7\% & 0.0\% \\
\texttt{llama-3.3-70b} & 3.2\% & 0.0926 & 2.6\% & 34.6\% & 3.7\% & 14.3\% & 0.0\% \\
\bottomrule
\end{tabular}
\end{minipage}

\noindent\textbf{Takeaway.} P3 shows that escaping a corrupted trajectory is not the same as recovering: high escape with low success marks continuity or cascade failure, while successful recovery requires both reset and stabilization.

\Needspace{14\baselineskip}
\paragraph{P4 metrics and readout.}
P4 removes search from the problem: all candidates are oracle-feasible before the model ranks them. Full Kendall Tau measures global order agreement with the oracle policy ranking; exact match requires the entire five-item order to be correct; top-1 and top-2 set accuracy measure whether the highest-priority choices are preserved; Pareto Tau isolates ranking consistency on Pareto-relevant comparisons; BARS emphasizes balanced-active policy-sensitive rows; parse and violation rates record formatting failures and attempts to alter fixed candidates. The table shows that P4 is not reducible to one ranking statistic. gpt-5.4 leads Full Tau, qwen3.6-plus is strongest on exact match, and gemini-3.1-pro-preview has a high BARS despite lower Tau, indicating that policy execution has multiple sub-behaviors rather than a single scalar notion of ``ranking skill.''

\begin{center}
\captionof{table}{P4 full table. Compact notation keeps ranking, Pareto, and formatting diagnostics in one table.}
\label{tab:appendix_p4_full}
\scriptsize
\setlength{\tabcolsep}{3.5pt}
\renewcommand{\arraystretch}{1.08}
\begin{tabular}{llllllll}
\toprule
Model & Tau & Exact & Top1 & Top2 & Pareto & BARS & Parse / viol. \\
\midrule
gpt-5.4 & 0.887 & 57.2\% & 83.0\% & 81.1\% & 1.000 & 0.6258 & 0.0\% / 0.0\% \\
\texttt{qwen3.6-plus} & 0.877 & 66.3\% & 79.1\% & 83.4\% & 1.000 & 0.6912 & 0.0\% / 0.0\% \\
deepseek-v3 & 0.860 & 54.7\% & 77.4\% & 79.2\% & 0.984 & 0.5751 & 0.0\% / 0.8\% \\
mimo-v2.5-pro & 0.843 & 52.8\% & 72.3\% & 79.2\% & 0.993 & 0.6012 & 0.0\% / 0.3\% \\
claude-4.6-sonnet & 0.840 & 56.6\% & 72.3\% & 79.2\% & 1.000 & 0.6781 & 0.0\% / 0.0\% \\
hunyuan-hy3-preview & 0.839 & 50.9\% & 77.4\% & 79.9\% & 0.943 & 0.5610 & 0.0\% / 2.8\% \\
qwen3-max & 0.835 & 49.7\% & 78.0\% & 76.7\% & 0.983 & 0.5367 & 0.0\% / 0.8\% \\
deepseek-r1 & 0.833 & 56.0\% & 71.7\% & 76.7\% & 1.000 & 0.6835 & 0.0\% / 0.0\% \\
gemini-3.1-pro-preview & 0.824 & 54.1\% & 71.7\% & 78.6\% & 0.975 & 0.6988 & 1.3\% / 1.3\% \\
deepseek-v4-pro & 0.794 & 43.4\% & 60.4\% & 74.2\% & 0.977 & 0.5832 & 0.0\% / 1.2\% \\
\texttt{o4-mini} & 0.780 & 50.0\% & 60.0\% & 72.0\% & 1.000 & 0.5988 & 0.0\% / 0.0\% \\
\texttt{llama-3.3-70b} & 0.714 & 34.0\% & 54.1\% & 65.4\% & 0.959 & 0.5386 & 0.0\% / 2.0\% \\
\bottomrule
\end{tabular}
\end{center}

\noindent\textbf{Takeaway.} P4 ranking has multiple sub-behaviors: global order quality, exact endpoint order, top-choice preservation, and policy-sensitive BARS can diverge, which is why P4 is not reducible to one scalar preference score.

\section{Split-Resolved Tables}

This block checks that the main-paper claims are not driven by a single split. The P2 closure pattern and P4 non-monotonic ranking pattern remain visible across splits, while P3 contains genuine near ties that should be read with the confidence intervals in Appendix~\ref{tab:appendix_ci_table}. P1 uses matched partitions rather than DOI-disjoint source splits, so its split labels audit robustness rather than held-out generalization. The split-resolved audit tables below use the same 12 complete P1--P4 model runs as the main leaderboard and stage-aware selection simulation. Exact split-invariant rows, such as \texttt{qwen3.6-plus} and claude-4.6-sonnet in P1, are retained as audit signals from the released snapshot. Each table reports the main stage headline plus one supporting diagnostic in a split-resolved view.

\begin{center}
\captionof{table}{Split and subtype reporting plan. The appendix presents split-resolved views as mechanism audits rather than as extra leaderboards.}
\label{tab:appendix_split_subtype_map}
\centering
\small
\begin{tabular}{lp{28mm}p{32mm}p{42mm}}
\toprule
Probe & Split-resolved view & Subtype families & Main claim audited \\
\midrule
P1 & matched-partition composite and diagnostics (legacy \texttt{dev}/\texttt{test\_id}/\texttt{test\_ood} labels) & infeasible-margin, infeasible-structural, missing-info families & certification robustness is not driven by one matched partition or blocker family \\
P2 & dev / test\_id / test\_ood headline and feasible rate & paper-like, resonance-tuned, power-tight, boundary-binding & closure dominates anchoring as the main search distinction \\
P3 & dev / test\_id / test\_ood success and recovery quality & trap families and cascade diagnostics & corrupted-state recovery is not reducible to nominal design-search quality \\
P4 & dev / test\_id / test\_ood Full Tau and Exact & balanced, performance-first, reliability-first, BARS & structural ranking and policy-sensitive ranking remain partially separable \\
\bottomrule
\end{tabular}
\end{center}
\FloatBarrier

\begin{center}
\captionof{table}{P1 matched-partition table. Each cell reports P1-Composite and raw accuracy.}
\label{tab:appendix_p1_split}
\centering
\scriptsize
\setlength{\tabcolsep}{4pt}
\renewcommand{\arraystretch}{1.08}
\begin{tabular}{lccc}
\toprule
Model & dev & test\_id & test\_ood \\
\midrule
qwen3-max & 0.532 / 61.3\% & 0.622 / 70.0\% & 0.567 / 65.0\% \\
gemini-3.1-pro-preview & 0.545 / 70.0\% & 0.525 / 66.2\% & 0.578 / 71.2\% \\
\texttt{o4-mini} & 0.542 / 71.2\% & 0.509 / 70.0\% & 0.502 / 68.8\% \\
deepseek-r1 & 0.517 / 68.8\% & 0.512 / 66.2\% & 0.482 / 63.7\% \\
gpt-5.4 & 0.374 / 37.5\% & 0.469 / 47.5\% & 0.440 / 45.0\% \\
hunyuan-hy3-preview & 0.448 / 67.5\% & 0.422 / 66.2\% & 0.404 / 65.0\% \\
deepseek-v3 & 0.357 / 63.7\% & 0.385 / 65.0\% & 0.364 / 65.0\% \\
\texttt{llama-3.3-70b} & 0.362 / 61.3\% & 0.393 / 65.0\% & 0.328 / 61.3\% \\
mimo-v2.5-pro & 0.336 / 62.5\% & 0.308 / 61.3\% & 0.306 / 61.3\% \\
\texttt{qwen3.6-plus} & 0.306 / 61.3\% & 0.306 / 61.3\% & 0.306 / 61.3\% \\
deepseek-v4-pro & 0.275 / 60.0\% & 0.336 / 62.5\% & 0.306 / 61.3\% \\
claude-4.6-sonnet & 0.207 / 57.5\% & 0.207 / 57.5\% & 0.207 / 57.5\% \\
\bottomrule
\end{tabular}
\end{center}
\FloatBarrier

\begin{center}
\captionof{table}{P2 split-resolved table. Each cell reports the final feasible power ratio and final feasible coverage.}
\label{tab:appendix_p2_split}
\centering
\scriptsize
\setlength{\tabcolsep}{4pt}
\renewcommand{\arraystretch}{1.08}
\begin{tabular}{lccc}
\toprule
Model & dev & test\_id & test\_ood \\
\midrule
gemini-3.1-pro-preview & 0.4017 / 98.4\% & 0.4324 / 90.0\% & 0.3672 / 97.1\% \\
deepseek-r1 & 0.2560 / 62.5\% & 0.3049 / 82.5\% & 0.2078 / 65.4\% \\
claude-4.6-sonnet & 0.2199 / 54.7\% & 0.3095 / 75.0\% & 0.2245 / 62.5\% \\
\texttt{qwen3.6-plus} & 0.2355 / 57.8\% & 0.2316 / 65.0\% & 0.1786 / 61.5\% \\
qwen3-max & 0.2182 / 50.0\% & 0.2486 / 60.0\% & 0.1612 / 57.7\% \\
hunyuan-hy3-preview & 0.1689 / 43.8\% & 0.1703 / 50.0\% & 0.1523 / 51.0\% \\
deepseek-v3 & 0.2052 / 48.4\% & 0.1761 / 47.5\% & 0.1189 / 40.4\% \\
\texttt{o4-mini} & 0.1587 / 42.2\% & 0.1220 / 35.0\% & 0.1655 / 46.2\% \\
gpt-5.4 & 0.0991 / 26.6\% & 0.0541 / 12.5\% & 0.1841 / 51.0\% \\
deepseek-v4-pro & 0.1891 / 51.6\% & 0.1165 / 40.0\% & 0.0977 / 35.6\% \\
\texttt{llama-3.3-70b} & 0.1781 / 43.8\% & 0.1046 / 32.5\% & 0.0896 / 34.6\% \\
mimo-v2.5-pro & 0.1682 / 40.6\% & 0.0819 / 30.0\% & 0.0796 / 29.8\% \\
\bottomrule
\end{tabular}
\end{center}
\FloatBarrier

\Needspace{22\baselineskip}
\begin{center}
\captionof{table}{P3 split-resolved table. Each cell reports final recovered feasible rate and post-escape recovery quality.}
\label{tab:appendix_p3_split}
\centering
\scriptsize
\setlength{\tabcolsep}{4pt}
\renewcommand{\arraystretch}{1.08}
\begin{tabular}{lccc}
\toprule
Model & dev & test\_id & test\_ood \\
\midrule
hunyuan-hy3-preview & 50.0\% / 0.4811 & 42.5\% / 0.5333 & 48.1\% / 0.4605 \\
mimo-v2.5-pro & 50.0\% / 0.6591 & 40.0\% / 0.5435 & 44.2\% / 0.6406 \\
deepseek-r1 & 45.3\% / 0.3103 & 37.5\% / 0.2105 & 44.2\% / 0.2708 \\
gpt-5.4 & 48.4\% / 0.4322 & 40.0\% / 0.3816 & 36.5\% / 0.2841 \\
gemini-3.1-pro-preview & 43.8\% / 0.3770 & 32.5\% / 0.2917 & 32.7\% / 0.2660 \\
deepseek-v3 & 42.2\% / 0.2368 & 27.5\% / 0.3571 & 32.7\% / 0.3068 \\
qwen3-max & 39.1\% / 0.3182 & 22.5\% / 0.1912 & 25.0\% / 0.2444 \\
deepseek-v4-pro & 37.5\% / 0.2456 & 25.0\% / 0.1714 & 21.2\% / 0.1633 \\
\texttt{qwen3.6-plus} & 28.1\% / 0.1774 & 27.5\% / 0.1389 & 26.9\% / 0.1100 \\
\texttt{o4-mini} & 28.1\% / 0.1556 & 30.0\% / 0.1667 & 21.2\% / 0.0972 \\
claude-4.6-sonnet & 20.3\% / 0.1270 & 10.0\% / 0.0513 & 15.4\% / 0.0918 \\
\texttt{llama-3.3-70b} & 4.7\% / 0.1429 & 0.0\% / 0.0000 & 3.8\% / 0.1429 \\
\bottomrule
\end{tabular}
\end{center}
\FloatBarrier

\Needspace{22\baselineskip}
\begin{center}
\captionof{table}{P4 split-resolved table. Each cell reports Full Tau and Exact.}
\label{tab:appendix_p4_split}
\centering
\scriptsize
\setlength{\tabcolsep}{4pt}
\renewcommand{\arraystretch}{1.08}
\begin{tabular}{lccc}
\toprule
Model & dev & test\_id & test\_ood \\
\midrule
gpt-5.4 & 0.886 / 53.8\% & 0.900 / 58.3\% & 0.873 / 66.7\% \\
\texttt{qwen3.6-plus} & 0.867 / 65.6\% & 0.856 / 52.8\% & 0.920 / 80.0\% \\
deepseek-v3 & 0.888 / 57.0\% & 0.783 / 47.2\% & 0.867 / 56.7\% \\
mimo-v2.5-pro & 0.852 / 51.6\% & 0.794 / 44.4\% & 0.873 / 66.7\% \\
claude-4.6-sonnet & 0.832 / 59.1\% & 0.833 / 41.7\% & 0.873 / 66.7\% \\
hunyuan-hy3-preview & 0.856 / 51.6\% & 0.767 / 41.7\% & 0.873 / 60.0\% \\
qwen3-max & 0.832 / 49.5\% & 0.850 / 47.2\% & 0.827 / 53.3\% \\
deepseek-r1 & 0.834 / 57.0\% & 0.800 / 44.4\% & 0.867 / 66.7\% \\
gemini-3.1-pro-preview & 0.826 / 55.9\% & 0.778 / 38.9\% & 0.873 / 66.7\% \\
deepseek-v4-pro & 0.804 / 44.1\% & 0.772 / 38.9\% & 0.787 / 46.7\% \\
\texttt{o4-mini} & 0.778 / 47.3\% & 0.733 / 44.4\% & 0.800 / 56.7\% \\
\texttt{llama-3.3-70b} & 0.720 / 35.5\% & 0.667 / 22.2\% & 0.753 / 43.3\% \\
\bottomrule
\end{tabular}
\end{center}
\FloatBarrier

\section{Reasoning Coverage and Response-Control Profiles}

This block remains secondary evidence rather than a substitute for the model-by-model tables. It is a negative control for the response-control profile account: if the paper were only measuring whether thinking mode was on, the stage patterns should collapse under this grouping. Instead, thinking mode helps some workflow regimes more than others and still does not erase stage dissociation.

\clearpage
\begin{table}[p]
\centering
\caption{Reasoning/thinking run-mode audit for the 12 complete P1--P4 model runs. The column records whether thinking mode was used in the reported run.}
\label{tab:appendix_reasoning_grouping}
\scriptsize
\setlength{\tabcolsep}{3.2pt}
\renewcommand{\arraystretch}{1.08}
\begin{tabularx}{\linewidth}{@{}>{\raggedright\arraybackslash}p{26mm}Y>{\centering\arraybackslash}p{13mm}Y@{}}
\toprule
Model & Public / run evidence & Think used & Caveat \\
\midrule
\texttt{o4-mini} & OpenAI documentation identifies it as a reasoning model~\citep{openai2026o4mini}. & Yes & reasoning-labeled row \\
gemini-3.1-pro-preview & Google documentation describes internal thinking / thought-summary support~\citep{google2026geminithinking}. & Yes & thinking support active in the reported run \\
deepseek-r1 & DeepSeek documentation positions R1 as a reasoning-oriented model~\citep{deepseek2025r1}. & Yes & key within-family contrast is R1 versus V3 \\
hunyuan-hy3-preview & Public provider pages describe Hy3 as a fast/slow or multi-mode reasoning model~\citep{siliconflow2026hy3preview}. & Yes & run manifests are retained as the source of truth for provider settings \\
deepseek-v4-pro & DeepSeek documentation describes V4 as supporting Thinking and Non-Thinking modes~\citep{deepseek2026v4preview}. & No & manifest records \texttt{thinking=disabled} for JSON/action-schema stability \\
mimo-v2.5-pro & Xiaomi Mimo API row is treated as a thinking-capable provider model in the run plan. & No & manifest records \texttt{thinking=disabled} for JSON/action-schema stability \\
qwen3-max & Alibaba documentation describes Qwen3-Max as hybrid-thinking with thinking disabled by default~\citep{alibaba2026qwenDeepThinking}. & No & DashScope call used provider default; no explicit \texttt{enable\_thinking=True} override \\
\texttt{qwen3.6-plus} & Qwen documentation describes explicit \texttt{enable\_thinking=True} usage~\citep{qwen2026qwen36plus}. & No & DashScope call used provider default without an explicit \texttt{enable\_thinking} flag \\
gpt-5.4 & Frontier GPT row with provider-default reasoning effort in our logs. & No & reported as no explicit thinking-mode run \\
deepseek-v3 & No reasoning-first grouping used here. & No & used as the within-family non-R1 contrast \\
claude-4.6-sonnet & No reasoning-first grouping used here. & No & grouped by observed profile, not by hidden implementation details \\
\texttt{llama-3.3-70b} & No reasoning-first grouping used here. & No & open model baseline row in the complete P1--P4 roster \\
\bottomrule
\end{tabularx}
\end{table}
\FloatBarrier

\begin{center}
\captionof{table}{Stage-wise group means on the 12 complete model runs, grouped only by whether thinking mode was used in the reported run.}
\label{tab:appendix_reasoning_groupmeans}
\centering
\scriptsize
\setlength{\tabcolsep}{5pt}
\renewcommand{\arraystretch}{1.08}
\begin{tabular}{lcc}
\toprule
Stage & Think & No-think \\
\midrule
P1 Composite & 0.499 & 0.359 \\
P2 headline & 0.237 & 0.161 \\
P3 Success (\%) & 38.5 & 28.6 \\
P4 Full Tau & 0.819 & 0.831 \\
\bottomrule
\end{tabular}
\end{center}
\FloatBarrier

\clearpage
\Needspace{30\baselineskip}
\Needspace{30\baselineskip}
\noindent\begin{minipage}{\textwidth}
\centering
\captionof{table}{Stable profile scores on the 12 complete P1--P4 runs. Scores are diagnostic boundary-fit indicators, not a monolithic leaderboard. Column aliases: Action=action discipline, Edit=edit style, Feedback=feedback conditioning, State=state-reset effort, Policy=policy execution.}
\label{tab:profile_quantification}
\scriptsize
\setlength{\tabcolsep}{4pt}
\renewcommand{\arraystretch}{1.08}
\begin{tabular}{lccccc}
\toprule
Model & Action & Edit & Feedback & State & Policy \\
\midrule
qwen3-max & 0.737 & 0.571 & 0.407 & 0.646 & 0.679 \\
gemini-3.1-pro-preview & 0.668 & 0.753 & 0.508 & 0.425 & 0.729 \\
gpt-5.4 & 0.657 & 0.482 & 0.406 & 0.453 & 0.743 \\
deepseek-r1 & 0.652 & 0.667 & 0.455 & 0.455 & 0.728 \\
\texttt{o4-mini} & 0.635 & 0.470 & 0.391 & 0.495 & 0.666 \\
hunyuan-hy3-preview & 0.564 & 0.485 & 0.429 & 0.599 & 0.688 \\
deepseek-v3 & 0.536 & 0.493 & 0.389 & 0.629 & 0.708 \\
\texttt{llama-3.3-70b} & 0.528 & 0.586 & 0.326 & 0.452 & 0.622 \\
mimo-v2.5-pro & 0.499 & 0.385 & 0.391 & 0.627 & 0.707 \\
deepseek-v4-pro & 0.491 & 0.404 & 0.332 & 0.737 & 0.666 \\
qwen3.6-plus & 0.491 & 0.613 & 0.404 & 0.523 & 0.761 \\
claude-4.6-sonnet & 0.428 & 0.622 & 0.392 & 0.443 & 0.730 \\
\bottomrule
\end{tabular}
\end{minipage}
\FloatBarrier

\clearpage
\section{Stage-Aware Selection and Error Decomposition}
\label{sec:appendix_stage_awareness}

This appendix reports the two analyses used in Section~\ref{sec:experiments} to connect stage-local scores to actionable model selection and visible failure modes. Both analyses reuse existing split scores and oracle-log profile metrics; no new model calls are introduced.

\subsection{Stage-Rank Correlation Matrix}
\label{sec:appendix_stage_rank_corr}

Table~\ref{tab:appendix_stage_rank_corr} reports the full Spearman rank-correlation matrix among the four stage leaderboards. The off-diagonal correlations range from \(-0.26\) to \(0.38\), showing that no single stage ordering is a reliable proxy for the others.

\begin{center}
\captionof{table}{Spearman rank correlations between stage leaderboards.}
\label{tab:appendix_stage_rank_corr}
\centering
\small
\setlength{\tabcolsep}{5pt}
\renewcommand{\arraystretch}{1.08}
\begin{tabular}{lrrrr}
\toprule
 & P1 & P2 & P3 & P4 \\
\midrule
P1 & 1.00 & 0.29 & 0.31 & -0.26 \\
P2 & 0.29 & 1.00 & 0.06 & 0.07 \\
P3 & 0.31 & 0.06 & 1.00 & 0.38 \\
P4 & -0.26 & 0.07 & 0.38 & 1.00 \\
\bottomrule
\end{tabular}
\end{center}
\FloatBarrier

\subsection{Stage-Aware Selection Simulation}
\label{sec:appendix_stage_router}

The router simulation uses the 12 complete P1--P4 model runs. Model selection is performed on the validation split. Held-out evaluation combines \texttt{test\_id} and \texttt{test\_ood} with task-count weights: P1 80/80, P2 40/104, P3 40/52, and P4 36/30. Stage scores are normalized by the held-out best score for that stage before averaging. Validation ties are broken by the corresponding validation auxiliary metric: P1 accuracy, P2 final feasible rate, P3 recovery quality, and P4 exact match.

\begin{center}
\captionof{table}{Full stage-aware selection simulation. The selected model column records the model used for each stage.}
\label{tab:appendix_stage_router_full}
\centering
\scriptsize
\setlength{\tabcolsep}{3pt}
\renewcommand{\arraystretch}{1.08}
\begin{tabularx}{\textwidth}{lYYYYr}
\toprule
Strategy & P1 & P2 & P3 & P4 & Mean norm. \\
\midrule
Validation aggregate leader & 0.552 (gemini-3.1-pro-preview) & 0.385 (gemini-3.1-pro-preview) & 32.6\% (gemini-3.1-pro-preview) & 0.821 (gemini-3.1-pro-preview) & 0.892 \\
Best held-out single model & 0.552 (gemini-3.1-pro-preview) & 0.385 (gemini-3.1-pro-preview) & 32.6\% (gemini-3.1-pro-preview) & 0.821 (gemini-3.1-pro-preview) & 0.892 \\
Validation-selected stage router & 0.552 (gemini-3.1-pro-preview) & 0.385 (gemini-3.1-pro-preview) & 42.4\% (mimo-v2.5-pro) & 0.821 (deepseek-v3) & 0.945 \\
Ex-post stage-best envelope & 0.594 (qwen3-max) & 0.385 (gemini-3.1-pro-preview) & 45.7\% (hunyuan-hy3-preview) & 0.888 (gpt-5.4) & 1.000 \\
\bottomrule
\end{tabularx}
\end{center}
\FloatBarrier

The validation-selected stage router improves over the validation aggregate leader because it assigns different models to different design roles instead of forcing one model to cover the full workflow. In the full 12-model pool, the aggregate validation leader and the best held-out single model are both gemini-3.1-pro-preview, with a mean normalized held-out score of 0.892. This equality is not a duplicate-row error: gemini-3.1-pro-preview is both the validation aggregate choice and the ex-post best single model under the held-out pool. The validation-selected router keeps gemini-3.1-pro-preview for P1 and P2, switches P3 to mimo-v2.5-pro, and selects deepseek-v3 for P4; this raises mean normalized held-out score to 0.945. The realized gain comes almost entirely from P3, where held-out recovery improves from 32.6\% under the aggregate leader to 42.4\% under the routed choice. The router does not reach the ex-post stage-best envelope because validation selection misses the held-out P1 leader (qwen3-max), the held-out P3 leader (hunyuan-hy3-preview), and the held-out P4 leader (gpt-5.4). This is the desired diagnostic behavior: the router is deployable because it uses only validation information, while the envelope reports remaining headroom.

\paragraph{Worked normalization example.}
Table~\ref{tab:stage_aware_router} reports raw held-out stage scores and a normalized mean. For qwen3-max, the raw held-out scores are \(0.594\) on P1, \(0.185\) on P2, \(23.9\%\) on P3, and \(0.839\) on P4. The held-out stage-best denominators are \(0.594\), \(0.385\), \(45.7\%\), and \(0.888\), respectively. Its mean normalized score is therefore
\[
\frac{1}{4}\left(\frac{0.594}{0.594}+\frac{0.185}{0.385}+\frac{0.239}{0.457}+\frac{0.839}{0.888}\right)=0.738.
\]
The ex-post stage-best envelope has mean normalized score \(1.000\) by construction because each stage uses its own held-out best model as the denominator.

\subsection{Stage-Wise Error Decomposition}
\label{sec:appendix_stage_error_decomposition}

Table~\ref{tab:appendix_stage_error_decomp} expands the compact main-text decomposition to all 12 full-coverage models. The error families are non-exclusive diagnostic rates; they are not a partition of all failures and should not be summed within a row. The exact column formulas are given in Appendix~\ref{sec:appendix_fulltable_metric_dictionary}. P2 invalid/no-op uses protocol-invalid rate as the stable released proxy; raw identical-edit no-op extraction is not required for the reported scores.

\begin{center}
\captionof{table}{Stage-wise error decomposition by model. All values are rates in [0,1]; lower is better.}
\label{tab:appendix_stage_error_decomp}
\centering
\scriptsize
\setlength{\tabcolsep}{2.2pt}
\renewcommand{\arraystretch}{1.04}
\resizebox{\textwidth}{!}{%
\begin{tabular}{lrrrrrrrrrrrrrr}
\toprule
Model & P1 over-act & P1 over-ref. & P1 miss & P2 no close & P2 destr. & P2 inv. & P2 util. loss & P3 no escape & P3 casc. & P3 dead & P3 post-esc. fail & P4 policy & P4 dom. & P4 parse \\
\midrule
qwen3-max & 0.390 & 0.233 & 0.467 & 0.442 & 0.000 & 0.000 & 0.646 & 0.141 & 0.127 & 0.257 & 0.649 & 0.701 & 0.008 & 0.000 \\
gemini-3.1-pro-preview & 0.524 & 0.067 & 0.333 & 0.038 & 0.000 & 0.000 & 0.590 & 0.077 & 0.507 & 0.258 & 0.597 & 0.711 & 0.013 & 0.013 \\
\texttt{o4-mini} & 0.600 & 0.027 & 0.600 & 0.572 & 0.000 & 0.005 & 0.634 & 0.308 & 0.176 & 0.266 & 0.620 & 0.879 & 0.000 & 0.000 \\
deepseek-r1 & 0.581 & 0.088 & 0.444 & 0.322 & 0.000 & 0.000 & 0.642 & 0.077 & 0.264 & 0.251 & 0.535 & 0.719 & 0.000 & 0.000 \\
gpt-5.4 & 0.343 & 0.552 & 0.311 & 0.639 & 0.000 & 0.000 & 0.616 & 0.096 & 0.213 & 0.252 & 0.532 & 0.566 & 0.000 & 0.000 \\
hunyuan-hy3-preview & 0.705 & 0.000 & 0.667 & 0.514 & 0.000 & 0.005 & 0.662 & 0.224 & 0.091 & 0.253 & 0.388 & 0.657 & 0.028 & 0.000 \\
deepseek-v3 & 0.790 & 0.010 & 0.644 & 0.558 & 0.000 & 0.000 & 0.642 & 0.173 & 0.271 & 0.276 & 0.574 & 0.617 & 0.008 & 0.000 \\
\texttt{llama-3.3-70b} & 0.752 & 0.039 & 0.800 & 0.630 & 0.000 & 0.000 & 0.672 & 0.654 & 0.037 & 0.143 & 0.907 & 0.851 & 0.020 & 0.000 \\
mimo-v2.5-pro & 0.848 & 0.000 & 0.756 & 0.668 & 0.000 & 0.000 & 0.672 & 0.365 & 0.040 & 0.267 & 0.283 & 0.657 & 0.003 & 0.000 \\
\texttt{qwen3.6-plus} & 0.857 & 0.000 & 0.800 & 0.389 & 0.000 & 0.000 & 0.657 & 0.051 & 0.230 & 0.256 & 0.709 & 0.661 & 0.000 & 0.000 \\
deepseek-v4-pro & 0.857 & 0.000 & 0.800 & 0.587 & 0.000 & 0.000 & 0.680 & 0.096 & 0.170 & 0.243 & 0.681 & 0.783 & 0.012 & 0.000 \\
claude-4.6-sonnet & 0.943 & 0.000 & 1.000 & 0.375 & 0.000 & 0.005 & 0.611 & 0.032 & 0.053 & 0.257 & 0.834 & 0.710 & 0.000 & 0.000 \\
\bottomrule
\end{tabular}%
}
\end{center}
\FloatBarrier

These error families explain what the response-control profiles summarize: action discipline appears as over-action or over-refusal; bounded editing and feedback conditioning appear as infeasible closure, destructive edits, invalid updates, and utility loss; state-reset effort appears as escape, cascade, dead-budget, and post-escape failures; and policy execution appears as policy mismatch, dominance error, parse failure, and top-choice failure.

\noindent\textbf{Readout.} The main text reports the cross-model means and interpretation. This table provides the per-model values used to identify which failure family drives each stage score and to support the profile readout.

\clearpage
\section{Extended Related Work}
\label{sec:appendix_extended_related_work}

\begin{center}
\centering
\footnotesize
\setlength{\tabcolsep}{3.5pt}
\renewcommand{\arraystretch}{1.12}
\captionof{table}{Positioning VEHBench among LLM-assisted engineering design benchmarks.}
\label{tab:appendix_vehbench_positioning}
\begin{tabularx}{\textwidth}{p{0.19\textwidth}YYYY}
\toprule
\textbf{Benchmark / line of work} & \textbf{What it evaluates} & \textbf{How validity is checked} & \textbf{What remains hard to see} & \textbf{VEHBench distinction} \\
\midrule
Broad engineering design benchmarks & General engineering design capability & Rubric / task-specific checks & Which design-stage behavior failed & Stage-local workflow diagnosis \\
Artifact / simulation benchmarks & Whether generated artifacts work & Simulator / physical feedback & Whether failure came from triage, search, recovery, or selection & Verifier used as workflow boundary \\
Mechanical-design agents & Iterative mechanical design or CAD/CAE reasoning & Simulation or expert/problem checks & Cross-stage role dissociation & Separate assistant roles \\
Engineering QA / documentation benchmarks & Understanding engineering documents & QA labels / expert answers & Design-loop behavior & Moves from comprehension to design assistance \\
Materials / selection benchmarks & Choosing suitable materials or options & Labels / property rules & Search/recovery behavior before selection & Separates generation, recovery, and selection \\
\textbf{VEHBench} & LLM-assisted VEH co-design & Analytical physical oracle & N/A & Engineering-native diagnostic benchmark \\
\bottomrule
\end{tabularx}
\end{center}
\FloatBarrier

VEHBench sits between engineering-design benchmarks and benchmark-validity work. General agent and software benchmarks are useful context for endpoint task completion, but they are not the primary comparators for this paper because VEHBench targets coupled physical design rather than coding, web navigation, or generic tool use. The relevant comparison is therefore not whether an agent completes a broad interactive task, but whether an engineering assistant behaves reliably at a specific verifier-grounded design boundary: pre-search triage, verifier-guided design search, corrupted-state recovery, or policy-conditioned selection.

Appendix Table~\ref{tab:appendix_vehbench_positioning} gives the benchmark-positioning summary; this appendix expands the surrounding context.

A growing line of work evaluates LLMs on engineering design tasks that require external physical or simulation-based verification. EngDesign, BuildArena, mechanical-design agents, DesignQA, and MSEval evaluate design satisfaction, simulation loops, document reasoning, or material-selection behavior~\citep{guo2025engdesign,xia2025buildarena,jadhav2024mechanical,doris2024designqa,jain2024mseval}. VEHBench differs by treating design as a sequence of decision regimes rather than one endpoint objective. Its VEH domain is not arbitrary: vibration and piezoelectric energy harvesting is a canonical small-scale power setting for wireless and embedded systems, from early micro-generator analysis to modern cantilever-based, mechanical, and IoT-oriented harvesting surveys~\citep{williams1996microelectric,safaei2019review,maamer2019review,sadaf2024cantilever,sadaf2025harnessing,citroni2024efficient,zeadally2020design,naifar2024energy}. Cantilevered piezoelectric models provide the closed-form electromechanical structure used by the oracle~\citep{erturk2008bimorph,erturk2009experimental}, and high-impact nanogenerator work further established piezoelectric transduction as a route to self-powered microsystems~\citep{wang2006nanogenerators,xu2010nanowire}. This makes VEH a useful closed-form engineering domain: it is physically meaningful, constraint-coupled, and analytically auditable.

VEHBench also builds on benchmark design and evaluation-validity work. BIG-bench, HELM, Dynabench, datasheets, model cards, Croissant, and benchmark-accountability work motivate transparent task construction, metadata, contamination boundaries, and reproducibility claims~\citep{srivastava2022bigbench,liang2023helm,kiela2021dynabench,gebru2021datasheets,mitchell2019modelcards,akhtar2024croissant,raji2021aiaccountability}. LLM-as-judge and prompt-sensitivity studies caution that evaluation can fail when the judge is comparable to the tested model or when models recognize evaluation contexts~\citep{dorner2025limits,xiong2025stealtheval,pu2025overbench}. VEHBench responds by using an external physical oracle and by reporting a controlled-prompt audit rather than relying on conversational closure or LLM judgment.

Finally, VEHBench is adjacent to black-box optimization, but not reducible to it. COCO/BBOB and CMA-ES provide optimization context and baselines~\citep{hansen2021coco,hansen2023cma}. VEHBench's target is broader: numerical search matters in P2, but P1 tests actionability, P3 tests recovery after state contamination, and P4 tests policy-conditioned selection.

\clearpage
\section{CMA-ES Classical Optimization Baseline}
\label{sec:appendix_cmaes}

CMA-ES~\citep{hansen2023cma} is run under the same analytical oracle as the LLM agents, with population size 8 and initial sigma 0.15. The 6-query run covers the released split-resolved P2 subset used for this optimizer audit (156 tasks), while the relaxed 40-query run is a dev-subset ceiling check (64 tasks). These runs are therefore calibration references rather than full-bank competitors to the model results in Table~\ref{tab:main_results}.

\begin{center}
\captionof{table}{CMA-ES P2 baseline results. The 6-query budget matches the LLM agents on the 156-task optimizer-audit subset; 40 queries tests a 64-task dev-subset optimizer ceiling. The anchor-fixed heuristic (0 oracle queries) and gemini-3.1-pro-preview (2.63 avg.\ queries) are shown for context.}
\label{tab:appendix_cmaes_p2}
\centering
\small
\begin{tabular}{lcccc}
\toprule
Method & Budget (queries) & Final feasible rate & Mean power ratio & Mean queries \\
\midrule
CMA-ES & 6 & 0.269 & 0.099 & 6.0 \\
CMA-ES & 40 & 0.578 & 0.305 & 40.0 \\
Anchor-fixed heuristic & 0 & 1.000 & 0.306 & 0.0 \\
gemini-3.1-pro-preview (LLM) & 6 & 0.962 & 0.390 & 2.63 \\
\bottomrule
\end{tabular}
\end{center}
\FloatBarrier

CMA-ES under 6 queries underperforms most frontier LLM agents on P2. With 40 queries on the dev subset it matches the anchor-fixed heuristic (0.305 vs.\ 0.306) but remains below gemini-3.1-pro-preview (0.390). This supports two complementary interpretations: (1) LLM agents carry physical priors from pretraining that accelerate feasibility-preserving search beyond what a pure numerical optimizer achieves under the same oracle budget; and (2) P2 is not explained away by a generic optimizer within the benchmark's query constraints, so the observed inter-model differences reflect genuine behavioral variation rather than a ceiling effect.

\clearpage
\section{Prompt Templates, Oracle Protocol, and Task Examples}
\label{sec:prompts}

\subsection{Evaluation Harness Protocol}

All model runs use a shared evaluation harness. Each task instance specifies the design variables, variable bounds, task objective, and query budget. The harness parses the first valid JSON object in each model response, normalizes field aliases (e.g., \texttt{R\_ohm} vs.\ \texttt{R\_value}), and applies a fixed retry policy: up to two retries for transient malformed or length-truncated outputs before counting the row as a parse failure. Iterative probes (P2, P3) run under bounded query budgets. P2 can stop on feasible closure; P3 preserves the full recovery trace. All runs use temperature 0.0, and provider-level default max output tokens (typically 4096--8192 depending on the provider). Run dates, provider endpoints, and model version strings are recorded in per-model run manifests released with the benchmark artifact.

\subsection{P1 Prompt Template (Triage)}

The P1 prompt presents the task specification and asks the model to output exactly one action.

\begin{promptlisting}{P1 Prompt: Specification Triage}
System:
You are an engineering triage agent. Given a design specification, choose whether the task is feasible, infeasible, or underspecified. Do not output numeric design variables; output one action label as a single JSON object.

User:
{
  "task": "Evaluate whether the following VEH design specification
           is feasible, infeasible, or missing critical information.",
  "spec": {
    "beam_length_mm": [20, 80],
    "tip_mass_g": [0.5, 5.0],
    "substrate_thickness_um": [100, 500],
    "piezo_thickness_um": [50, 300],
    "target_frequency_hz": 50,
    "target_power_uw": 15,
    "excitation_acceleration_g": 0.5,
    "material": "PZT-5A",
    "substrate": "stainless_steel"
  },
  "allowed_actions": ["propose_design", "declare_infeasible",
                      "request_missing_info"]
}

Expected output format:
{
  "action": "propose_design | declare_infeasible | request_missing_info",
  "reason": "<brief justification>",
  "missing_fields": ["spec.xxx", ...]
}
\end{promptlisting}

The \texttt{missing\_fields} array is only required for \texttt{request\_missing\_info}. P1 evaluator scoring is described in Appendix~\ref{sec:metric_formulas}.

\clearpage
\subsection{P2 Prompt Template (Verifier-Guided Design Search)}

P2 runs as a multi-turn loop. Each turn presents the current design state and the oracle's constraint feedback.

\begin{promptlisting}{P2 Prompt: Verifier-Guided Design Search}
System:
You are an engineering design agent. You are given a design task with coupled physical constraints. At each step, propose an improved design. The verifier will return feasibility and violation information. Your goal is to reach a feasible design within the query budget while maximizing the objective.

User (turn 1):
{
  "task": "Design a piezoelectric cantilever VEH meeting the
           following constraints.",
  "design_variables": ["beam_length_mm", "tip_mass_g",
       "substrate_thickness_um", "piezo_thickness_um",
       "beam_width_mm", "load_resistance_ohm"],
  "variable_bounds": {
    "beam_length_mm": [20, 80],
    "tip_mass_g": [0.5, 5.0],
    "substrate_thickness_um": [100, 500],
    "piezo_thickness_um": [50, 300],
    "beam_width_mm": [5, 30],
    "load_resistance_ohm": [1000, 100000]
  },
  "initial_design": {
    "beam_length_mm": 50, "tip_mass_g": 2.0,
    "substrate_thickness_um": 300, "piezo_thickness_um": 150,
    "beam_width_mm": 15, "load_resistance_ohm": 10000
  },
  "objective": "maximize output power (uw)",
  "query_budget": 6
}

Oracle response (example, after turn 1):
{
  "feasible": false,
  "metrics": {
    "resonant_freq_hz": 72.3,
    "load_power_uw": 3.2,
    "tip_stress_mpa": 12.4,
    "tip_disp_mm": 0.8
  },
  "violations": [
    {"constraint": "frequency_target", "value": 72.3,
     "target": 50.0, "tolerance": 0.05,
     "message": "frequency too high; increase mass or length"}
  ],
  "total_violation": 8.47,
  "oracle_feedback": "frequency=72.3 Hz exceeds [47.5, 52.5] Hz
      target; tip stress=12.4 MPa within limit."
}

Model output format (each turn):
{
  "design": {
    "beam_length_mm": 55.0,
    "tip_mass_g": 2.5,
    "substrate_thickness_um": 280,
    "piezo_thickness_um": 150,
    "beam_width_mm": 15,
    "load_resistance_ohm": 10000
  },
  "action": "continue",
  "reason": "Increased beam length and tip mass to reduce
             resonant frequency toward 50 Hz target."
}
\end{promptlisting}

\clearpage
\subsection{P3 Prompt Template (Post-Trap Recovery)}

P3 is structurally identical to P2 but pre-loads a corrupted trajectory into the conversation context. The corruption varies by trap type: unit-flip, wrong-formula-direction, false-feasibility, topology-trap, verifier-ignored, or progressive-contamination. The model receives the full pseudo-history as if it were prior conversation turns.

\begin{promptlisting}{P3 Prompt: Corrupted-State Recovery}
System:
You are an engineering recovery agent. You are given a design task, a prior trajectory, and verifier feedback. Some earlier trajectory steps may be corrupted or physically misleading. Your goal is to recover a feasible design within the remaining query budget. You may reuse a trusted prior design, reset to an earlier state, replan, or propose a new bounded edit, but the final response must be a single valid JSON object.

User:
{
  "task": "Recover a feasible VEH design from the following trajectory.",
  "design_variables": ["beam_length_mm", "tip_mass_g",
       "substrate_thickness_um", "piezo_thickness_um",
       "beam_width_mm", "load_resistance_ohm"],
  "variable_bounds": {...},
  "objective": "maximize output power (uw) subject to verifier constraints",
  "query_budget_remaining": 4,
  "corrupted_history": [
    {
      "step": 0,
      "design": {...},
      "verifier_response": {...},
      "note": "prior assistant claimed frequency was too low"
    },
    {
      "step": 1,
      "design": {...},
      "verifier_response": {
        "feasible": false,
        "oracle_feedback": "frequency is too high; the prior direction
                            increased violation"
      }
    }
  ],
  "allowed_actions": ["propose_design", "replan", "reset_to_prior",
                      "declare_infeasible"]
}

Model output format:
{
  "action": "propose_design | replan | reset_to_prior | declare_infeasible",
  "design": {
    "beam_length_mm": 60.0,
    "tip_mass_g": 2.8,
    "substrate_thickness_um": 260,
    "piezo_thickness_um": 150,
    "beam_width_mm": 15,
    "load_resistance_ohm": 10000
  },
  "trusted_state_decision": "discard step 1 because verifier feedback
                             contradicts the claimed update direction",
  "reason": "Reset from the corrupted direction and reduce frequency
             without introducing stress or displacement violations."
}
\end{promptlisting}

For the state-summary intervention, the raw trajectory is replaced with a deterministic verifier-authored summary containing only already-observed trusted fields: current step, latest proposal, latest verifier state, best-so-far feasible proposal (if any), short-horizon objective and violation trends, and whether new violations were introduced. The oracle evaluator is unchanged.

\clearpage
\subsection{P4 Prompt Template (Policy-Conditioned Ranking)}

\begin{promptlisting}{P4 Prompt: Policy-Conditioned Selection}
System:
You are an engineering evaluator. Given a set of oracle-feasible design candidates and a stated ranking policy, rank the candidates according to the policy. Do not modify the designs or propose new ones.

User:
{
  "task": "Rank the following feasible VEH designs according to
           the stated policy.",
  "policy": "Prioritize output power, then tip-stress margin,
             then component cost. All candidates are feasible.",
  "candidates": [
    {"id": "A", "design": {...}},
    {"id": "B", "design": {...}},
    {"id": "C", "design": {...}},
    {"id": "D", "design": {...}},
    {"id": "E", "design": {...}}
  ]
}

Expected output:
{
  "ranking": ["B", "A", "D", "E", "C"],
  "reason": "B has highest power; A second on power with better
             stress margin than D; E and C follow by power."
}
\end{promptlisting}

\subsection{Oracle Specification (VEH Domain)}

The VEH oracle follows the single-mode piezoelectric cantilever model defined in Appendix~\ref{sec:appendix_veh_oracle_formulas}. That section gives the complete symbol table, composite-section calculation, electromechanical coupling, dynamic frequency-response displacement, average load-power formula, root-stress approximation, and feasibility slacks. We do not repeat a shortened formula list here because the prompt templates use oracle feedback fields rather than asking the model to execute the equations manually. The release includes the oracle code, task manifests, and validation artifacts needed to audit these computations.

\subsection{Example Task Walkthrough (VEH P2)}

We illustrate a complete P2 trajectory for a single task to make the evaluation protocol concrete.

\textbf{Task.} Design a PZT-5A cantilever VEH with target frequency 50 Hz, target power $\ge 15\,\mu$W, excitation 0.5 g, stress limit 50 MPa, displacement limit 2 mm. Budget: 6 queries. Initial design: $L = 50$ mm, $m_t = 2.0$ g, $h_s = 300\,\mu$m, $h_p = 150\,\mu$m, $w = 15$ mm, $R_L = 10$ k$\Omega$.

\textbf{Step 1.} Model proposes $\{L: 55, m_t: 2.5, h_s: 280, h_p: 150, w: 15, R_L: 10000\}$. Oracle returns: $f_r = 63.8$ Hz (violation: too high), $P_{\text{out}} = 4.1\,\mu$W, stress 11.2 MPa, displacement 0.7 mm. Feasible: false. Total violation: 5.24.

\textbf{Step 2.} Model proposes $\{L: 65, m_t: 3.0, h_s: 250, h_p: 120, w: 15, R_L: 10000\}$. Oracle returns: $f_r = 48.7$ Hz (within $[47.5, 52.5]$), $P_{\text{out}} = 9.8\,\mu$W, stress 9.1 MPa, displacement 0.8 mm. Feasible: true. Violation: 0. Objective: $9.8/15 = 0.653$ (relative to BKF). End of trajectory (feasible closure).

This task is considered a P2 success. The model used 2 queries, achieved feasible closure, and the final feasible power ratio is 0.653.

\subsection{Example Task Walkthrough (Circuit P3 Dual-Trap)}

\textbf{Task.} RC low-pass filter, target $f_c = 1$ kHz $\pm 2\%$, source current limit 0.5 mA, $R \in [1\text{k}\Omega, 100\text{k}\Omega]$, $C \in [1\text{nF}, 1\mu\text{F}]$, $V_{\text{in}} = 5$ V. Corrupted history (phase-1 trap): prior steps repeatedly increased $C$ under the false assumption that $f_c$ was too low; actual $f_c = 3.18$ kHz (too high). Current bad design: $R = 5\text{k}\Omega$, $C = 100$ nF. Budget: 5 queries.

\textbf{Step 1.} Model recognizes the contradiction between oracle feedback (``$f_c$ too high'') and corrupted history (``$f_c$ too low''). Proposes $\{R: 20000, C: 7.96\text{nF}\}$. Oracle: $f_c = 1000$ Hz (feasible), $I_{\text{source}} = 0.25$ mA (feasible). Phase 1 escaped.

\textbf{Step 2 (phase-2 trap).} With $R = 20\text{k}\Omega$, the model's edit has triggered the near-boundary margin: a second constraint on component tolerance (previously dormant) now requires $C \ge 1.5$ nF. Oracle returns a new violation. Model must adjust $R$ downward slightly and $C$ upward.

\textbf{Step 3.} Model proposes $\{R: 15000, C: 10.6\text{nF}\}$. Oracle: $f_c = 1001$ Hz (feasible), $I_{\text{source}} = 0.33$ mA (feasible). All constraints satisfied. Recovery success: true. Cascade: false.

This task demonstrates the progressive-dual-trap mechanism: the model must escape the primary corrupted state (phase 1) and then handle a secondary coupled constraint triggered by the escape move itself (phase 2).

\subsection{Run Configuration}

All model runs share the following configuration unless noted otherwise in per-model manifests:

\begin{itemize}[leftmargin=5mm,itemsep=1mm,topsep=1mm]
    \item Temperature: 0.0 (deterministic decoding).
    \item Max output tokens: provider default (4096--8192).
    \item Retry policy: up to 2 retries on parse failure; row counted as parse error after 3 consecutive failures.
    \item P2 stop rule: trajectory terminates on feasible closure or budget exhaustion.
    \item P3 stop rule: trajectory runs full budget; no early stop on feasibility.
    \item Provider endpoints and model version strings are recorded in per-model run manifests.
    \item Core roster run dates: April 2026. Thinking-model extension run dates: April 25--27, 2026. Circuit audit run dates: April 28--30, 2026.
\end{itemize}

\clearpage
\section{Representative Failure Cases}
\label{sec:failure_cases}

The following notes give one auditable failure case per probe. They are not additional metrics; they make the response-control interpretation concrete and show what the probe-specific scores count as failure. We use prose rather than a table because the cases contain long task identifiers and multi-sentence readouts.

\paragraph{P1: optimistic over-proposal.}
In audit case \path{p1::dev::infeasible_hard_conflict::0000::s42}, the gold action is \texttt{declare\_infeasible}, but the model chooses \texttt{propose\_design}. The verifier-side label makes the failure observable: continuation is counted as a spurious propose and unsafe entry into search. This is an action-prior failure. P1 is therefore not rewarding raw willingness to generate a design; it rewards disciplined entry control.

\paragraph{P2: over-edit / wandering.}
In the macro-unimorph tip-mass design-search case derived from DOI 10.3390/mi14020421 (case 0016), the trajectory makes large moves across violation families instead of preserving the feasible neighborhood. The oracle trace shows oscillation and failure to re-enter feasibility within budget; over-edit and low directed-update scores capture this behavior. This is an edit-style failure: the model can react to feedback, but the reaction is too global and can chase one constraint while breaking another.

\paragraph{P3: continuity trap.}
In audit case \path{p3::dev::0093::0000}, the exposed history has already moved substrate thickness in a harmful direction. Continuity-biased runs keep treating that corrupted history as trusted state. P3 separates this into escape, cascade, dead-budget, and final-success fields: merely noticing the trap is not enough if the model then cascades or fails to return to a feasible design. This is why the state-summary intervention is diagnostic; it tests whether verifier-authored state isolation reduces raw-history continuity bias.

\paragraph{P4: policy mismatch.}
In audit case \path{vehbench::test_id::p4::0032}, all candidates are physically feasible, but the model ranks a local trade-off neighborhood as \(B>C>E>A>D\) while the oracle policy ranks \(C>E>D>A>B\). The resulting full Kendall tau is \(-1.0\) and the dominance-violation rate is \(1.0\). The error is therefore not search failure. It is preference-execution failure: the model does not follow the policy-conditioned ordering over feasible designs.

Together, these cases clarify why VEHBench reports stage-specific metrics instead of a single endpoint score. A P1 over-proposal can look productive but is unsafe triage; a P2 over-edit can contain valid engineering arithmetic but still destroy design-search feasibility; a P3 continuity trap can escape one violation and cascade into another; and a P4 policy mismatch can occur after the physical verifier has already accepted every candidate.

\clearpage
\section{Selection--Generation Isomorphism Audit}
\label{sec:appendix_isomorphic}

The following tables document the frozen cleanup protocol behind the SG-gap analysis in the main text. Residual failures after one cleanup pass and at most one targeted rerun are reported as final outcomes rather than recursively cleaned away.

\subsection{Selection--Generation Isomorphism Audit}

This audit bank tests whether the same underlying feasible design remains accessible when the response regime changes. We construct 46 probe groups from the hard/medium P2--P4 overlap and instantiate each group in three matched forms: \textbf{A-selection} (pick the one feasible candidate from a five-way pool), \textbf{B-generation} (synthesize a feasible candidate from a near-feasible seed), and \textbf{C-completion} (fill in two masked variables of a gold feasible design). We freeze the protocol at one cleanup pass plus at most one targeted rerun for malformed outputs; residual failures after that pass are reported as final failures rather than recursively cleaned away.

\begin{center}
\captionof{table}{Isomorphic probe results after one cleanup pass and at most one targeted rerun. Success rates are group-level percentages over 46 probe groups. SG-gap is A-selection minus B-generation. gemini-3.1-pro-preview remains partially parse-limited only on B-generation; all other model/form cells are parse-clean after the frozen cleanup protocol.}
\label{tab:appendix_isomorphic_probe}
\centering
\scriptsize
\setlength{\tabcolsep}{3pt}
\renewcommand{\arraystretch}{1.08}
\begin{tabular}{p{30mm}ccccccc}
\toprule
Model & A-sel. & B-gen. & C-comp. & SG-gap & Parse-B & Lat. (s) & Bad rows \\
\midrule
claude-4.6-sonnet & 41.3\% & 4.3\% & 17.4\% & 37.0 pts & 100.0\% & 18.1 & 0 \\
gpt-5.4 & 32.6\% & 6.5\% & 21.7\% & 26.1 pts & 100.0\% & 2.4 & 0 \\
gemini-3.1-pro-preview & 65.2\% & 4.3\% & 30.4\% & 60.9 pts & 41.3\% & 29.8 & 27 \\
\texttt{qwen3.6-plus} & 52.2\% & 8.7\% & 15.2\% & 43.5 pts & 100.0\% & 70.0 & 0 \\
\bottomrule
\end{tabular}
\end{center}
\FloatBarrier

For gemini-3.1-pro-preview, B-generation is deliberately reported on the full 46-group denominator because parse robustness is part of the generation regime. Among the 19 parse-clean B-generation rows, its feasible-generation success is 2/19 = 10.5\%; the full-denominator value remains 2/46 = 4.3\%. This distinction prevents the SG-gap from being misread as only a semantic generation failure or only a formatting failure.

\begin{center}
\captionof{table}{Protocol log for the isomorphic probe reruns. Residual bad rows after the first targeted rerun are treated as part of the measured behavior, not recursively retried until clean.}
\label{tab:appendix_isomorphic_protocol}
\centering
\scriptsize
\setlength{\tabcolsep}{3pt}
\renewcommand{\arraystretch}{1.08}
\begin{tabularx}{\textwidth}{p{28mm}Y p{28mm}Y}
\toprule
Model & First-pass issue & Final residual after one rerun & Readout \\
\midrule
gpt-5.4 & no malformed-output cleanup needed & 0 bad rows & stable baseline; all A/B/C cells remain parse-clean \\
claude-4.6-sonnet & transient transport failures in the first pass, cleaned once & 0 bad rows & clean selector after one cleanup pass; low B-generation remains a capability result, not a format artifact \\
gemini-3.1-pro-preview & first pass mixed timeouts and malformed JSON & 27 bad rows, all B-generation & residual failures are no longer timeout-dominated; they remain half-written JSON under the hardest six-variable generation form \\
\texttt{qwen3.6-plus} & four timeout rows in the first pass & 0 bad rows & fully clean after one cleanup pass, but still by far the slowest model in the audit \\
\bottomrule
\end{tabularx}
\end{center}
\FloatBarrier

\clearpage
\clearpage
\section{Tier~3 Controlled-Prompt Audit --- Full Results}
\label{sec:tier3_full}

The main text (Section~\ref{sec:experiments}) reports the four primary confirmatory cells with targeted-minus-neutral paired bootstrap deltas. This appendix provides the complete per-condition results, the C4 per-model split, and per-task pattern decomposition that support the interpretation advanced in the main text.

\subsection{Design and Execution}

The experiment was fully pre-registered. Four confirmatory cells pair known deficit models with their weak stages: C1 (gpt-5.4, P2), C2 (claude-4.6-sonnet, P3), C3 (gemini-3.1-pro-preview, P4), and C4 (gpt-5.4 and claude-4.6-sonnet, P1). Each cell samples 36--60 tasks under frozen seed 42 with subtype-level stratification. Five prompt conditions are tested per cell: \texttt{default} (the standard benchmark prompt), \texttt{neutral} (generic structured execution without boundary-specific diagnosis), \texttt{targeted} (boundary-specific diagnostic prompt designed to address the known deficit), \texttt{wrong-boundary} (a mismatched diagnostic prompt from a different stage), and for C2, a \texttt{state-summary-plus} variant stacking targeted with default. All prompts were frozen before execution; no iterative refinement occurred. The analysis uses 10,000 paired task-level bootstrap resamples with 95\% confidence intervals.

\begin{center}
\captionof{table}{Tier~3 prompt-condition logic. The audit asks whether boundary-specific prompt controls can erase the profile deficits identified in the main benchmark.}
\label{tab:tier3_prompt_conditions}
\centering
\scriptsize
\setlength{\tabcolsep}{4pt}
\renewcommand{\arraystretch}{1.08}
\begin{tabular}{lp{27mm}p{34mm}p{39mm}}
\toprule
Condition & Purpose & Used where & Prediction if profiles are prompt artifacts \\
\midrule
\texttt{default} & standard benchmark prompt & all cells & baseline behavior \\
\texttt{neutral} & controls for extra structure and length & all cells & small generic improvement at most \\
\texttt{targeted} & boundary-specific control instruction & all cells & large gain over neutral \\
\texttt{wrong-boundary} & mismatched control block & C1, C3, C4 & no gain or degradation \\
\texttt{state-summary-plus} & targeted P3 plus verifier-authored summary & C2 & extra gain if reset wording adds value beyond state representation \\
\bottomrule
\end{tabular}
\end{center}
\FloatBarrier

\subsection{Prompt Conditions}

The \texttt{default} condition uses the standard P1--P4 benchmark prompt templates in Appendix~\ref{sec:prompts}. The other conditions prepend one frozen control block before the standard task prompt. The \texttt{neutral} block controls for extra structure and length without naming a boundary-specific failure mode:

\begin{promptlisting}{Neutral Control Prompt}
SYSTEM CONTROL: Structured task execution.
Break down the task into steps. For each step:
- State what you are doing.
- Execute the step.
- Verify the result before proceeding.
Complete all steps before outputting the final answer.
\end{promptlisting}

The targeted prompts are boundary-specific. P1 targets entry discipline:

\begin{promptlisting}{Targeted P1 Control Prompt}
SYSTEM CONTROL: Triage protocol.
1. List every missing parameter needed for a valid design decision.
2. Check whether any hard constraints conflict.
3. Determine whether a feasible design exists within stated bounds.
Only output `propose_design` if all three conditions pass.
Output `request_missing_info` if condition 1 fails.
Output `declare_infeasible` if condition 2 or 3 fails.
\end{promptlisting}

P2 targets bounded verifier-guided design search rather than broad regeneration:

\begin{promptlisting}{Targeted P2 Control Prompt}
SYSTEM CONTROL: Verifier-guided design search.
- Edit exactly ONE design variable per iteration.
- Preserve constraints that are currently satisfied unless the verifier
  feedback implies a necessary trade-off.
- Target the constraint with the largest violation first.
- Report which constraint you targeted and whether it improved.
\end{promptlisting}

P3 targets skeptical reset under corrupted history:

\begin{promptlisting}{Targeted P3 Control Prompt}
SYSTEM CONTROL: Corrupted trajectory recovery.
The history below may contain steps that moved the design in a harmful
direction. Before acting:
1. Identify the direction the design moved over the last 3 steps.
2. Locate the best feasible design in the trajectory (if any).
3. Explicitly state which portion of the history you are discarding.
Proceed from the best feasible state, not from the last step.
\end{promptlisting}

P4 targets evaluator-role discipline:

\begin{promptlisting}{Targeted P4 Control Prompt}
SYSTEM CONTROL: Policy-conditioned evaluation.
You are an evaluator. Do NOT propose or modify designs.
All candidates are fixed and must be treated as given.
1. Extract from the policy the primary objective and any secondary
   objectives, with their priority ordering.
2. Compare candidates only on policy-relevant quantities.
3. Rank candidates according to the policy's priority order.
Ignore personal design preferences not stated in the policy.
\end{promptlisting}

The \texttt{wrong-boundary} condition intentionally applies an incompatible control block: P1 and P2 receive the P4 evaluator prompt, while P4 receives the P2 bounded-edit prompt. This tests whether improvement comes from generic instruction length or from matching the control block to the trusted-state boundary. The C2 \texttt{state-summary-plus} condition stacks the P3 targeted prompt with a verifier-authored state summary, allowing us to check whether explicit reset instructions add value beyond the state representation itself.

\subsection{Full Per-Condition Results}

\clearpage
\begin{center}
\captionof{table}{Tier~3 Phase~A complete per-condition results. All metrics are primary confirmatory endpoints per cell. The C1 targeted secondary objective is retained with an audit flag because it is a scale outlier; the confirmatory endpoint for C1 is feasible rate.}
\label{tab:tier3_all_conditions}
\centering
\scriptsize
\setlength{\tabcolsep}{3.5pt}
\renewcommand{\arraystretch}{1.08}
\begin{tabular}{lllcccc}
\toprule
Cell & Model & Condition & $n$ & Primary metric & Secondary & Parse \\
\midrule
C1 & gpt-5.4 & default   & 40 & 0.150 feasible & 61.0 mean obj & 0.000 \\
   &         & neutral   & 40 & 0.175 feasible & 4.3 mean obj  & 0.000 \\
   &         & targeted  & 40 & 0.150 feasible & 461.8$^\dagger$ mean obj & 0.000 \\
   &         & wrong     & 40 & 0.075 feasible & 2.1 mean obj  & 0.000 \\
\midrule
C2 & claude-4.6-sonnet & default   & 36 & 0.167 success & 0.694 cascade & 0.000 \\
   &            & neutral   & 36 & 0.389 success & 0.528 cascade & 0.000 \\
   &            & targeted  & 36 & 0.194 success & 0.778 cascade & 0.000 \\
   &            & wrong     & 36 & 0.222 success & 0.694 cascade & 0.000 \\
   &            & s+        & 36 & 0.222 success & 0.750 cascade & 0.000 \\
\midrule
C3 & gemini-3.1-pro-preview & default   & 50 & 0.692 $\tau$ & 0.320 exact & 0.000 \\
   &            & neutral   & 50 & 0.692 $\tau$ & 0.320 exact & 0.000 \\
   &            & targeted  & 50 & 0.708 $\tau$ & 0.320 exact & 0.000 \\
   &            & wrong     & 50 & 0.700 $\tau$ & 0.300 exact & 0.000 \\
\midrule
C4 & claude-4.6-sonnet & default   & 60 & 0.650 acc & --- & 0.050 \\
   &            & neutral   & 60 & 0.617 acc & --- & 0.050 \\
   &            & targeted  & 60 & 0.567 acc & --- & 0.183 \\
   &            & wrong     & 60 & 0.633 acc & --- & 0.050 \\
\cmidrule{2-7}
   & gpt-5.4    & default   & 60 & 0.433 acc & --- & 0.000 \\
   &            & neutral   & 60 & 0.400 acc & --- & 0.000 \\
   &            & targeted  & 60 & 0.467 acc & --- & 0.000 \\
   &            & wrong     & 60 & 0.400 acc & --- & 0.000 \\
\bottomrule
\end{tabular}
\end{center}
\FloatBarrier

Table~\ref{tab:tier3_all_conditions} reports the complete per-condition primary metric. The C1 targeted mean-objective value marked with $\dagger$ is a secondary-display scale outlier relative to the rest of the C1 rows; we leave it visible for auditability but do not use it for the Tier~3 conclusion. Four patterns merit attention beyond the main-text targeted-minus-neutral deltas:

\begin{enumerate}[leftmargin=5mm,itemsep=1mm,topsep=1mm]
    \item \textbf{C2 neutral improvement.} The generic structured neutral prompt lifts claude-4.6-sonnet P3 success from 0.167 to 0.389 (95\% CI for neutral$-$default: [0.083, 0.361]), while reducing cascade from 0.694 to 0.528. This is the largest single-condition effect in the experiment and suggests that claude-4.6-sonnet benefits from structured execution guidance but not from skeptical-reset framing.
    \item \textbf{C2 cascade--success inversion.} Cascade rate under targeted (0.778) is the highest across all C2 conditions, and targeted is the only condition where cascade exceeds the default rate. Per-task analysis identifies four tasks with cascade exclusively under targeted, consistent with the interpretation that the skeptical-reset prompt induces overcorrection.
    \item \textbf{C2 state-summary-plus does not rescue targeted.} The \texttt{s+} condition raises success only from 0.194 to 0.222 relative to targeted and leaves cascade high at 0.750. This is informative because it shows that simply stacking state-summary representation onto the same targeted wording does not recover the neutral-prompt gain.
    \item \textbf{C4 parse-error interaction.} claude-4.6-sonnet's parse-error rate under targeted P1 (0.183) is more than triple the rate under other conditions (0.050). Parse errors concentrate in \texttt{infeasible\_margin} tasks (4/7) and \texttt{declare\_infeasible} gold labels (5/14). Excluding parse errors, claude-4.6-sonnet's classification accuracy under targeted is 0.694, exceeding neutral (0.617), indicating that the targeted prompt content aids reasoning while its length impairs output-format compliance.
\end{enumerate}

\subsection{Per-Task Decomposition}

\paragraph{C2 per-task transitions.} Of 36 tasks, 20 fail under all three main conditions; 7 succeed under neutral but fail under default and targeted; 4 succeed under all conditions. The neutral$\to$targeted transition loses 9 successes and gains 2. Cascade under targeted increases in 10 tasks relative to neutral and decreases in only 1.

\paragraph{C3 per-task decomposition.} The +0.016 mean $\tau$ gain is driven entirely by one task whose $\tau$ moves from $-$0.4 (negatively correlated ranking) under neutral to +0.4 under targeted; the remaining 49 tasks show identical $\tau$ values across both conditions. Top-1 accuracy improves on 2 tasks (94\%$\to$100\%). The targeted prompt thus eliminates a single ranking failure rather than systematically improving order quality.

\paragraph{C1 subtype pattern.} gpt-5.4 never solves \texttt{boundary\_binding} tasks under any condition (0/40). The \texttt{paper\_like} subtype accounts for most feasible cases across all conditions (19/22 total feasible designs). This subtype specificity, together with the uniformly low feasible rates (0.075--0.175), indicates a capability floor that prompt variation does not shift.

\subsection{Interpretive Summary}

The four-cell audit supports the following interpretation, which we advance in the main text: behavioral fingerprints identified by Tier~1 profiles are not shallow prompt artifacts. If they were, targeted boundary-specific prompts should erase them. Instead, C1 shows no effect, C2 shows significant reversal, C3 shows a one-task marginal effect, and C4 shows model-dependent interaction. The resilience of stage dissociation to prompt-level control reinforces the benchmark's construct validity: what VEHBench measures is durable prior-boundary compatibility, not prompt sensitivity.

\clearpage
\section{Circuit Audit: Cross-Domain Construct Validity}
\label{sec:appendix_circuit_pilot}

This appendix reports the final circuit audit described in Section~\ref{sec:experiments}. The audit is a construct-validity check, not an independent circuit benchmark: it asks whether the same P1--P4 diagnostic regimes remain discriminative under a second closed-form engineering domain. The released task bank hardens P1 with near-boundary infeasible and missing-information cases, hardens P2 with dual-constraint design-search tasks where fixing one violation can break another, and uses matched P3/P4 recovery and ranking banks to test corrupted-state recovery and policy-conditioned selection. P1/P2 use the hardened final audit bank, while P3/P4 use the retained original audit bank to preserve the paper-facing snapshot.

\Needspace{12\baselineskip}
\begin{center}
\captionof{table}{Circuit audit task inventory. The task bank changes the physics while preserving the P1--P4 decision boundaries.}
\label{tab:appendix_circuit_inventory}
\scriptsize
\setlength{\tabcolsep}{3pt}
\renewcommand{\arraystretch}{1.08}
\begin{tabularx}{\textwidth}{lcYYY}
\toprule
Probe & n & Source & Hardening & Validation target \\
\midrule
P1 & 32 & final circuit audit & Near-boundary infeasible cases; lower raw \texttt{propose\_design} prior; subtype-balanced entry decisions & Action-discipline transfer \\
P2 & 32 & final circuit audit & Dual constraints; update directions can conflict; objective-preserving feasible design search is rewarded & Edit style and feedback conditioning \\
P3 & 18 & final circuit audit & Progressive dual traps; escape can trigger a second violation & State-reset behavior and trap sensitivity \\
P4 & 24 & final circuit audit & Policy-flip ranking tasks under feasible candidate pools & Policy execution \\
\bottomrule
\end{tabularx}
\end{center}

\paragraph{Why P1-Composite is the circuit P1 headline.}
The circuit P1 bank follows the VEH P1 hardening principle: raw accuracy is too easy to inflate when the action distribution contains many feasible proposals. The bank therefore lowers the trivial \texttt{propose\_design} prior and adds near-boundary infeasible/request cases. P1-Composite is used as the headline because it jointly rewards correct entry actions and penalizes spurious proposals, unsafe proposals, missed missing-information cases, and missed infeasibility cases. This makes P1 an entry-discipline metric rather than a willingness-to-generate metric.

\Needspace{12\baselineskip}
\begin{center}
\captionof{table}{Circuit P1 action triage. P1-Comp. is the headline; abbreviations: Spur.=spurious propose, Unsafe=unsafe propose, Req.=request recall, Inf.=infeasible recall.}
\label{tab:appendix_circuit_p1}
\scriptsize
\setlength{\tabcolsep}{3.5pt}
\renewcommand{\arraystretch}{1.08}
\begin{tabular}{lccccccccc}
\toprule
Model & n & P1-Comp. & Acc. & F1 & Spur. & Unsafe & Req. & Inf. & Parse \\
\midrule
qwen3-max & 32 & 0.971 & 0.969 & 0.965 & 0.000 & 0.000 & 1.000 & 1.000 & 0.000 \\
hunyuan-hy3-preview & 32 & 0.944 & 0.938 & 0.933 & 0.000 & 0.000 & 1.000 & 1.000 & 0.000 \\
deepseek-r1 & 32 & \textbf{0.974} & 0.969 & 0.971 & 0.000 & 0.000 & 1.000 & 1.000 & 0.000 \\
gemini-3.1-pro-preview & 32 & \textbf{0.974} & 0.969 & 0.971 & 0.000 & 0.000 & 1.000 & 1.000 & 0.000 \\
gpt-5.4 & 32 & 0.923 & 0.906 & 0.917 & 0.000 & 0.000 & 1.000 & 1.000 & 0.000 \\
claude-4.6-sonnet & 32 & 0.932 & 0.938 & 0.940 & 0.031 & 0.031 & 1.000 & 0.875 & 0.000 \\
\bottomrule
\end{tabular}
\end{center}

\Needspace{12\baselineskip}
\begin{center}
\captionof{table}{Circuit P2 verifier-guided design search. P2b is the final feasible objective score; abbreviations: Feas.=final feasible rate, Dir.=directed update, O-edit=over-edit.}
\label{tab:appendix_circuit_p2}
\scriptsize
\setlength{\tabcolsep}{4pt}
\renewcommand{\arraystretch}{1.08}
\begin{tabular}{lcccccc}
\toprule
Model & n & Feas. & P2b & Dir. & O-edit & Parse \\
\midrule
qwen3-max & 32 & 0.906 & 0.494 & 0.938 & 0.400 & 0.000 \\
hunyuan-hy3-preview & 32 & 0.938 & 0.512 & 0.914 & 0.771 & 0.000 \\
deepseek-r1 & 32 & 0.969 & 0.594 & \textbf{1.000} & 0.605 & 0.000 \\
gemini-3.1-pro-preview & 32 & \textbf{1.000} & \textbf{0.662} & 0.969 & 0.938 & 0.000 \\
gpt-5.4 & 32 & 0.875 & 0.457 & 0.883 & 0.489 & 0.000 \\
claude-4.6-sonnet & 32 & 0.906 & 0.487 & 0.914 & 0.646 & 0.000 \\
\bottomrule
\end{tabular}
\end{center}

\Needspace{12\baselineskip}
\begin{center}
\captionof{table}{Circuit P3 corrupted-state recovery retained from the original bank. Abbreviations: Esc.=escape, Replan=explicit replan, Reset=history reset, Casc.=cascade, Dead=dead budget, Succ.=final success.}
\label{tab:appendix_circuit_p3}
\scriptsize
\setlength{\tabcolsep}{3.5pt}
\renewcommand{\arraystretch}{1.08}
\begin{tabular}{lccccccccc}
\toprule
Model & n & Esc. & Replan & Reset & Casc. & Dead & Succ. & Rec. q. & Parse \\
\midrule
qwen3-max & 18 & 0.944 & 0.000 & 0.000 & 0.471 & 0.000 & 0.778 & 0.366 & 0.000 \\
gemini-3.1-pro-preview & 18 & 1.000 & 0.000 & 0.000 & \textbf{0.000} & 0.000 & \textbf{1.000} & \textbf{0.611} & 0.000 \\
gpt-5.4 & 18 & 0.944 & 0.111 & 0.111 & 0.176 & 0.056 & 0.778 & 0.401 & 0.000 \\
deepseek-r1 & 18 & 0.778 & 0.000 & 0.000 & \textbf{0.000} & 0.000 & 0.722 & 0.314 & 0.000 \\
hunyuan-hy3-preview & 18 & 0.889 & 0.000 & 0.000 & 0.062 & 0.000 & 0.778 & 0.364 & 0.000 \\
claude-4.6-sonnet & 18 & 1.000 & 0.000 & 0.000 & \textbf{0.000} & 0.000 & \textbf{1.000} & 0.568 & 0.000 \\
\bottomrule
\end{tabular}
\end{center}

\Needspace{12\baselineskip}
\begin{center}
\captionof{table}{Circuit P4 policy-conditioned ranking. Abbreviations: Top2=set agreement for the top two candidates; Flip=policy-flip accuracy.}
\label{tab:appendix_circuit_p4}
\scriptsize
\setlength{\tabcolsep}{3.5pt}
\renewcommand{\arraystretch}{1.08}
\begin{tabular}{lccccccccc}
\toprule
Model & n & Tau & Exact & Top1 & Top2 & Pair & Flip & BARS & Parse \\
\midrule
qwen3-max & 24 & 0.608 & 0.375 & 0.583 & 0.583 & 0.804 & 0.769 & 0.710 & 0.000 \\
gemini-3.1-pro-preview & 24 & 0.475 & 0.250 & 0.500 & 0.500 & 0.738 & 0.679 & 0.625 & 0.000 \\
gpt-5.4 & 24 & \textbf{0.625} & 0.375 & 0.583 & 0.625 & 0.812 & \textbf{0.801} & \textbf{0.722} & 0.000 \\
deepseek-r1 & 24 & 0.542 & 0.292 & 0.583 & 0.500 & 0.771 & 0.716 & 0.661 & 0.000 \\
hunyuan-hy3-preview & 24 & 0.492 & 0.208 & 0.458 & 0.417 & 0.746 & 0.647 & 0.614 & 0.000 \\
claude-4.6-sonnet & 24 & 0.608 & 0.375 & 0.542 & 0.500 & 0.804 & 0.759 & 0.707 & 0.000 \\
\bottomrule
\end{tabular}
\end{center}

\Needspace{10\baselineskip}
\begin{center}
\captionof{table}{Circuit profile scores. Action/edit scores use the updated P1/P2 audit; state/policy scores are retained from the original P3/P4 banks.}
\label{tab:appendix_circuit_profile}
\scriptsize
\setlength{\tabcolsep}{5pt}
\renewcommand{\arraystretch}{1.08}
\begin{tabular}{lcccc}
\toprule
Model & Action & Edit & State & Policy \\
\midrule
qwen3-max & 0.983 & 0.877 & 0.650 & 0.706 \\
hunyuan-hy3-preview & 0.968 & 0.792 & 0.721 & 0.612 \\
deepseek-r1 & \textbf{0.985} & 0.579 & 0.700 & 0.672 \\
gemini-3.1-pro-preview & \textbf{0.985} & 0.766 & \textbf{0.800} & 0.633 \\
gpt-5.4 & 0.956 & 0.830 & 0.720 & \textbf{0.714} \\
claude-4.6-sonnet & 0.962 & 0.815 & \textbf{0.800} & 0.696 \\
\bottomrule
\end{tabular}
\end{center}

Table~\ref{tab:appendix_cross_domain_patterns} summarizes the signature patterns that reproduce across VEH and circuit domains.

\Needspace{14\baselineskip}
\begin{center}
\captionof{table}{Cross-domain pattern reproduction. The circuit audit checks whether the same diagnostic reversals appear under different physics.}
\label{tab:appendix_cross_domain_patterns}
\scriptsize
\setlength{\tabcolsep}{3pt}
\renewcommand{\arraystretch}{1.08}
\begin{tabularx}{\textwidth}{p{31mm}YY}
\toprule
Pattern & VEH observation & Circuit observation \\
\midrule
gpt-5.4 P4 lead / weaker search & P4 Tau 0.887 (1st); P2 ratio 0.133 & retained P4 Tau 0.625 (1st); circuit P2b 0.457 (6/6) \\
gemini-3.1-pro-preview P2 lead / weak P4 & P2 ratio 0.390 (1st); P4 Tau 0.824 (7/12) & circuit P2b 0.662 (1st); retained P4 Tau 0.475 (6/6) \\
No single model dominates & qwen3-max P1; gemini-3.1-pro-preview P2; hunyuan-hy3-preview P3; gpt-5.4 P4 & gemini-3.1-pro-preview/deepseek-r1 P1; gemini-3.1-pro-preview P2; gemini-3.1-pro-preview/claude-4.6-sonnet retained P3; gpt-5.4 retained P4 \\
P3 trap-geometry sensitivity & Multi-step contamination penalizes continuity-biased models (gemini-3.1-pro-preview cascade 0.507 in VEH) & Formula-level circuit traps give gemini-3.1-pro-preview zero cascade, while qwen3-max cascades at 0.471; this is a trap-geometry contrast, not a claim that the same model-level cascade mechanism transfers unchanged \\
Profile diagonal direction & Action discipline tracks P1; edit style tracks P2; policy execution tracks P4 & P1-Composite, P2b, and P3 success remain separable under the circuit oracle \\
\bottomrule
\end{tabularx}
\end{center}

\paragraph{What the circuit audit proves and does not prove.}
The audit supports a narrow construct-validity claim: when the physics changes from VEH to closed-form circuit design, the P1--P4 scaffold still produces separable action, design-search, recovery, and ranking behaviors. It also shows that response-control reversals such as gemini-3.1-pro-preview's search/ranking split and gpt-5.4's ranking/search split are not artifacts of VEH-specific formulas. It does \emph{not} establish stable circuit-domain model rankings, cover modern circuit design broadly, or replace a full circuit benchmark; the task count and family coverage are intentionally pilot-scale.

\clearpage
\section*{NeurIPS Paper Checklist}

\begin{enumerate}

\item {\bf Claims}
    \item[] Question: Do the main claims made in the abstract and introduction accurately reflect the paper's contributions and scope?
    \item[] Answer: \answerYes{}
    \item[] Justification: The abstract and introduction (§1) state that VEHBench evaluates LLM-assisted vibration energy harvester co-design through four verifier-grounded design roles (P1--P4), that frontier models dissociate across these roles, that response-control profiles provide a diagnostic account of the dissociation, and that stage-local results support stage-aware model selection and control. Each claim is supported by the experimental evidence in §4, with robustness and scope checks reported in the appendix and limitations discussed in §5.

\item {\bf Limitations}
    \item[] Question: Does the paper discuss the limitations of the work performed by the authors?
    \item[] Answer: \answerYes{}
    \item[] Justification: Section 5 states the main scope boundary: VEHBench is scoped to analytically verifiable cantilever VEH co-design and does not replace FEM or hardware certification. The appendix details the construct-validity circuit pilot, P1 matched certification split, P3 intervention subset, SG-gap audit, prompt-control audit, profile definitions, uncertainty, split coverage, and closed-API snapshot drift (Appendix~\ref{sec:appendix_circuit_pilot}, Appendix~\ref{sec:appendix_isomorphic}, Appendix~\ref{sec:tier3_full}, Appendix~\ref{sec:metric_formulas}).

\item {\bf Theory assumptions and proofs}
    \item[] Question: For each theoretical result, does the paper provide the full set of assumptions and a complete (and correct) proof?
    \item[] Answer: \answerNA{}
    \item[] Justification: This paper does not include theoretical results. It is a benchmark and empirical evaluation paper. The VEH oracle uses closed-form engineering equations documented in Appendix~\ref{sec:appendix_veh_oracle_formulas}; the circuit construct-validity audit is documented in Appendix~\ref{sec:appendix_circuit_pilot}.

\item {\bf Experimental result reproducibility}
    \item[] Question: Does the paper fully disclose all the information needed to reproduce the main experimental results of the paper to the extent that it affects the main claims and/or conclusions of the paper?
    \item[] Answer: \answerYes{}
    \item[] Justification: Section 3 describes the benchmark construction pipeline (209-paper extraction audit, 52 cleaned anchors, source-anchor separated optimizer-facing splits, P1 matched certification-style triage, manifest-backed provenance). Section 4.1 describes the experimental setup (12 complete model runs, headline metrics, shared oracle). The Appendix provides full metric definitions, P1--P4 prompt templates, oracle specifications, run configuration, and the CMA-ES baseline configuration. All task banks, evaluator code, and per-model JSONL logs are released with the artifact.

\item {\bf Open access to data and code}
    \item[] Question: Does the paper provide open access to the data and code, with sufficient instructions to faithfully reproduce the main experimental results, as described in supplemental material?
    \item[] Answer: \answerYes{}
    \item[] Justification: The release artifact is available at \href{https://huggingface.co/datasets/AnonymousVehbench/vehbench}{huggingface.co/datasets/AnonymousVehbench/vehbench}. The artifact includes P1--P4 JSONL task banks, manifests, oracle and evaluator code, generation scripts, split reports, prompt templates, per-model JSONL logs, licenses, and Croissant RAI metadata. The dataset is released under CC-BY-4.0; code under MIT license. Build and evaluation commands are provided in the Appendix and artifact README.

\item {\bf Experimental setting/details}
    \item[] Question: Does the paper specify all the training and test details (e.g., data splits, hyperparameters, how they were chosen, type of optimizer) necessary to understand the results?
    \item[] Answer: \answerYes{}
    \item[] Justification: This paper evaluates pre-trained frontier LLMs as engineering agents; no training is performed. Evaluation details are specified in §4.1 and the Appendix: temperature 0.0, max output tokens provider-default (4096--8192), up to 2 retries on parse failure, P2 stops on feasible closure, P3 uses full budget. Task splits (dev/test\_id/test\_ood) are described in §3. Model provider endpoints and run dates are recorded in per-model manifests. All task generation uses fixed random seeds.

\item {\bf Experiment statistical significance}
    \item[] Question: Does the paper report error bars suitably and correctly defined or other appropriate information about the statistical significance of the experiments?
    \item[] Answer: \answerYes{}
    \item[] Justification: Table~\ref{tab:main_results} reports the main headline scores, while Appendix Table~\ref{tab:appendix_ci_table} reports 95\% confidence intervals and the key paired deltas for the best point estimates. Appendix Table~\ref{tab:appendix_p3_intervention_delta_ci} reports paired task-bootstrap 95\% confidence intervals for the P3 state-summary intervention. Spearman profile-stage and stage-rank correlations are reported as descriptive diagnostics. The CI methodology (non-parametric bootstrap, Wilson intervals for binary proportions, and paired resampling for deltas) is described in the Appendix.

\item {\bf Experiments compute resources}
    \item[] Question: For each experiment, does the paper provide sufficient information on the computer resources (type of compute workers, memory, time of execution) needed to reproduce the experiments?
    \item[] Answer: \answerYes{}
\item[] Justification: All experiments use API-based inference with no local GPU compute. Model runs use provider APIs (temperature 0.0, max output tokens 4096--8192). The analytical oracle (VEH) runs at ~0.5 ms per evaluation on CPU; the circuit oracle is faster. CMA-ES baseline on the full P2 bank (208 tasks × 40 oracle calls) completes in under 1 second on a standard laptop. Total experimental API cost is approximately USD 300--500 across all models and probes. No training runs were performed.

\item {\bf Code of ethics}
    \item[] Question: Does the research conducted in the paper conform, in every respect, with the NeurIPS Code of Ethics?
    \item[] Answer: \answerYes{}
    \item[] Justification: The research conforms to the NeurIPS Code of Ethics. No human subjects were involved. The benchmark uses publicly available engineering design principles and published literature data; no private or sensitive data was collected. The LLMs evaluated are publicly accessible models. The artifact is released under open licenses (CC-BY-4.0 for data, MIT for code) with documented intended use as an engineering-agent diagnostic benchmark, explicitly noting that it is not intended for production safety certification.

\item {\bf Broader impacts}
    \item[] Question: Does the paper discuss both potential positive societal impacts and negative societal impacts of the work performed?
    \item[] Answer: \answerYes{}
    \item[] Justification: Section 3 describes the release boundary and states that VEHBench is intended as a diagnostic tool for engineering-agent evaluation, not for production safety certification or as a standalone engineering solver. The Discussion (§5) provides deployment guidance for verifier-gated and stage-routed systems. Potential positive impact includes more auditable engineering-agent evaluation; the main negative risk is over-trusting benchmark scores as evidence of safe real-world hardware design. This is mitigated by intended-use documentation, analytical-oracle scope limits, and release metadata.

\item {\bf Safeguards}
    \item[] Question: Does the paper describe safeguards that have been put in place for responsible release of data or models that have a high risk for misuse?
    \item[] Answer: \answerNA{}
    \item[] Justification: This paper releases a benchmark dataset and evaluation code, not a deployable model. The dataset consists of closed-form engineering design tasks (VEH and circuit domains) with analytical oracles. It poses no risk of misuse for disinformation, surveillance, or generation of harmful content. The artifact documentation includes an intended use statement explicitly scoping the benchmark to research on engineering-agent evaluation.

\item {\bf Licenses for existing assets}
    \item[] Question: Are the creators or original owners of assets (e.g., code, data, models), used in the paper, properly credited and are the license and terms of use explicitly mentioned and properly respected?
    \item[] Answer: \answerYes{}
    \item[] Justification: Prior engineering benchmarks and task families (EngDesign, BuildArena, mechanical-design agents, DesignQA, and MSEval) are cited in §2, with expanded related work in the Appendix. The VEH domain is supported by modern VEH/IoT harvesting reviews, Erturk--Inman analytical VEH references, and self-powered piezoelectric nanogenerator work. Evaluated LLM models are cited with provider documentation references where available. The CMA-ES baseline is cited through the CMA-ES reference.

\item {\bf New assets}
    \item[] Question: Are new assets introduced in the paper well documented and is the documentation provided alongside the assets?
    \item[] Answer: \answerYes{}
    \item[] Justification: VEHBench is a new benchmark. Documentation includes task schemas (Appendix~\ref{sec:prompts}), metric definitions with formulas (Appendix~\ref{sec:metric_formulas}), construction pipeline and split policy (§3), P1--P4 prompt templates (Appendix~\ref{sec:prompts}), oracle specifications with equations (Appendix~\ref{sec:appendix_veh_oracle_formulas}), Croissant RAI metadata, release manifests, JSONL task banks, evaluator/oracle code, model logs, and manifest-backed provenance. The artifact is released under CC-BY-4.0 (data) and MIT (code).

\item {\bf Crowdsourcing and research with human subjects}
    \item[] Question: For crowdsourcing experiments and research with human subjects, does the paper include the full text of instructions given to participants and screenshots, if applicable, as well as details about compensation (if any)?
    \item[] Answer: \answerNA{}
    \item[] Justification: This paper does not involve crowdsourcing or human subjects research. All evaluations are automated via LLM API calls to pre-trained models. The benchmark uses an analytical physics oracle rather than human annotation for ground truth.

\item {\bf Institutional review board (IRB) approvals or equivalent for research with human subjects}
    \item[] Question: Does the paper describe potential risks incurred by study participants, whether such risks were disclosed to the subjects, and whether Institutional Review Board (IRB) approvals (or an equivalent approval/review based on the requirements of your country or institution) were obtained?
    \item[] Answer: \answerNA{}
    \item[] Justification: This paper does not involve human subjects research. All data is either derived from published engineering literature or synthetically generated via deterministic oracles. No human participants were recruited, surveyed, or studied.

\item {\bf Declaration of LLM usage}
    \item[] Question: Does the paper describe the usage of LLMs if it is an important, original, or non-standard component of the core methods in this research?
    \item[] Answer: \answerYes{}
    \item[] Justification: LLMs are the primary subjects of evaluation in this benchmark paper. All 12 evaluated frontier models are LLMs used as engineering design agents. Their usage is fully described in §4.1 (experimental setup) and Appendix~\ref{sec:prompts} (prompt templates, run configuration). LLM assistants were also used during manuscript preparation for code/debugging assistance, figure integration, formatting migration, and language editing; all scientific claims, reported numbers, and final text were reviewed by the authors.

\end{enumerate}

\end{document}